*Full Research Paper*

# Three Dogmas, a Puzzle and its Solution

## Elnaserledinellah Mahmood Abdelwahab*

GridSAT Stiftung, Hanover (Germany)

## ABSTRACT

Modern Logics, as formulated notably by Frege, Russell and Tarski involved basic assumptions about Natural Languages in general and Indo-European Languages in particular, which are contested by Linguists. Based upon those assumptions, formal Languages were designed to overcome what Logicians claimed to be *'defects'* of Natural Language. In this paper we show that those assumptions contradict basic principles of Arabic. More specifically: The Logicians ideas, that within Natural Language words refer to objects, *'ToBe'*-constructions represent identity statements, Indefinite Descriptions must be replaced by existential quantifiers to form meaningful Sentences and Symbols can have no interpretation-independent meanings, are all falsified using undisputed principles of Arabic. The here presented falsification serves two purposes. First, it is used as a factual basis for the rejection of approaches adopting Semantic axioms of Mathematical Logics as Models for meaning of Arabic Syntax. Second, it shows a way to approach the important computational problem: Satisfiability (*SAT*). The described way is based upon the realization that parsing Arabic utilizes the existence of *'meaning-particles'* within Syntax to efficiently recognize words, phrases and Sentences. Similar meaning-particles are shown to exist in *3CNF* formulas, which, when properly handled within the machinery of *3SAT-Solvers*, enable structural conditions to be imposed on formulas, sufficient alone to guarantee the efficient production of non-exponentially sized Free Binary Decision Diagrams (*FBDDs*). We show, why known exponential Lower Bounds on sizes of *FBDDs* do not contradict our results and reveal practical evidence, obtained for multiplication circuits, supporting our claims.

*Corresponding author: elnaser@gridsat.io

# I-    Introduction and Motivation

Adequate formal treatment of Natural Language has become, at least since the emergence of computational devices and human-machine interactions, a necessary condition for the realization of practical information Systems. Such a treatment usually includes three parts: *Syntax*, *Semantics* and *Computation*, the latter being a direct consequence of assumed relations between the former two.

But even before the era of computation there has been a lot of interest in analyzing Language in a formal way, especially in the context of studies related to the logical foundations of mathematics. Although those studies aimed at building artificial Languages, they were in reality a by-product of broader linguistic contemplations of notorious western philosophers and mathematicians, whose views on purpose and nature of Language are reflected in the state-of-the-art Model theory of first order logics (see: [1]):

1- A *structure S* is a set of objects of a given Domain of Discourse, together with interpretation *I* of each of the Symbols of a given signature $\sum$ as relation or function on *D*.

2- The structure $S = <D, I>$ is said to *Model* a set of Sentences *T* formed using $\sum$, if each Sentence *t* in *T* is true for objects of *D*, when all Symbols are interpreted as given in *I*.

3- The Semantic Truth Definition for any Sentence *t* follows the rule: *'t' is true iff 't' is satisfied by any object in D; 't' is false iff 't' is satisfied by no object in D.*

4- *Truth* is thus defined as a special case of *Satisfaction*: Open formulas, i.e., the ones which contain free variables, are satisfied or not, depending upon how those variables are interpreted in *D*, but Sentences (i.e., formulas without free variables) are just *true* or *false*.

Take for example the formula 'x is a man'. Let *D* consist of names of men and women. This formula is satisfied by putting for x: *'Ali'*, but not *'Susan'*. Furthermore, the Sentence *'Ali is a man'* is *true* in structure *S*, but the Sentence *'Susan is a man'* is *false* in *S*. Thus: Satisfaction converts open formulas into true Sentences, non-satisfaction converts them to false ones.

5- Satisfaction procedures require a Semantic component called an *Assignment Function*. This function maps individual variables in formulas to objects of *D*. The interpretation of expressions containing variables is thus relative to both a Model and an Assignment Function.

6- Satisfaction procedures implement Assignment Functions in a way completely independent of both Semantics (Models) and Syntax (formulas). There is neither a logical nor a structural reason for the choice of a particular variable to refer to a particular object in *D* or for selecting a particular order in which variable assignments are made. In this work we call this property: *Selection Arbitrariness*[1].

---

[1] Although as per postulate 4, *Truth* is a special case of satisfaction, i.e., our intuitions about when a Sentence is *true* can guide our intuitions about assigning objects to free variables, none of this could enter into the formal definition of *Truth*, since, for Tarski: *'taking a variable as a name of an object is a Semantic notion'*, and his *Truth*

We can trace the above postulates back to the work of Frege [3], whose linguistic views were influenced by what has been called: *'Language Paradoxes'*. In this paper we call those views, to which also Russell and Tarski largely contributed: ***Dogmas***, because they are rarely questioned.

We select the three in our opinion most influential ones for our investigation:

a- ***Dogma1***: Words and phrases of Natural Language refer to objects of the world or relations among those objects, while Sentences refer to Truth values (this is reflected in all the above postulates).

b- ***Dogma2:*** Indefinite Descriptions must be replaced by existential assertions in order for Sentences of Natural Language to mean something[2] (substitute *'Indefinite Descriptions'* with *'free variables'* in postulate 4).

c- ***Dogma3:*** Symbols have, independent of a particular interpretation, which makes them refer to objects or relations between those objects, *no meanings for themselves*. An interpretation attributes meanings to Symbols in an *arbitrary* way: There is no reason why the string *'liar'*, for example, must denote *'the one who does not tell the Truth'*. In an interpretation $I$ of a structure $S$: A string like *'xyz'* can do the same. Put in another way: Symbol structure **cannot** reveal any intrinsic meaning (we call this claim: *'no-meaning postulate'*).

One consequence of this Dogma, used in Tarskian formal Systems, is:

> ***Consequence3:*** *Free variables can be renamed in open formulas without altering meaning nor loosing generality*[3].

When Generative Grammar emerged and studies showed that Syntax of Natural Languages possesses a high degree of structure, allowing detailed formal investigations, attempts were made to enforce the above *Dogmas* on Semantic Models of Natural Language.

This is best illustrated by the volume [4], which contains contributions from generative Linguists, Logicians and Philosophers, whose unifying leitmotif was the refusal of the purely syntactic nature assigned to *'deep structure'* representations in Chomsky's Standard Theory. The common claim was that *deep structure* of a Natural Language Sentence is to be identified with *Logical Form* (*LF*)[4], that is to say, it must be equated with the place in which the hidden logical structure of Natural Language is explicitly encoded.

---

definition had to be built, other than on set theory, upon notions from Syntax only (cf. [2]).

[2] Recall that existentially quantified formulas like: '$\exists x$ P($x$)' are true **iff** there is a way to choose an object of $D$ for $x$ such that P($x$) is satisfied.

[3] Compare this consequence to the way all programming Languages work today: *Naming conventions of used variables are left to the human user.*

[4] While there was a difference between what Logicians understood under *LF* and what Linguists understood, for a long time there was consensus that *LF* is hidden behind the ordinary or: *'Surface structure'* of a Sentence.

Approximately in the same time Montague Grammar [5] emerged and soon became the basis for many modern attempts of formalizing Semantics, not only of English, but recently also of Arabic (e.g., see [6], [7]).

Montague adopted a version of Frege's distinction between *'Sense'* and *'Reference'*. Although he did acknowledge a hierarchy of senses like Frege, he did not employ it for the analysis of iterated indirect contexts. Instead, he identified Frege's senses with *'Intensions'* along the lines of Carnap: Set theoretic functions on a logical space of possible worlds (or world-time-pairs), whose values are the references of expressions, their *'Extensions'*. In particular, the way in which a Description refers to its referent is captured by its dependence on contingent facts. The famous Fregean Venus-Descriptions differ in intension as long as there is a possible world, in which the brightest star at dawn is not the same object as the brightest star at night [8]. Technically, his formal System was realized using higher order Logics and Lambda Calculus, making use of the notions of *Intensional Logics*, via Kripke possible world Models.

Where we stand today in terms of success of Logic-based approaches to Semantics of Natural Language is best illustrated by the following reply of Chomsky to the authors questions about this subject (as well as *Partee's Paradoxes*, which are handled in detail in part A below) [9]:

>  *"The idea of relating deep structure to logical form was given up in the 1960s, with the discovery of surface structure effects on meaning. Decades ago deep structure was given up altogether as superfluous. I don't see the force of Partee's paradox. It presupposes some simple-minded notion of denotation/reference that doesn't hold for Language. It follows that Montague Grammar and its offshoots are forms of syntax in the technical sense of Frege-Tarski, etc., and we have to evaluate them in those terms, with no pretense that they are reaching outside the symbolic System."*

In part A of this paper, we strengthen the above opinion of Chomsky by showing that *the above three Dogmas cannot apply to Arabic.*

We do this by comparing them to basic, uncontested Arabic Language principles, known for more than a millennium now. We demonstrate, in particular, that many of the claimed paradoxes, attributed to Language and linked to those *Dogmas*, do not occur within Arabic in the first place, shedding a strong shadow of doubt upon their correctness and/or universality.

We show in particular that:

1- *Sense* and *Reference* related ideas postulated by Frege

2- Usages of the verb *'ToBe'* reflecting identity statements

3- Indefinite Descriptions, the way Russell has seen them

4- The *no-meaning postulate* of *Dogma3*

can all be refuted using Arabic Language principles, which we call here: *Anti-Dogmas*.

We provide, in addition to this overwhelming counter-evidence from ancient and modern Arabic Linguistics, a list of basic examples, two of which become *anomalies*, occurring when Dogmas of Logicians are used to obtain meanings of Natural Language Sentences. In addition, many so called *'quantifier scope ambiguities'*, motivating the use of *LF* as a basis for *deep structure*, are seen to be either avoidable or non-existent in Arabic. Arabic Sentences containing quantifier expressions are shown to use readily available Grammar constructs to overcome ambiguity, whenever it is possible.

Our findings may serve as arguments against the adoption of Logic-based formalisms, pretending to capture Semantics of Arabic, including Montague-type Models.

Is there any other reason, why one might be unwilling to adopt *Dogmas* of Logicians, when designing modern information Systems?

There is the computational aspect.

Computation paradigms are burdened by the intractability of thousands of practically important problems, all of which can be reduced to the notorious *SAT*, the problem of efficiently linking Syntax to Semantics of simple Propositional Logics formulas, upon which *Dogmas* cast also their dark shadow.

There have been many solution attempts of *SAT*, all of them, except those published by the author in [10], [11] and [12], have one thing in common: *They neither question those Dogmas, nor even recognize their influence on Selection Arbitrariness of Satisfaction procedures.*

From the NLP angle: The realization that Natural Language recognition is NP-complete came very early on and succeeded to put Natural Language formalization efforts in the center of all computational complexities. Studying reduction proofs given in relevant papers doesn't reveal unexpected surprises, though: Even distinguishing basic word-types (like Verbs and Nouns) in an English Sentence involves, in the worst case, trying both options. One is easily reminded of *Selection Arbitrariness*.

In part B of this paper, we investigate the influence of *Dogmas* on *SAT* by studying the difference between Satisfaction in Logics and Language Recognition in Arabic more thoroughly, viewing the *SAT* efficiency puzzle as the answer to the question, whether or not there is a set of sufficient conditions imposed on the Syntax of *CNF* formulas, enabling the efficient production of non-exponentially sized *BDDs*.

In part C we show that enlightenment comes through the positive answer to a similar question related to recognition of Arabic Sentences, namely: *Is there a set of sufficient conditions imposed on subsets of Arabic Syntax, which makes efficient recognition of Sentence constituents possible?*

Using the idea of *meaning-particles*, which helps *I'rab-procedures* in NLP to be efficient, a simple pattern-oriented Satisfaction procedure is defined, which applies to *2SAT* as well as *3SAT* cases, and is seen to guarantee that *FBDDs* cannot possess sub-graphs in the form of complete binary trees for N>=3 (*2SAT*) and N>=4 (*3SAT*), respectively, where N is the number of variables in the corresponding *CNF* formula.

The formal proofs given for this fact are non-constructive in nature, but sufficient to show that known Lower Bound techniques don't apply to the here described methods, i.e., generated *FBDDs cannot* contain complete binary trees and are thus non-exponential in size. Constructive proofs, dealing with a weaker set of sufficient conditions and setting the complexity upper bound of *FBDD* sizes at $O(M^4)$, where *M* is the number of clauses in the *CNF* formula, were already given and thoroughly investigated in our three publications [10], [11] and [12], of which the latter contains also a proposed solution to the hard #2SAT counting problem. This time we also include practical testing results, performed on *3CNF* clause sets representing multiplication circuits, which clearly show non-exponential behavior of our methods.

Far from being only about a new *SAT-Solver* technique, this work investigates epistemological reasons, which stand behind *Anti-Dogmas* of Arabic.

It turns out, that Arabic represents a philosophical middle-way, in which Natural Language is not subordinated to necessities of Logics, neither in Syntax, nor in Semantics, while in the same time possessing its own *descriptive essentials*, easily amenable to formalization, covering important aspects of Symbol-Meaning relationships, to the extent, that this Philosophy may also serve as a new paradigm for computation.

## II-    Part A: Dogmas of Logics, Anti-Dogmas of Arabic

### i.   Semantic Principles of Arabic

Long ago, studies of the Arabic Language became mature sciences, very close to what is understood today to be *Modern Linguistics* [13]. Some of the marking features of those sciences were:

1- The existence of a large body of authentic references to correct Arabic Sentences, against which theories could be validated or falsified, constituting of a variety of linguistic products of Arabic speaking populations, before and after Islam. Among those authentic references, the Quran, believed to be the literal word of God expressed in plain Arabic, played the most distinguished role.

2- Theories were abstracted from data in all fields of modern relevance: *Syntax*, *Grammar*, *Phonetics*, *Semantics* as well as *Meta-Language*.

3- Language faculties were considered to be unique cognitive abilities of humans, not evolving from- nor comparable to more primitive animal communication Systems.

4- Mainstream Philosophers of Language linked Arabic to Logics in a non-subordinated way, to be described throughout this work, but some saw Logics as Language-dependent, relativizing the concept of Truth to reflect what is found to be valid through linguistic principles, not according to logical necessities of the outside world[5].

As part A of this work is mainly concerned with the relation between Syntax and Semantics in Arabic, we focus on basic Semantic principles governing this relation, as expressed in ancient as well as recent Arabic Language references (like [15] for the latter and [16] for the former), which highlight what we call here: *Anti-Dogmas*:

---

[5] The most prominent example of this Philosophy can probably be found in the chapter about causality in *Al-Ghasali's* famous *'Critic of Philosophy'*, in which he attacks the concept of *'logical necessity'* and refuses to link it to linguistic implication, claiming that implications express contingent, not necessary facts [14].

***Anti-Dogma1*:** *Words and phrases in Arabic denote meanings, not objects, nor relations between those objects. Meanings are patterns recognized by the mind either by definition, perception or abstraction[6]. They might not correspond to anything in the world or even be expressible using Symbols. Sentences denote meanings related to word combinations. Purpose of Language Rules is not determination of Truth or Falsity of Sentences or ensuring their non-contradiction, but disambiguation of their meanings (Arabic: 'Bayan').*

Noteworthy here is the comparison with De-Saussure's idea of the relation between a *signifier* and its *signified*, which links words and phrases to concepts, not necessarily standing for existing objects or their relations [17]:

> *"A sign is a combination of a 'concept' and a 'sound pattern', a union that cannot be separated in people's associative mind. It is a 'form made up of something physical, sounds, letters, gestures, etc.', which Saussure called the signifier or 'sound-image'; **to stand for an object, image, event, etc., which he termed the signified or 'concept'**"*

The common ground between *Anti-Dogma1* and Saussure's ideas is the fact, that a phrase like *'the present king of France'* can denote a *signified*, although only imagined. Allowing such *imaginary*- or *no-Object*-denotations is in direct contrast to *Dogma1*, for which the same phrase must denote nothing, because only objects or relations defined on objects are allowed.

Denotations of an important category of Arabic expressions are given in the following *essentialist* principle related to *Anti-Dogma1* (see [15]):

***Anti-Dogma2*:** *Arabic Definite and Indefinite Descriptions denote meaning categories, abstracting from the external existence of objects of a particular type to the imaginary existence of one or more representatives of the mental concept standing for this type. Degrees of explicitness of syntactic signs, used to express grammatical roles, mark degrees of definiteness of such Descriptions.*

In a Sentence like: *'Sokrat$^u$ R$^a$G$^u$L$^{un}$' = 'Sokrates is a man'*, the Indefinite Description *'R$^a$G$^u$L$^{un}$' = 'man'* denotes an *unspecified*, *imaginary representative* of the mental concept *'man'*, same goes for *'Sokrat$^u$ iNS$^a$n$^{un}$' = 'Sokrates is a human'*, *manhood* and *humanity* being the respective mental concepts, existing in the mind *per se*, without needing any further justification, as explained by Aladod [15], p128:

> *"Anything has an essence, which differentiates it from other things, whether those other things are necessarily related to it or not. Humanity, for example, is just humanity. **It is neither existent, nor non-existent, neither one nor many, nor is it attached to any contradicting properties.**"*

---

[6] Using the notion of *'cognitive pattern'* is our choice. We use it here as a working definition of what *'meaning'* is, without going into further detail, since we are only concerned with the interface between Linguistics and Logics as they stand today, not any deeper insights about Language evolution. This helps also in focusing the attention on recognition of features of surface structures rather than searching for *'deeper'* layers of meaning-representation.

In Contrast: Descriptions in Logics, as shall be seen, must either denote objects known to speakers (and are thus called: *Definite*) or just existing objects, considered: *Indefinite*.

As per *Anti-Dogma2*, Descriptions are not attached to objects at all: *Definite Descriptions denote agreed-upon-, while Indefinite Descriptions denote unspecified representatives of mental concepts.*

Note the crucial difference between the idea of a *mental concept* and what Logicians call *'Predicate*[7]: While predicates must be attached to objects (via their arguments), mental concepts are cognitive constructions standing for themselves. What is meant by *a 'representative'* of a mental concept is only*: An imaginary entity, exhibiting all properties of the given mental concept, without necessarily being attached to any object.*

Arabic treats *no-Objects* and *unspecified Objects* in two different ways. While there is no distinction between *Objects* and *no-Objects*, neither in Grammar nor in meaning, because only mental concepts and their representatives are dealt with in the first place, *Indefinite Descriptions* (modelling *unspecified Objects*) possess special grammatical Rules, which constrain their usage. This makes perfect sense, considering that the function of Language Rules in Arabic is disambiguation as per *Anti-Dogma1*. A clear manifestation of this distinctive treatment can be seen in the following:

**Principle of constructing Noun Sentences in Arabic:** *Noun Sentences are assertions, directly assigning to a Definite Description (whose role in the Sentence is named: 'Mubtada'), among other possible forms of grammatical categories, one or more Indefinite Descriptions (role name: 'Khabar'), without using any Copula. Indefinite Descriptions cannot become Mubtada, unless they are preceded by expressions affirming existence of the representatives of mental concepts denoted by them.*

In other words: Noun Sentences, which assert something about an Indefinite Description, are grammatically complete, only if they contain direct existential assertions about the indefinite entity in question (like location- or time-related information). Moreover: Indefinite Descriptions retain their positions in such Sentences, as if they were the asserted predicates.

All this indicates Arabic Grammar's sensibility to the logical distinction between Definite- and Indefinite: *The issue of existence of imaginary entities representing Indefinite Descriptions is not only raised to the level of affecting grammatical correctness of the Sentence, but even determining positions of the respective Indefinite Descriptions within Syntax.*

---

[7] As per [18]: *'A predicate is a Semantic relation that applies to one or more arguments. A one-place predicate would be "(be) green." A two-place predicate takes two arguments. For example, the two-place predicate "hit" involves both at hitter and the entity being hit. Nouns, Verbs, and Adjectives all correspond to Semantic predicates.'*

Seeing all this in a working example: The English Sentence *'A girl is beautiful.'* has no counterpart in Arabic, for instance.

Any literal translation gives only: *'$F^aT^ut^{un}$ $G^aM^iL^at^{un}$'*, which is *not* a full Noun Sentence. To complete the meaning, we need to add words conveying existence, like: *'Ga'At $F^aT^ut^{un}$ $G^aM^iL^at^{un}$.'* = *'A beautiful girl came.'*, or *'honaka $F^aT^ut^{un}$ $G^aM^iL^at^{un}$.'* = *'There is a beautiful girl.'*, or *'heya $F^aT^ut^{un}$ $G^aM^iL^at^{un}$.'* = *'She is a beautiful girl.'*, or *'Fi-lmadrasati $F^aT^ut^{un}$ $G^aM^iL^at^{un}$.'* = *'A beautiful girl is in the school.'*, etc. Note that the Indefinite Description comes in all those cases only *after* the modifiers signifying existence. According to Grammar Rules: This is the right order.

The reader might be seeing our line of argument already: *Pretending that the surface structure of Natural Language Sentences using Indefinite Descriptions does not reflect intended logical meanings, because existential assertions are missing, is an unfounded claim, easily refuted by Arabic.* More on that on the section dedicated to Russell's Indefinite Descriptions.

Most important for the current work, however, is the following principle:

***Anti-Dogma3***: *Root- and template morphemes[8] used in Verbs and Nouns have intrinsic meanings, independent of any context. They constitute, among other morphemes, 'meaning-particles'. Meaning-particles are those parts of the Syntax of a word, a phrase or a Sentence, which contain Semantic information. On the word level, they form the basis for lexical definitions. Lexica are procedures composing denotations of words from meaning-particles. Moreover: Special types of meaning-particles help classify constituent's grammatical roles in a Sentence correctly.*

A *'liar'* is called in Arabic: *'$K^aZ^iB'$*, where *'K Z B'* is the root- and *{@ a $ i %}* is the template-morpheme, filled, in order, with root characters in the place of special characters to convey the meaning. This template holds the meaning nuance: *'the one who does'*, which is a template of type *'Noun'*. We may use the same template for *'writer'* = *'$K^aT^iB'$*, where root is *'K T B'* or *'just'* = *'$A^aD^iL'$* (like in *'a just person'*). If we want to express the fact that someone is a *'frequent liar'*, another template with meaning nuance *'the one who repeats doing the verb'* is used: *{@ a $$ a %}* combined with the same root *'K Z B'* forms: *'$K^aZZ^ab'$*. Verbs are treated in the same way: The verb *'writes'* = *'$y^aKT^uB'$*, *'wrote'* = *'$K^aT^uB^a'$*, *'looks'* = *'$y^aNZ^oR'$*, *'saw'* = *'$N^aZ^aR^a'$'*, etc.

It is important to note, that using roots or templates, other than *'K Z B'* and *{@ a $ i %}*, to express the meaning of: *'the one who does the lying'* is not allowed under *any* contextual circumstance. It is not even correct to use *'$K^aZZ^ab'$* for *'$K^aZ^iB'$'*, although they are close in meaning.

---

[8] Among the possible formal representations of Arabic morphology, root-and-pattern morphology is a natural representation, as well as for other Semitic Languages. It is so widely used that this Model is also known as '*Semitic morphology*'. A (surface) *root* is a morphemic abstraction, a sequence of letters, which can only be consonants or long vowels. A pattern is a *template* of characters surrounding the slots for the root letters [19].





Symbols expressing Verbs and Nouns in Arabic have, therefore, some basic, unaltered denotations, which are *independent of any interpretation*, in direct contrast to what *Dogma3* postulates. This is also partially true for English Verbs and Nouns: You cannot call *'the one who does the lying'* anything other than: *'Liar'*, of course.

*Anti-Dogma3* shows, therefore, that Semantic Models of Logics *relativize* otherwise fixed denotations of Symbols by letting them vary arbitrarily within logical interpretations. They are *too powerful* to model Natural Language Semantics appropriately: *Accepting more Sentences than necessary, namely those, in which Symbols are assigned denotations, other than the ones given to them in ordinary Natural Language*[9].

For example, the Sentence: *'A writer is the one, who does the reading'*, is a semantically rejected Sentence in English, but an accepted one in all Models of Logics, where the denotation of the symbol *'writer'* is set to *'the one who does the reading'*. We don't expect the Sentence: *'A writer is the one, who does the writing'* to be logically valid, precisely because we allow such semantically rejected English Sentences to be true in some Models. *Logical validity of a Sentence is unrelated, in principle, to its Semantic validity in Natural Language.*

This inadequacy goes deeper in Arabic: Taking variables (and for that matter: *Any Symbols*) to be names of objects or relations is, for Logicians, a Semantic notion as we saw (recall: *Footnote 1*), i.e.: One, which has no place in the formal System and must, therefore, be *arbitrary* in nature.

However: Neither the word *'lair'* nor any Arabic Noun or Verb was *arbitrarily* selected to denote its meaning, in any *absolute* sense of the notion *arbitrary*. It may be argued, that morphemes constituting them were arbitrarily chosen, but this is of no relevance, when we are only interested in meanings of whole words.

Here again a comparison with De-Saussure's might be of value. Although his first principle states clearly (see [55]):

---

[9] The reader is reminded, that, while *'relativizing'* may seem in the context of modelling ordinary Natural Language a trivial problem, because it can be fixed for most denotations of Symbols by letting Semantic Models adhere to Natural Language denotations, as per Skolem, it touches some important set-theoretic notions used in math Languages in a very non-trivial way, which cannot be fixed. This makes those relativized notions inappropriate to capture the ordinary English meanings, which are also intended by mathematicians: *"Given any first-order axiomatization of set theory and any formula $\Omega(x)$ which is supposed to capture the notion of uncountability, the Löwenheim-Skolem theorems show that we can find a countable Model $M$ which satisfies our axioms. As in Section 1, therefore, we can find an element $m^\wedge \in M$ such that $M \vDash \Omega(m^\wedge)$ but $\{m \mid M \vDash m \in m^\wedge\}$ is only countable.* **Thus, as long as the basic set theoretic notions are characterized simply by looking at the Model theory of first-order axiomatizations of set theory, then many of these notions — and, in particular, the notions of countability and uncountability — will turn out to be unavoidably relative."** [56]





"*There is no logical basis for the choice of a particular signal to refer to a particular signification. It is not the inherent physical properties of a signal that makes it suitable for the representation of a signification (concept) and it is not the characteristics of a signification that makes it choose a particular signal to represent it. So the linguistic sign is arbitrary.*"

He acknowledges the existence of *relatively arbitrary* signs, i.e., those whose signification can be calculated from more basic, *absolutely arbitrary* components.

Formal Systems of Logics, modern or ancient, fail to correctly model Symbols reflecting *relatively arbitrary* signs, like *'liar'* or Verbs and Nouns of Arabic. The cost of this failure is computationally very high as can be seen in the following examples:

In first order Logics, if we want to express, that the meaning of the predicate *'teacher'* is *'the one who does the teaching'*, we need to include this in an explicit statement:

-   *For all x: [teacher(x) iff teaches(x)].*

And we do this for all similar Nouns:

-   *For all x: [liar(x) iff lies(x)].*

-   *For all x: [speaker(x) iff speaks(x)].*

-   *For all x: [reader(x) iff reads(x)], etc. ...*

There is *no way* of expressing the same information using one single template-rule like:

-   *For all x, <verb>: [<verb>-er(x) iff <verb>-es(x)].*

Not even in any higher order Logics, because Symbols, all the way up the quantification hierarchy, must *firstly, be taken in their entirety (i.e.: No morphological analysis is allowed to intervene in the Semantic Model structure) and secondly: Assumed not to be holding any intrinsic, interpretation-independent information (Dogma3).*

In Arabic, the above rule is embedded in the root-/template-origin of Verbs and Nouns, without any need for explicit logical statements: The one who performs the verb *'writes' = 'y$^a$KT$^u$B'* must be called *'writer' = 'K$^a$T$^i$B'*, per canonical definition of Arabic word-templates.

Generally speaking: Template-morphemes used in Arabic are meta-Symbols, holding most of the meaning nuances we need to be aware of, when calculating the overall meaning of a word or a Sentence or when deducing information. Ignoring this fact leads to considerable computational losses.

For example: A *'pathological liar'* is a *'liar'*, who repeats his ill-doings more than once. Suppose we have a first order premise stating that: *'Every liar is hated by someone'* and a fact: *'The man is a pathological liar'*, expressed as:





- *For all x: There is y: [liar(x) implies hates(y,x)].*

- *pathological_liar('the man').*

Since in Logics we do not analyze the morphological structure of predicate Symbols, we cannot deduce: *'There is someone, who hates the man'*, unless we add an explicit rule:

- *For all x: [pathological_liar(x) implies liar(x)]*

In Arabic: *'pathological liar' = 'K$^a$ZZ$^a$b'* contrasts *'liar' = 'K$^a$Z$^i$B'*, but because of the intrinsic relation between template meta-Symbols: *{@ a $ i %}* and *{@ a $$ a %}*, we don't need to explicitly list implications relating individual predicates to each other. We have just to include one meta-symbolic rule of the form:

- *For all x and all root-Symbols '@', '$', '%': [{@ a $$ a %}(x) implies {@ a $ i %}(x)]*

Which states something like: *'All entities exhibiting frequent behavior, exhibit also normal behavior'*.

Using such a rule: *'the man is a pathological liar' = 'alR$^a$G$^u$L$^u$ K$^a$ZZ$^a$b$^{un}$'*, can help us deduce: *'the man is a liar' = 'alR$^a$G$^u$L$^u$ K$^a$Z$^i$b$^{un}$'* and then apply the first order premise to find out, that someone must hate him.

In Arabic: All Verb- and Noun-templates are linked to such meta-symbolic Rules expressing either unique- or shared meaning nuances[10]. This property, which is a *trueness*, was called by IbnDjinni [16]: *'the small deduction property' = 'Al-ishtikak al-azghar'.*

What about roots of Verbs and Nouns?

IbnDjinni conjectures that permutations of characters used in roots are also linked to meaning nuances (he calls this *'the big deduction property' = 'Al-ishtikak al-alakbar'*):

All permutations of characters *'M L K'* produce roots related to the concept of *'power'*, for example. *'M L K'* is root for words like: *'king', 'owner'*, etc..., *'L K M'* root for *'punch', 'hit'*, etc., *'K M L'* root for *'complete', 'perfect'*, etc. However: Since most of the root-permutations are not activated by

---

[10] Importance of meta-symbolism is underlined in a Quote from [20]: *"The human being is the being of meta-levels. It can transcend beyond different meta-levels of Language by higher-level symbolization and interpretation. If we refer to the transgressing of sets within one and the same level by the Latin syllable 'trans' and to the ascending to higher-levels by 'super' or 'supra', we can call the human being the trans-interpreting and/or super- or suprainterpreting being, or for short: the meta-interpreting being, the level-transgressing interpreting being and by that in turn really the (meta)reflecting being. Abstract reflections are only possible if you can transcend the actual level at hand, if you can transgress the levels by going meta-symbolic or super-interpreting. Therefore, it is most plausible to notify the human being as the "animal metasymbolicum" (a sort of extension of Cassirer's terminology) or as the super-interpreting being (in extension of Nietzsche's conception of the interpreting being)."*





Language speakers, the legitimacy of using free meaning-nuances is questionable[11], if commonly used Language-conventions are to be strictly followed.

There are problems related to computing valid logical deductions, even if we disregard morphology and adopt the Logician's idea that Language-related axioms can be properly modelled by listing them as clauses or Rules in logical programs (like listing the fact that *'a bachelor'* is *'an unmarried man'*, for example, so that both terms can be substituted for each other in first order clauses).

Take the following two quantified Sentences:

- *Every newborn is beautiful.*

- *Every beautiful female is classy.*

It is *not* possible, even via Aristotelian syllogisms, to produce:

- *Every newborn female is classy.*

Because the middle-term is not the same in the first two Sentences [21]. In first order Logics, the above Sentences could be translated in this way:

- *For all x: [newborn(x) implies beautiful(x)].*

- *For all x: [(beautiful(x) and female(x)) implies classy(x)].*

And here also the conclusion:

- *For all x: [(newborn(x) and female(x)) implies classy(x)]*

Cannot directly follow, in spite of the fact, that all instances of the two Rules produce the conclusion. Fixing this problem involves either using inductive- or second order reasoning. Both solutions result in intractable computational difficulties.

An Arabic-enabled formal System, in which data and meta-data levels are treated in the same way, without necessarily dropping level-distinction[12], can easily recognize, through its lexicon, not through morphology, that *'newborn'* = *'m$^a$WLouD'* (root: *'W L D'*), is a neutral adjective used by both genders, so that a deduction taking this information into account may directly and efficiently be performed.

Summarizing the findings of this section we can say that:

1- *Dogmas* of Logicians cause formal Systems adopting them to *over- and under-accept* Natural Language Sentences, because on the one side they relativize otherwise fixed Symbol denotations in Semantic Models and on the other: Reject *no-Object*-denotations, *in principle*.

---

[11] In ancient texts like [16], this question was referred to as: *'the legitimacy of using analogy within Language'*-question.

[12] Remember that separating data- and meta-data levels is a logical principle, caused by logical fallacies, not Natural Language Rules.





2-     Such formal Systems fail, *in principle also*, in analyzing Symbol morphology, an important source of meaning in Arabic.

3-     The previous two points must lead either to substantial computational difficulties or loss of expressive power, if those Systems are used to model Semantics of Arabic Sentences.

4-     In comparison: Adhering to Arabic Language *Anti-Dogmas* enables not only a solid theoretical foundation for linguistic Models, but also overcoming processing shortcomings through extensive use of *meaning-particles*, which help in disambiguating denotations of words, phrases and Sentences, as shall be seen in next sections.

### ii.   What is essential in Naming and why?

In modern, western Language Philosophy there are many objections to essentialist approaches to meaning in Language as endorsed by Aristotle. Those objections are articulated notably by Quine, who regarded *essentialism* as: *'A relic from metaphysical dogmas, which a pure empiricist has to overcome'*.

As per [57]: Quine's objections relate to his rejection of the traditional distinction between analytic and synthetic Truth. There is no longer any *necessary Truth*, but only some *'web of beliefs'* which can be more coherent or less coherent. This is the reason why Quine claimed that the search for an object's essential properties would be in vain. For him, the external world is accessible via Descriptions only: *If there is necessity in logic at all, it must be de dicto, since this type of necessity is reducible to a semantic predicate*.

Quine's idea of a *pure descriptive necessity* is best illustrated by his example of a cycling mathematician, respectively, a mathematical cyclist [57]:

*"Mathematicians may conceivably be said to be necessarily rational and not necessarily two-legged; and cyclists necessarily two-legged and not necessarily rational. But what of an individual who counts among his eccentricities both mathematics and cycling? Is this concrete individual necessarily rational and contingently two-legged or vice versa? Just insofar as we are talking referentially of the object, with no special bias toward a background grouping of mathematicians as against cyclists or vice versa,* **there is no semblance of sense in rating some of his attributes as necessary and others as contingent"**

There are also notorious objections to Quine's objections, formulated by Kripke in [58], which amount to showing that Quine's approach (also called: *'anti-essentialist empiricism'*) does require for proper names a *'descriptive account of reference'*, like the one suggested by Frege and Russell[13]. In [58]

---

[13] As per [59]: Russell regarded a proper name to refer not to a referent, but to a set of true propositions that uniquely describe a referent, for example: *'Aristotle'* refers to *'The teacher of Alexander the Great'*. The common-sense view of reference was originally formulated by John Stuart Mill, when he defined it as *'a word that answers the purpose of showing what thing it is that we are talking about but not of telling anything about it'*. This view was criticized by Frege, who pointed out that proper names may apply to imaginary and nonexistent entities, without becoming meaningless. Rejecting descriptivism, Kripke held that names come to be associated





Kripke regards such an account as counter-intuitive as seen from the following quote:

> "Suppose that someone said, pointing to Nixon, 'That's the guy who might have lost'. Someone else says 'Oh no, if you describe him as 'Nixon', then he might have lost; but, of course, describing him as the winner, then it is not true that he might have lost'. Now which one is being the philosopher, here, the unintuitive man? **It seems to me obviously to be the second."**

As per [57]: A similar reasoning applies to Quine's cycling mathematician: If he is described as a *'mathematician'*, he is necessarily rational. If he is described as a *'cyclist'*, he is necessarily two-legged. But it would be absurd to answer the question whether a certain person, for example Smith, is necessarily two-legged in the following way: *'If you describe Smith as a cyclist, he is necessarily two-legged. But if you describe him as a mathematician, then he is not.'* Therefore, Quine's anti-essentialist understanding of modality must be rejected, because it presupposes a counter-intuitive way of Language use.

In Summary: Kripke tried to overcome an anti-essentialist Dogma in Philosophy, which was established by Kant and which heavily influenced Quine and other early analytic philosophers, like Russell and Wittgenstein He did this by recurring to linguistic convention, which, when naming things, gives *'logical essence'* to referents.

What about Arabic? How can we understand the *essentialism* expressed in the above *Anti-Dogmas*, i.e., letting Symbols refer to *mental entities*, not requiring manifestations in objects (*Anti-Dogma2*)? Or letting Symbols denote meaning-nuances, independent of interpretations (*Anti-Dogma3*)? Is this notion of necessity *descriptive* or *logical*?

To answer this question correctly, we need to quote [60], in which conventional Philosophy of Arabic and counter-positions of ancient Language philosophers, like Ibn Taymiyyah, are discussed, while in the same time drawing an interesting comparison to similar western counter-opinions, notably of Wittgenstein:

> "According to conventional Arabic Language Theory, the origins of Language lay in an act of conventional assignation (Arabic: 'wad'). Although there were various opinions on how words themselves came to have meaning – by virtue of this meaning being inherent to them, as argued by the Muʿtazilite theologian Al-Ṣaymarī, by a revelatory act of God, or by some combination of divine fiat and subsequent human convention, as explained by the Ashʿarī theologian and jurisprudent Al-Shīrāzī – the majority of philosophers and theologians held that Language was originally established through convention, whereby certain utterances and words were assigned to signify certain objects. This conventional account of the origin of Language, in

---

with individual referents, because social groups link the name to its reference in a naming event, which henceforth fixes the value of the name to the specific referent within that community. Another alternative is the *'direct reference theory'*, which holds that proper names refer to their referents (objects) without attributing any additional information, connotative or of sense, about them (in strict compliance to *Dogma1*).





*turn, explained the dichotomy between veridical and metaphorical expressions. For according to most forms of the conventional theory, the original assignation of words* **through a process of ostensive definition**[14] *was followed by subsequent instances of usage, in which words came to denote objects other than those originally assigned to them. This, according to the conventional theory, was how metaphorical utterances and expressions came into existence [...]*

*Ibn Taymiyyah's criticism of the categorical syllogism and the definition also form the overall context in which he attacks the conventional theory of Language and meaning, according to which ostensive definitions were the method by which particular utterances are first used to signify particular objects, or particular apprehensions regarding objects. Contra the philosophers and theologians, whom Ibn Taymiyyah faults for regarding definitions as a form of certain knowledge,* **Ibn Taymiyyah argues that knowledge is entirely possible without definitions because a person who coins a definition must already know the object he is defining before he defines it.** *This same criticism reappears in Ibn Taymiyyah's critique of the conventional theory of Language and its account of the origins of Language.*

*The conventional theory held, as we saw, that meaning in Language can only come about through convention, which then justifies subsequent use. Ibn Taymiyyah raises two objections to this claim. First, he says, there is no evidence that those who speak a Language ever came together to coin all the expressions used to denote objects in that Language and to assign particular expressions to particular objects. Second, and here Ibn Taymiyyah's criticisms of the conventional theory of Language tie into his critique of the epistemic tools and resources of Aristotelian thought and its privileging of the definition as a source of certain knowledge:* **Ibn Taymiyyah and his student Ibn al-Qayyim argue that conventional agreement on the meaning of words cannot arise prior to the use of a Language in ways that already convey meaning.** *Ibn Taymiyyah is, of course, reacting against what he perceives to be the error of the Muslim philosophers and theologians who, in his view, mistakenly regard philosophical definitions of conceptual terms*[15] *as a more certain source of knowledge than scripture and therefore call for scripture to be interpreted in light of those definitions. But his rejection of the views of the philosophers and theologians is based on his argument that the foundations on which these groups build their views on Language are rationally and philosophically unsound.*

---

[14] An ostensive definition conveys the meaning of a term by pointing out examples.

[15] *i.e.:* The adoption of: *The descriptive account of reference.*





Wittgenstein proves to be an interesting and perhaps indispensable philosopher to read alongside Ibn Taymiyyah here because in his later period he too famously attacked the theory of ostensive definition, associated in Western linguistic philosophy with the figure of Augustine. In place of the Augustinian view he rejects, **Wittgenstein advances an idea that is already present in Ibn Taymiyyah, namely that the meaning of Language arises out of use.**

As with Ibn Taymiyyah, Wittgenstein's attack on Augustine's conception of Language and meaning is based on Wittgenstein's rejection of Aristotelianism, **particularly the notion that a definition is essential to knowing a concept.** "When I give the description 'The ground was quite covered with plants', do you want to say that I don't know what I'm talking about until I can give a definition of a plant?" Wittgenstein asks rhetorically. Like Ibn Taymiyyah, who argued that a definition could only be coined by those who already knew the object being defined and that **meaning could not arise out of convention because every conventional act of definition already assumes certain ways of using Language in ways that already convey meaning**, Wittgenstein also insists that in order to define an object one already has to know something about the object and about using Language in ways that convey meaning: "So, one could say: an ostensive definition explains the use – the meaning – of a word if the role the word is supposed to play in the Language is already clear… . one has already to know (or be able to do) something before one can ask what something is called." […]

Wittgenstein's attacks on conventional theories of linguistic meaning also overlapped with his attack on the other mainstay or Aristotelianism: **the idea of universals, which Wittgenstein believed was unnecessarily privileged by philosophers in his own time.** Wittgenstein was critical of philosophers who exhibited what he called a "contemptuous attitude towards the particular case" while privileging universal categories as the source of meaning – criticisms that, as we have seen, also occupy a central place in Ibn Taymiyyah's criticisms of the conventional theory of Language. Wittgenstein also explicitly connects the lure of universals with the emergence of incorrect theories of meaning. The "craving for generality", Wittgenstein argues, leads to the philosophical and epistemological error of trying to "look for something in common to all the entities which we commonly subsume under a general term" and keeps us from analogizing between specific objects and seeing interconnectedness of things. **It is a mistake, Wittgenstein insists, to think of the meaning of a word as an image or a thing correlated to a word, as when we think that the person who understands the term "leaf" has come to possess a kind of general picture of a leaf, as opposed to pictures of particular leaves.**

Wittgenstein also regarded categorical syllogisms, the mainstay of the Aristotelian method, as tautological and senseless objects. Just as Ibn Taymiyyah argued that the initial premises of a syllogism already contained within them the conclusion that was inferred in the final proposition, Wittgenstein also argues that to say that one proposition follows from another is to say that the first proposition already says everything said by the inferred proposition. In other words, the first proposition already presents a proper picture of reality, unlike a syllogism, which according to Wittgenstein is not a picture of reality and is therefore nonsensical – a conclusion that follows from Wittgenstein's thesis that **a proposition with sense must be a picture or Model of reality if it is to have any sense; it cannot say what things are, only how they are.**





As with Wittgenstein, Ibn Taymiyyah's critique of Aristotelianism also goes hand in hand with his criticisms of the conventional linguistic philosophy prevalent in his day. Like Wittgenstein for whom syllogisms yield little or no knowledge, Ibn Taymiyyah argues that the definitions and categorical syllogisms championed by the philosophers do not constitute viable sources of knowledge of the sort that could be regarded as superior to scriptural proofs. **Knowledge, Ibn Taymiyyah insists, arises from the consideration of specific cases and not the extra mental universals that are utilized in categorical syllogisms. In fact, one's knowledge of universals and even of basic rational truths is, according to Ibn Taymiyyah, itself dependent on analogizing what is sensed and experienced to what is not. Thus, Ibn Taymiyyah says, the basis of knowledge is not the categorical syllogism but the sort of analogies used in Islamic juristic reasoning.** Foreshadowing Wittgenstein's criticisms of his contemporaries, Ibn Taymiyyah exclaims that Muslim philosophers, theologians and jurists have fallen in thrall to the spell of Hellenistic philosophy, from which they need to be awakened.

It is clear, then, that Ibn Taymiyyah and Wittgenstein's theories of Language are tied to their overall assault on Aristotelianism, particularly the idea that meaning in Language is created by ostensive definition. **For both Ibn Taymiyyah and Wittgenstein, the Aristotelian tradition is mistaken in emphasizing the importance of knowing the definition and meaning of individual words and what they signify in themselves, by reference to classificatory criteria such as genus and differentia.** Against this view, both Ibn Taymiyyah and Wittgenstein reject the notion that knowledge of objects depends upon definitions **and insist that meaning arises out of use**. Thus in his *Philosophical Investigations*, in which Wittgenstein attacks both what he regards as the illusion that words must correspond to images of real things and also the theory of Language associated with Augustine – whom Wittgenstein charges with adopting a pictorial conception of the relationship between objects and words – **Wittgenstein explicitly states that the meaning of Language is determined by use and does not arise prior to it.** In most cases, Wittgenstein says, the word "meaning" can be explained in this way: **"the meaning of a word is its use in the Language."** Elsewhere, Wittgenstein expresses the same idea thus: "But if we had to name anything which is the life of the sign, we should have to say that it was its use."

As mentioned before (see *Footnote 6*): In this work we concentrate on the Phenomenon of Natural Language as it exists today, abstracting away from the otherwise important aspect of its *Origin*. This helps focusing our attention on its relation to Logics, without losing generality. It also helps in filtering away some philosophical positions, which do not affect our investigations.

To facilitate an overview of relevant opinions of above philosophers, as we understand them, the following Table summarizes our findings:





| | Symbols denote | Logical Essence | name-meanings | Truth via | N. Language Grammar |
|---|---|---|---|---|---|
| Frege / Russell | Objects | no | via descriptions | universals | inappropriate |
| Quine | Objects | no | via descriptions | universals | inappropriate |
| Kripke | Objects | yes | also directly | universals | inappropriate |
| Wittgenstein / IbnTaymia | Objects | no | via use | analogy | Rules for words-use |
| Chomsky | cognitive patterns | irrelevant | problematic | structure | loosely connected |
| Conventional Arabic (*Anti-Dogmas*) | cognitive patterns | irrelevant | via meta-Symbols | irrelevant | has meaning particles |

Essence, meaning and Truth through Language – T1

This table shows, in the first three rows, the effect of seeing Natural Language as a subordinate System Serving Logics: Objects are considered to be the correct denotations of Symbols and universal generalizations about properties of those objects the right way to find Truth. Natural Language Grammar is regarded as inappropriate to adequately express meanings of Sentences, because, for Logicians, those meanings are compositions of references to objects, which Grammar is not really concerned with.

Since our arguments against the servitude of Natural Language to Logics are articulated throughout the rest of this paper, we focus in this section only on discussing ideas related to: *Naming.* For Logicians: *Naming* is the faculty of assigning to specific Symbols specific references, whether those references are objects or relations between those objects.

That *Naming* of objects is only possible through assigning them Descriptions, is refuted by adopting Wittgenstein's or Ibn Taymiyyah's above circularity argument (especially because *Origin* is of no concern to us): Meaning of a Symbol cannot arise out of a conventional act of definition, because *'defining'* assumes using Language in ways that already convey meanings of the defined (i.e.: *We don't name definitions*).

Kripke's opinion regarding proper names cannot count as truly different from Logicians views, to the contrary: He re-introduces *logical essence*, postulating that *Naming* creates an entity, which *necessarily* exists in all possible worlds.

Using his own example, there is a fairly simple objection. Even if everybody in all possible worlds necessarily knows, who Nixon is: Are his attributes also necessarily agreed upon? In other words, in the alleged, social *Naming* event, in which reference to *'Nixon'* was fixed: Did convention fix also all his attributes? Difficult to imagine that Nixon was *'good'* or *'bad'* or *'tall'* or *'small'* for everybody. How can we account for the relativity of attributes of an object, if its reference is fixed? And if we say, not all attributes are fixed, some of them may be relative, then we fall again into Quine's remark: *"There is no semblance of sense in rating some of his attributes as necessary and others as contingent."*





This objection is easily extendible to concepts: What is the nature of the concept we call: *'time'*? Is it logically necessary, i.e. a *trueness* in the outside world[16]? If it is necessary, then its attributes must also be *essential*, but then: How do we account for the fact that time is different for different observers? If it is contingent, then we must admit timeless places and/or entities, which are beyond rationality and science[17].

While Wittgenstein and Ibn Taymiyyah distance themselves from *essentialism*, universal generalizations and even logical deduction as a means of finding *Truth*, postulating that analogy is the only way to knowledge and meanings of Symbols are not understood by convention, but by use, they also fail to see any important role of Grammar, other than being a set of mere empirical *'Rules of conduct'* for Language speakers. They both don't seem to have *any* notion of necessity, neither in Description nor in Logics[18].

There are obviously two sides to the act of giving the name: *'N'* to an external object *O* or an imagined concept *C,* anyone of which possessing a set of attributes *A*:

The object- or concept side, i.e., the question, *why attributes A of O or C are eligible to be called: 'N'?* And the opposite side: *Why is Symbol 'N' appropriate to be selected, out of the arsenal of all Language Symbols, for denoting O or C, knowing that they have attributes A?*

While the first question relates to Logics, the second one relates clearly to Language. In their discussions about necessity and contingency, above Philosophers seem to be talking only about the first question.

Substantial difference to Logicians comes with Chomsky, whose opinions, mentioned in the fifth row of Table T1, we describe one by one in the following points, quoting directly from [61]:

---

[16] Newtonian time is *necessary* in the sense that time itself is distinct from our measures of time. But Newton also conceived of time as necessary in a more profound way, namely, that it exists independently of any physical objects whatsoever. In contrast, the concept of time for Kant is based upon the idea, that the human has an external and an internal sense and time can be only recognized by an internal sense. Time is nothing else than the form of the internal sense, that is, of the intuitions of self of our internal state. Interesting to mention here the position of Islamic epistemology, backed up by Quran itself, which clearly asserts that time is *not* necessary or absolute, but relative: *"And they urge you to hasten the punishment. But Allah will never fail in His promise. And indeed, a day with your Lord is like a thousand years of those which you count."* [Quran: 47, 22].

[17] Modern epistemology refuses of course to fall into ontological discussions, since it succeeded in separating *ontological Truth* from our Model of it. Kripke's revival of logical essentialism seems, however, not to follow this wise way.

[18] We are not familiar enough with the work of both Philosophers, though, so that our opinion might not be accurate in this respect. A deeper investigation of their philosophical work is beyond the scope of this paper.





1- Reference: *"Chomsky has interpreted the notion of reference used in Binding Theory as a syntactic notion, which could be **conceived as a relation between expressions and internal entities of some (unspecified) cognitive domain.** Chomsky assumes, in other words, that the notion of reference can be interpreted only as an internal notion, and every theory of reference which supports a hypothetical relation between words and external objects must be rejected (so, both the Fregean theory and the Direct Reference Theory)."*

2- Essence: *"Chomsky explicitly points out that **it would be perverse** to try to find a relation between the internal (mental) entities and the external objects in the world."*

3- Naming: "*According to Chomsky, to postulate a relation between words and the world is not scientific; is a 'kind of neo-scholastic picture'. Chomsky devotes many reflections to the criticism against the notion of reference. I quote one:*

> *'A good part of contemporary philosophy of Language is concerned with analyzing alleged relations between expressions and things, often exploring our intuitions about the technical notions "denote", "refer to", "true of", etc. said to hold between expressions and something else. But there can be no intuitions about these notions, just as there can be none about "angular velocity" or "protein". [...] it is not at all clear that the theory of natural Language and its use involves relations of "denotation", "true of", etc. in anything like the sense of the technical theory of meaning.'*

> *Chomsky's thought can be summarized this way: **The notion of reference is not a common sense or a scientific notion**; it is a pseudo-scientific notion, which cannot enter in the theoretical apparatus of the naturalistic analysis."*

4- & 5- Truth and Grammar: *"In fact, the logical form (which should be written as "Logical Form", to point out that we are dealing with a technical notion of generative Grammar) used by Chomsky **describes only those semantic aspects which are determined by the syntactic structure of the Sentence (like quantifier raising, scope relations, etc.), ignoring many other Semantic contents** which are usually expressed by the logical form used in the philosophical tradition, since Frege to Williamson."*

Returning to *Anti-Dogmas* and to the last row in the above table we notice that, while Chomsky's ideas are closest, i.e.: They adopt the same understanding of reference proposed in *Anti-Dogma1*, which refuses to see objects as meanings of Symbols and is, therefore, uninterested in their *logical essence*, there are important differences to Chomsky:





a- During *Naming* actions, he puts more emphasis on the will of speakers, rather than on Natural Language Rules, as can be seen in the following quote from [61]:

> "The externalized notion of reference, instead, can be part of the analysis of artificial Languages. Chomsky thinks that the notion of Bedeutung is adequate for the scientific Languages. In these Languages, in fact, we can construct symbolic objects which 'may well aim towards the Fregean ideal', since they have 'a semantics, based on the technical notion of Bedeutung, a relation between symbols and things'. This is not possible, instead, in natural Language, where:

>> 'There are complex conditions – poorly understood, though illustrative examples are not hard to find – that an entity must satisfy to qualify as a "naturally nameable" thing: these conditions include spatiotemporally contiguity, Gestalt qualities, functions within the space of human actions [...] A collection of leaves on a tree, for example, is not a nameable thing, but it would fall within this category if a new art form of "leaf arrangement" was devised and some artist had composed the collection of leaves as a work of art. He could then stipulate that his creation was to be named serenity. **Thus it seems that there is an essential reference even to willful acts, in determining what is a nameable thing.**'

Contrarily: *Anti-Dogma3* postulates that Symbols representing Verbs and Nouns used by speakers *must* possess intrinsic meanings, calculated from meta-Symbols (root- and template-morphemes). *Such morphemes represent descriptive essentials/necessities, which cannot be overruled by willful actions*[19].

Using Chomsky's own example: Even if we admit, that a work of Art can, literally, be called any name we wish (which doesn't happen in the reality of serious artistic disciplines), it is not the case, that names or attributes we give to people, for example, are also arbitrary: *To express that someone didn't speak the Truth we call him: 'Liar', not anything else.*

Is there any binding reason, why we must use *'liar'*, not *'teaching'*, for example, to refer to the one, who didn't tell the Truth? There is of course the fact, that he must be an Actor of the verb *'to lie'* and there is a *descriptive necessity*, coming from convention, prescribing

---

[19] *Naming* is highlighted in Islamic epistemology, backed up by Quran, as an innate and unique faculty of humans, which was directly taught to them by Allah: *"And He taught Adam all names (of everything), then He showed them to the angles and said: 'Tell me the names of those, if you are truthful'."* [Quran: 31, 2]. The fact that, in contrast to the description of the same story in the Old Testament, Quran explicitly mentions: *'All names'*, led many ancient as well as contemporary Muslim scholars to regard this verse as referring to *the faculty of Naming anything, whether object or concept, Verb, Noun or proposition.* This clearly contrasts the *purely referentialist* Semantic account, attributed to Augustin, according to which every word is a referential expression, whose meaning is its extension, i.e.: *The idea, that Names of objects alone are the essence of Natural Language.*





for Actors to be named using an End-Symbol containing the *'-er'* morpheme. The reason why we don't use: *'teach-er'* is similarly obvious: *'teach'* is fixed by convention to denote some other, unrelated meaning.

In Summary: While respecting the idea of Chomsky, that *Naming* has a component related to the free will, *our free will is always constrained by Rules and necessities of Language.*

b- *Descriptive necessities*, which are part of the broader concept of *meaning-particles* (a concept representing meta-Symbolic Semantic properties of words, phrases and Sentences used in Grammar), are not taken into account in Chomsky's views of Grammar. However, as seen in below sections: *The influence of meaning-particles on Arabic Grammar is uncontestable.*

c- While *AntiDogma1* emphasizes the fact, that purpose of Language Rules is not determination of *Truth* or *Falsity* of Sentences or even ensuring their non-contradiction, but only disambiguation of their meanings, Chomsky has an *internal* concept of *Truth*, which is not *material*, but *structural* or *formal* and as per [61] this concept is externalized by factors, not amenable to scientific investigation:

> *"he thinks in fact that denotation and reference are determined by many factors (beliefs, desires, etc.) which interact with the faculty of Language and determine how speakers use the formal expressions generated by the Grammar in their speech acts.* **However, no scientific theory can consider and study all these factors together, since no one seeks to study everything"***.*

Coming back to the question posed in the beginning of this section, concerning the nature of the *essentialism* used in the here described convention-based *Anti-Dogmas* and taking into account the empirical nature of convention, which is a psychological act, unrelated to facts or objects of the outside world and relevant only to participants:

*AntiDogmas deal only with the question, why a Symbol is appropriate to be selected from the arsenal of Language-Symbols for denoting specific objects or concepts. Because conventions require imaginary ideas for people to agree upon and sounds and signs to materialize this agreement, AntiDogmas only prescribe Language-dependent agreement necessities (descriptive necessities).*

We can understand the Philosophy of Arabic in the following obvious way: *Whatever is real in the outside world, epistemologically it can only be accessed through Language, which is built upon conventions requiring imaginary ideas like: 'concept', 'entity', etc... as well as empirical principles governing sounds and signs to express those ideas in words and combine them in phrases and Sentences* (cf. [62]). In contrast to the non-increasable adjective real, *reality* only applies to ontology outside any language (cf. [63]).





In this way we can understand the role of *mental concepts* as referents of Descriptions in *Anti-Dogma2*, as opposed to objects or predicates related to objects (*Dogma1* and *Dogma2*): *Since imaginary concepts are necessary for convention, they become essentials to be reflected in all meanings of words, phrases and Sentences.*

We can also explain *Anti-Dogma3*: *Root- and Template-morphemes, being cognitive patterns, existent empirically as Symbols in Syntax, are the only way to express Nouns and Verbs and as such: Necessary for calculating their meanings.*

Arabic regards, therefore, *'the imaginary'* or *'the cognitively existent'* as *'the essential'*: Essential for convention, not for Truth.

### iii.  Meaning-particles and their use

The best illustration of the use of *meaning-particles* on the word-level of Arabic is given by IbnMalik in the first verses (from 10-14) of his famous poem [22], summarizing Rules of distinguishing Nouns from Verbs, among other signs, in the following:

***Noun distinction criteria****: The following syntactic occurrences in a Sentence are sufficient to recognize related words as Nouns: The vowel mark 'Garr', the vowel mark 'Tanween' (both at the end of a word-string), the word used in a vocative case, the leading string: 'Al-'.*

For example, in the Sentence: *'Ali is in the school' = 'Aaliy$^{on}$ fi al-madrasa$^{ti'}$* both words *'Aaliy$^{on}$'* and *'al-madrasa$^{ti'}$* can directly be recognized as Nouns, because *'on'* is a vowel mark of type: *'Tanween'* and the substring *'al-'* leads the word *'madrasa$^{ti'}$*.

Vowel marks, which are usually placed as small marks on top of the words, as in all previous examples, are termed '*ḥarakāt'* and included in the broader term: *'tashkil'*, which represents *supplementary diacritics*.

The literal meaning of *'tashkil'* is '*forming'*. As the normal, every-day Arabic text does not provide enough information about the correct pronunciation, one of the purposes of *tashkīl* (and *ḥarakāt*) is to provide a phonetic guide, i.e., show the correct pronunciation [23].

On the Sentence level, Grammar and Semantics of Arabic are linked in the following obvious way:

***Role of Grammar:*** *If a Sentence S is semantically correct, then it is grammatically correct as well, but not vice versa[20].*

The most important purpose of *ḥarakāt* on the Sentence level is the disambiguation of roles of words and phrases in the grammatical structure of a

---

[20] The Sentence*: 'A reader is the one who writes' = 'Alqari$^u$ howa zak$^a$ al-lazy y$^u$kt$^o$b$^{u'}$* is obviously correct, grammatically, but wrong, semantically, because it confuses the meanings of *'reading'* and *'writing'*.





Sentence, a function, which can be expressed in the following important principle:

***Necessity of ḥarakāt:*** *If an unmarked Arabic Sentence S is grammatically correct, then there exists at least one correct list of ḥarakāt, which can be used to mark all words and phrases within S. Marking is done by putting appropriate ḥarakāt at the end of all words, if possible. If a word ends with the special character 'Alif-Layyena', then its marking cannot be explicitly put, but only estimated[21]. Because phrases are abstract entities: All ḥarakāt put on phrases of a Sentence are estimated.*

We call an Arabic Grammar *'Diacritic-Grammar'*, if ḥarakāt are included in its terminal Symbols and used in its production Rules [24].

Let *G* be such a Grammar[22]:

G = (V$_N$, V$_T$, P, S)

*V$_N$ = {S, NS, VS, Noun, Verb, Definite, Indefinite, Genitive, Al-Noun-Pattern}*

*V$_T$ = {All words formed using an Arabic lexicon and marked with vowels}*

*P = {*

> ➢ ***1- (S > NS | VS),***
> ➢ ***2- (NS > Assertion Definite),***
> ➢ ***3- (VS > Verb | Noun Verb),***
> ➢ ***4- (Noun > Definite | Indefinite),***
> ➢ ***5- (Definite > Al-Noun-Pattern | Genitive ),***
> ➢ ***6- (Indefinite > Noun)***
> ➢ ***7- (Genitive > $^i$Definite Indefinite | $^{in}$Indefinite Indefinite)***
> ➢ ***8- (Al-Noun-Pattern > Noun$^{Al}$)***
> ➢ ***9- (Noun > {… all words of type 'Noun' formed using an Arabic lexicon …})***
> ➢ ***10- (Verb > {… all words of type 'Verb' formed using an Arabic lexicon …})***
> ➢ ***11- (Assertion > Definite | Indefinite | VS)***

>   *}*

*S = Starting Symbol*

Recognition procedures implementing *G*, also called: *'I'rab-procedures'*, follow the following steps:

---

[21] For example, in the Sentence: *'Isa hit Musa' = 'D$^a$R$^a$B$^a$ Isa> Musa>'*, vowel markings: *'u'* and *'a'* should be put on the subject *Isa* and the object *Musa*, respectively, but because of their ending with *'Alif-Layyena'* (depicted here by 'a>'), this cannot be done and the marking is only *estimated*.

[22] Right parts of production are processed from *right to left*.





1- Since templates of the lexicon are marked *'Noun'* or *'Verb'*: Use those markings to decide in a first scan, what types of words occur in the Sentence to be recognized (productions: *8-, 9- and 10-*).

2- If there is still ambiguity on the word level: Let *Noun distinction criteria*, embedded in productions of *G*, guide your derivation by prioritizing productions with matching *ḥarakāt*.

An example shows the added value of using *ḥarakāt* to guide derivations:

Let *S* be an unmarked Sentence:

*'the boy's opinion is the best opinion' = 'RAY al-WaLaD KHaYR RAY'.*

We may use *I'rab* (productions: *5-, 6-, 8-* ) to find, after a first search in the lexicon, that the Non-Terminals used in the Sentence are[23]:

$$S' = <\textbf{\textit{Indefinite Indefinite Definite Indefinite}} >.$$

But to continue processing, the two productions: *4-, 11-* create additional 16 possibilities, resulting from the fact, that a *'Definite'* or an *'Indefinite'* can both be either *'Noun'* or *'Assertion'*.

Consider now the Sentence *S*, fully marked with *ḥarakāt*:

$$S'' = \textit{'RAY}^u \textit{ al-WaLaD}^i \textit{ KHaYR}^u \textit{ RAY}^{in}\textit{'}.$$

Following the mark *'in'* in the production Rules, we get production *7-*, which tells us after one step, that *S''* has the form:

$$\textit{'RAY}^u \textit{ al-WaLaD}^i \textit{ <Genitive>'}.$$

We can use the same production in a next step (via. The mark *'i'*) to get:

$$\textit{'<Genitive> <Genitive>'},$$

which is, using *5-*, of the form: *'<Definite> <Definite>'*. At this stage *4- and 11-* come into play again, but this time only 4 possibilities are checked, before concluding that *S* has the form: *'<Assertion> <Noun>'*, which is *'<NS>'*.

This example shows that: To classify *S* without using *ḥarakāt* is in most cases exponentially more expensive, than when *ḥarakāt* guide derivations.

Nevertheless: *The question of whether there exists a Diacritic-Grammar, used by an I'rab-procedure, which recognizes any Arabic Sentence using a number of steps not exponential in N, the number of words in the Sentence, is still an open, uninvestigated question.*

---

[23] Because of the use of Latin characters for Arabic ones: Read Grammar templates from *right* to *left* and the corresponding Arabic Sentences written in Latin characters from *left* to *right*.





### iv.  Sense and Reference

In his famous article entitled *'On Sense and Reference'* [25], Frege wrote the following:

*"**Equality gives rise to challenging questions which are not altogether easy to answer**. Is it a relation? A relation between objects, or between names or signs of objects? In my Begriffsschrift I assumed the latter. The reasons which seem to favor this are the following: a = a and a = b are obviously statements of differing cognitive value; a = a holds a priori and, according to Kant, is to be labelled analytic, while statements of the form  a = b often contain very valuable extensions of our knowledge and cannot always be established a priori. [......]*

*What is intended to be said by a = b seems to be that the signs or names 'a' and 'b' designate the same thing, so that those signs themselves would be under discussion; a relation between them would be asserted. **But this relation would hold between the names or signs only in so far as they named or designated something**. **It would be mediated by the connexion of each of the two signs with the same designated thing**. [......]*

*To every expression belonging to a complete totality of signs, there should certainly correspond a definite sense; **but natural Languages often do not satisfy this condition, and one must be content if the same word has the same sense in the same context**. [......]*

*The words 'the celestial body most distant from the Earth' have a sense, but it is very doubtful if they also have a reference. The expression 'the least rapidly convergent series' has a sense but demonstrably has no reference, since for every given convergent series, another convergent, but less rapidly convergent, series can be found. **In grasping a sense, one is not certainly assured of a reference**. [......]*

*Footnote: In the case of an actual proper name such as 'Aristotle' opinions as to the sense may differ. It might, for instance, be taken to be the following: 'the pupil of Plato and teacher of Alexander the Great'. Anybody who does this will attach another sense to the Sentence 'Aristotle was born in Stagira' than will a man who takes as the sense of the name: 'the  teacher of Alexander the Great who was born in Stagira'. **So long as the reference remains  the same, such variations of sense may be tolerated**, although they are to be avoided in  the theoretical structure of a demonstrative science and ought not to occur in a perfect Language."*

Those passages contain Frege's essential assumptions about Natural Language, reflected in the *Dogmas* quite clearly:

1- Referents of words and phrases are objects or relations between those objects. Referents of Sentences are the values: *'True'* or *'False'*, considered to be objects as well (this is *Dogma1*, which we shall call henceforth also: *'Referential Doctrine'*)

2- Noun Sentences using *'ToBe'* constructions in English are *identity statements*, their analysis is a matter of understanding what *Equality* means.

3- While *identity statements* of the form *'a=a'* do not add any new information, those of the form *'a=b'* are cognitively interesting, assuming that *a*, *b* refer to the same object.

4- *'Sense'* is some condition an object must satisfy in order to count as a referent.





5- It is also the mode of presentation of the referent and can, in a perfect Language, only exist, if the referent exists.

6- In Natural Languages: Definite Descriptions express senses, which might not have referents.

7- Natural Languages do not require an expression to have one definite sense.

8- As long as referents remain the same, variations of senses may be tolerated.

Frege's ideas gave rise to what has been called afterwards: *'Language Puzzles'*, whose essential types were:

a- *Identity Puzzles*: How do we account for the difference between the Sentences: *'The Morning Star is the Morning Star'* and *'The Morning Star is the Evening Star'*, if *'The Morning Star'* and *'The Evening Star'* both refer to the planet Venus?

b- *Substitutivity Puzzles*: How do we explain the fact that *'Frege thinks that Venus is the Morning Star'* might be *true*, while *'Frege thinks that Venus is the Evening Star'* might be *false*, if Frege is not aware of the relation identifying morning- with evening stars? Such Sentences, representing aspirations, form what is called: *An opaque context*.

c- *No-Object Puzzles*: What is the denotation/Truth value of a Sentence like: *'The present king of France is bald'*?

d- *Negative Existentials Puzzles*: What is the denotation/Truth value of a Sentence like: *'The present king of France does not exist'*?

Before discussing Frege's attempted solutions, we notice the following:

1- The interplay between the *Referential Doctrine*, understanding *ToBe*-constructions as mathematical identity statements and Leibniz's substitution law is the primary cause of all the above puzzles.

2- There seems to be a contradiction going beyond Natural Language between viewing senses as modes of presentation of referents on the one hand (point 5-), i.e., letting them only exist, when referents exist, and on the other: Allowing Definite Descriptions to express senses referring to nothing (point 6-). Frege points out that the expression *'the least rapidly convergent series'* has a sense, but demonstrably no reference. However: This expression is used in mathematics, not in ordinary Natural Language. One is inclined to ask, then: Does Frege's *'perfect Language'* require senses to always have references or not? Mathematics, for example, cannot depend only on expressions possessing references, since otherwise many precisely defined and very useful, but imaginary math objects will cease to exist.

3- Ambiguity is seen by Frege to be a negative property of Natural Language, not expected to occur in his *'perfect'* Language. His opinion was formed in a time, when such misconceptions about formal Languages were still believed to be true. Later, when results





of Goedel, Tarski, Skolem, and others, were finally accepted, ambiguity became an indispensable property for all types of Languages and the price to be paid for expressive power. We shall see below how similar misconceptions about properties of Natural Languages were the basis for Russell's beliefs as well.

*Identity puzzles* are claimed to be solved in Frege's opinion, because any Sentence not only has a referent, but a sense as well, so that the distinction between *'a=a'* and *'a=b',* for any *a, b* referring to the same thing, can be attributed to difference in sense.

However: What do we really mean by the *'Equality of the two expressions a, b'*? Are we equalizing their senses or their references? How do we know that *'a is b'* does not stand for: *'The sense of the morning star is the sense of the evening star'*?

Frege's clear choice is equalizing references, i.e., the identity relation, expressed using the verb *'ToBe'* in the above Sentences, reflects for him: *'Referential Identity'* (above quote):

> "What is intended to be said by *a = b* seems to be that the signs or names 'a' and 'b' designate the same thing"

What about *'Conceptual Identity'*? Obviously: Since the identity relation, which the verb *'ToBe'* is supposed to represent, is borrowed from mathematics, *Conceptual Identity* cannot be meant.

Can we restrain our understanding of identity to *Referential Identity*, without causing serious damage in Natural Language contexts?

The following anomaly provides a negative answer to this question:

$S_1$=*'Colonel Dr. Sam Daniels __is__ the fictional character representing a genius medical doctor who fights a deadly virus in the movie Outbreak'*

$S_2$=*'Raymond Babbitt __is__ the fictional character representing a sensitive autistic older brother who can barely speak two phrases without stuttering in the movie Rain man'*

$S_3$=*'Dustin Hoffman __is__ one of the key actors of New Hollywood, known for his versatile portrayals of antiheroes and emotionally vulnerable characters.'*

$S_4$= *'Colonel Dr. Sam Daniels __is__ Dustin Hoffman'*

$S_5$= *'Raymond Babbitt __is__ Dustin Hoffman'*

Since *'is'* in all above Sentences represents *Referential Identity*, the following absurd Sentences are deducible via Leibniz's substitution law:

$S_6$=*'Colonel Dr. Sam Daniels __is__ Raymond Babbitt'*

In other words:

$S_7$=*'The fictional character representing a genius medical doctor who fights a deadly virus in the movie Outbreak __is__ the fictional character representing a sensitive autistic older brother who can barely speak two phrases without stuttering in the movie Rain man '*





Note that $S_4$ and $S_5$, are enforced upon us by both the *Referential Doctrine* and the claim, that *'ToBe'* expresses identity.

This anomaly (called here: *Actor Anomaly*) contradicts Frege's assertion, that fixing referents makes variations of senses tolerable (point 8-).

It unveils a problem we already had with Frege's own example: While *'The Morning Star is the Evening Star'* is *referentially true*, it is *conceptually false*: *'Morning Stars'* and *'Evening Stars'* express conceptually different things. Frege handles this problem only in opaque contexts, where he lets references of expressions be senses, not objects, claiming that differences between senses are relevant only within aspirations.

The general principle governing this phenomenon in Natural Language, whose bypassing leads to the anomaly, is the following:

***Priority of meanings[24]:*** *No two Natural Language expressions, intending to describe different meanings, are semantically identical, even when they refer to one and the same object.*

Applying this principle to $S_7$ lets us reinterpret the intension of the verb: *'is'* in that Sentence as follows:

$S_7'$ = *'The fictional character representing a genius medical doctor who fights a deadly virus in the movie Outbreak **is played by the same Actor as the one playing** the fictional character representing a sensitive autistic older brother who can barely speak two phrases without stuttering in the movie Rain man'*

Similarly, what we mean to say by: *'The Morning Star is the Evening Star'* is obviously: *'The morning star **is physically the same object** as the evening star'.*

Both formulations are not identity statements, neither in English, nor in mathematics. They are simply: *Predicative assertions*.

If we drop the *Referential Doctrine* altogether and try to fix this situation, by letting Definite Descriptions denote senses, like in opaque contexts, restricting identity assertions to $S_1$, $S_2$ and $S_3$ only, then we are faced with the problem: How do we express the fact, that both fictional characters have one real person in common?

In English this is easily done using the Sentences:

$S_8$= *'Colonel Dr. Sam Daniels **is played by** Dustin Hoffman'*

$S_9$= *'Raymond Babbitt **is played by** Dustin Hoffman'*

Here again: The verb *'ToBe'* is not expressing identity, but affirming a relation between a *Subject* and its *Predicate*.

---

[24] We use *'meanings'* instead of *'senses'* here, so that we can refer to this principle in the context of Arabic *Anti-Dogmas* as well.





### v.   The verb *'ToBe'*

Tracing back the history of the verb *'ToBe'*, [26] distinguishes three uses of it, the first one reflecting *Referential Identity* as seen in the last section and understood by Frege, Russell and many other Logicians, the second supporting *affirmation* or *negation*, as per Aristotle:

> *"Let's summarize. In deciphering the scaffolding of the affirmative Sentence, we've distinguished two fundamental pillars (the subject and the predicate) and another additional component (tense) usually expressed syncretically on the predicate, when the predicate is a verb, or by an autonomous verb, the verb to be, when the predicate is expressed not by a verb but, for example, by a noun. **In every case, the verb to be is not a predicate; it's "only" a verb.** In this scenario, the verb to be appears, in a certain sense, as secondary compared to the two fundamental pillars: the subject and the predicate. **This is not surprising, given that the defining property of the Sentence is seen as the possibility of affirming whether a certain sequence of words expresses a Truth or not,** and considering that the possibility of expressing a Truth derives, in the final analysis, from the attribution (or negation of attribution) of a predicate to a subject and not from tense. Aristotle is certainly explicit about the secondary nature of the verb to be—indeed, on closer inspection a "tertiary" nature, since it comes after the subject and the predicate: "For example, in a man is just, I say that 'is' is the third component" (De int., 10, 19b, 20–22)."*

The third is its use as a *'Copula'*, i.e., a way to couple parts of Sentences together to form Truths:

> *"So, in the Port-Royal Grammar, the copula is to be found at the intersection of the trajectories that make up a complete vision of the human mind: **it is seen as the catalyst, the linchpin (sometimes, for brevity's sake, hidden in a verb) around which nothing other than the structure of human judgment turns and takes on substance.** Therefore, we have concluded the second stage of the history of the verb to be through the centuries: alongside the vision of the verb to be as a support of tense, we place that of the verb to be as the name of affirmation and as the "copula," the element of creative fusion between two independent concepts embodied in words that result in a judgment. We must also take these currents into account in order to understand the importance of copular Sentence analysis in the twentieth century, but a new controversy is lurking; **it insinuates itself into linguistics through the misgivings of mathematicians and logicians.**"*

The last Sentences of this quote refer to the use of *'ToBe'* as the *'Name of Identity'* (as the author calls it), promoted mainly by Russell in influential remarks like:

> *"It is a disgrace to the human race[25] that it has chosen to employ the same word 'is' for two entirely different ideas. [....] A disgrace which a symbolic logical Language of course remedies".*

---

[25] While it might have been normal in the first half of the last century to accept such hilarious generalizations, linking the *'human race'* to Grammar constructions used in one particular family of Languages, as if this family represents anything other than its own cultural context, this remark of Russell surely stands out today for the type of ignorance and arrogance influencing many of his false views about Language.





Which is strongly criticized in [26], without hiding a sense of surprise that such a view could be adopted by a prominent thinking figure like Russell:

*"First a preliminary and general note: **it's clear that Russell is not expressing the thought of a linguist.** To say that Socrates is human expresses a relationship of predication while Socrates is a man expresses an identity means shuffling the cards around on the table, in the sense that the relationship of identity is always mediated in natural Language by a relationship of predication, since we're still dealing with a Sentence. Even if one were to use an explicit predicate of identity, such as to be identical to, one would still have a relationship of predication, even in the presence of an explicit expression of an identity relationship; therefore, **identity and predication are not antagonists from a linguistic point of view**. But the question is much more complex and important. Russell's distance from linguistics becomes even more evident when he speaks about the verb to be followed by an adjective. **Had he adhered to the linguistic tradition up to that time, and of which an intellectual of his caliber must certainly have been aware, he would have had to say either that 'is' is the manifestation of time and the true predicate is the adjective human** (had he been an Aristotelian), **or that 'is' is the manifestation of affirmation and that the predicate is nevertheless human** (had he been a Port-Royalist). Evidently, Russell was neither an Aristotelian nor a Port-Royalist, and in this case, he deviates radically from tradition by asserting that the verb to be is a predicate [….] In a certain way, **he builds with his own two hands the very ambiguity against which he immediately and vehemently lashes out**. This idea that the verb to be is a predicate of identity was moreover not one that Russell invented from scratch: the credit goes to his illustrious predecessor, **also not a linguist**, the great logician Gottlob Frege, who with the famous example 'The evening star is the morning star' intended to create an indisputable case in which the verb to be produces an identity."*

Before going over to Arabic, we briefly list arguments held in [26] against Russell's idea, that Sentences like *'Sokrates is a man'* are identity statements:





a- If Russell had used an example with a female name, for instance, that of Socrates's wife, *Xanthippe*, the Italian clitic form[26] of the Sentence: *'Santippe è una donna'* = *'Xanthippe is a woman'* would have been: *'Santippe lo è'* = *'Xanthippe lo [so] is'* and not: *'Santippe la è'* = *'Xanthippe la [she] is'*, which demonstrates that *'una donna'* = *'a woman'* **functions as a predicate**. Obviously, there's no reason to think that *'un uomo'* = *'a man'* functions any differently.

b- If we say: *'Mary is her admirer'*, *'her'* can't refer to *'Mary'*; with any other verb, things change. If instead we say *'Mary knows her admirer'*, *'her'* can continue to refer to someone else, but this time it can also just as well refer to *'Mary'*. This special property of the verb *'ToBe'* has been the subject of many studies, and it seems to be invariable in all Languages, which use *'ToBe'*-constructions. It has been described this way: **If a Noun Phrase, has a so-called 'referential capacity', where by 'referential' it is meant the opposite of predication, then a pronoun that is contained in that phrase (such as 'her' in the previous examples) can refer to the subject, otherwise not.**

c- Hence: If in Frege's examples we replace the phrases: *'The evening star'* and *'The morning star'*, with *'The evening star* and *'its associate in the firmament'*, then: $S_1$ = *'The Evening Star <u>is</u> its associate in the firmament'* and $S_2$ = *'The Evening Star <u>is identical</u> to its associate in the firmament'* cause any native English speaker to recognize, that the pronoun *'its'* refers in $S_2$ only to *'The Evening Star'*, making *'its associate'* in $S_1$ definitely a predicate, i.e., **$S_1$ cannot be an identity statement.**

d- [26] concludes:

> *"This, of course, doesn't mean that we can't call the proposition in question an 'identity proposition', but it implies that the role of the verb 'ToBe' needs to be well defined and that, if anything, **identity needs to be explained independently** of it —starting, for example, from the interpretation and from the structure of the noun phrase involved in the Sentence."*

Where does Arabic, or for that matter: All Semitic Languages, not using *'ToBe'*-constructions to link *Subjects* to *Predicates*, stand in this discussion?

---

[26] As per [26]: *"In Italian and in many other Roman Languages, nouns may be replaced by pronouns of a type called 'clitic' —which are supported by other words because they are unstressed—and in most cases they agree in gender and number with the nouns they refer to."*





We already saw in section (i) of this part, that Noun Sentences are constructed without the use of Copula. Translated into Logicians views: *'Sokrat$^u$ R$^a$G$^u$L$^{un}$'* = *'Socrates is a man'* can only be a predicative assertion, having the following logical form (in first order Logics): *Man(Socrates)*. Looking back, we can easily see then, that the *Actor Anomaly* vanishes, since from:

$S_4$= *DustinHoffman(ColonelDr.SamDaniels),*

$S_5$=*DustinHoffman(Raymond Babbitt)*

One cannot, by any means, deduce:

$S_6$= *RaymondBabbitt (ColonelDr.SamDaniels)*

Summarizing our findings related to *Identity*- and *Substitutivity Puzzles* we can say then:

- They are caused by the *Referential Doctrine*, imposed upon Language by Logicians, who understood *'ToBe'*-constructions to semantically reflect *Referential Identity*.

- This understanding causes serious meaning anomalies, requiring for their solution the ability to respect *the Priority of meanings* principle, a condition, which is not fulfilled in Frege-type Systems.

- It is also an understanding, which is not based upon *any* linguistic principles, whatsoever. To the contrary:

  o *'ToBe'*-constructions in Indo-European Languages are mostly *predicative*, a fact, which is shown to be true for all Sentences marked as problematic by Logicians.

  o Important Language families (like Semitic-Languages) don't use Copula to link *Subjects* with *Predicates* in the first place. Alleged anomalies raised by Logicians and referring to Copula are thus inapplicable to such Languages, *in principle*.

  o This last fact sheds a strong shadow of doubt on either: The validity of Logicians ideas concerning Sense and Reference in general or the validity of the same for Indo-European Languages in particular or both.





### vi.  Russell's Indefinite Descriptions

Knowing that Noun Sentences in Arabic cannot represent Identity- and must, therefore, correspond to predicative statements in English, lets us get another clue in favor of the conjecture, that an Indefinite Description like: $'R^aG^uL^{un}' = 'a\ man'$ is used in a non-referential way in all Languages (at least: Not referential in the sense understood by Frege and Russell).

This goes against the *Referential Doctrine*, of course, and Russell needed to come up with an idea, which allowed Indefinite Description also to fit into this doctrine.

We shall elaborate first on the marking Semantic features of Indefinite Descriptions in Arabic (mentioned briefly in section (i), *AntiDogma2*), their usage in Grammar and their roles as quantifiers, before going over to Russell's ideas. In Arabic:

1- Indefinite Descriptions denote *unspecified imaginary entities representing mental concepts, not objects.* When they come in singular form, they refer to one- and when they come in plural form, they refer to a collection of such entities.
2- They are best represented in Grammar through *Common Nouns,* which explains the lack of specification and their usage in supporting quantifiers. *Common Nouns are words for types of things. None of those things are intended to be singled out with respect to the properties included in the given mental concept.*
3- They are mostly used in Noun Sentences as *Khabar* (the Sentence part corresponding to a *Predicate* in English).
4- Their less common use as *Mubtada* (the part of a Noun Sentence corresponding to a *Subject* in English) is bound to existential assertions, without which such Sentences are grammatically incomplete.
5- When they are used as *Mubtada*, they retain the position of *Khabar*, coming always *after* existential assertions.
6- Only singular forms are permitted with quantifiers. For example, the English Sentence: *'All mammals are creatures'* has no counterpart in Arabic. Either we say: $'K^oll^u\ TH^aDY^{yen}\ K^aIN^{un}'$ = *'Every mammal is a creature'* or $'K^oll^u\ alTH^aDY^{y}at^i\ K^aINat^{un}'$ = *'All **the** mammals are creatures'*, i.e., use the definite form.
7- They may be used with *All*-quantifiers only. When this happens, the intended meaning becomes distributive, i.e., applied to every single element of the collection. For example: $'K^oll^u\ INS^an^{in}\ F^aN^{in}'$ = *'Every human is mortal'* means there is no single human, who will live forever. Note the difference to Sentences like: $'K^oll^u\ alR^iG^aL^i\ Y^oH^iBBun^a\ aNN^iSa^{a'}$ = *'All (the) men love women'*, where there is ambiguity between understanding: *'Every single man loves every single woman'* or *'The collective of all men loves women'*.





Believing the *Referential Doctrine* to be a *trueness*, Russell's misunderstanding of Natural Language was developed further into his famous thesis about the use of Indefinite- and Definite Descriptions in Natural Language.

In [26] the author explains the reason why Russell went to such length to prohibit *'a man'* in *'Socrates is a man'* from being understood as a predicate:

> *"... when I say Socrates is a man I must be careful not to admit that a man—which of course belongs to the same type as Socrates, which is a noun phrase—can be a predicate of Socrates. In other words, I must at all costs avoid a noun's (or a noun phrase's) being the predicate of another noun (or noun phrase); this would reintroduce the antinomy[27] and we'd be back where we started. **So, for Russell the only alternative to keep from demolishing type theory was to admit that the verb to be was necessarily a predicate of identity when followed by a noun (or noun phrase), or that the two nouns (or noun phrases) that precede and follow the verb to be are both referential—which is to say neither of the two is a predicate—exactly as it is with any other transitive verb, that is, those involving two nouns or two noun phrases.** Since, however, a Sentence like Socrates is human displays no conflict between elements of the same type—because human is an adjective and therefore doesn't belong to the same type as Socrates—there is no longer any need (or possibility, since adjectives can't be referential) to admit that the verb to be is an identity predicate"*

We hope the reader agrees, that type-theory, being useful in formal Systems of Logics, is of no concern to Natural Language, *which does not even care about the fundamental, logical law of non-contradiction.* This makes Russell's concerns about type-theory completely pointless in Natural Language contexts.

In [27] a detailed discussion of Russell's ideas may be found, as well as critical opinions of his contemporary scholars (like Strawson).

We focus here on the following passages, related to his attempted solutions of *No-Object-* and *Negative Existentials* Puzzles:

> *"Russell's theory of Description entails an answer to the question how a Sentence such as:*
>
> *(5) The present king of France is bald.*
>
> *Can be meaningful even when France is not presently a monarchy. The difficulty, from Russell's perspective, was that this Sentence appears to be a Sentence in* **subject-predicate form** *and as such it appears to attribute a certain property to a certain object. However, that cannot, he reasoned, be the true logical form of the Sentence. There is currently no King of France.* **If we grant Russell's underlying assumption that all there can be to the meaning of a bona fide referring term is its role of standing for a certain object, then it follows that (5) is entirely meaningless.** *Since our Sentence is evidently meaningful, it follows that 'the present King of France' cannot really be a term and so cannot really serve as the logical subject of (5)."*

---

[27] Here: The Barber antinomy.





And further:

*"The key to solving the foregoing puzzles, according to Russell, is to see that definite Descriptions are what he calls* **incomplete Symbols**. **If a is an incomplete symbol, then a has no meaning in isolation, but every Sentence in which it occurs does have a meaning.** *This way of stating matters may mislead, however. The point is not that incomplete Symbols lack meaning in isolation, but have meaning in context. Nor is it that such expressions are devoid of meaning in the way that, say, nonsense is.* **The point is rather that contrary to appearances, incomplete Symbols are not proper grammatical constituents of the Sentences in which they occur.** *For example, our initial assessment might be that the expression 'the round square' is a bona fide referring term and as such occupies subject position in: 'The round square does not exist'. But Russell holds that 'the round square' is not a referring term at all and is* **not really a proper grammatical subject.** *[......]*

*The puzzles arise only because we are* **misled by surface grammatical form and are insufficiently attentive to logical form.** *Consider the following relatively (though less so than Russell imagined) straightforward example:*

*(7) Smith met a man.*

*Taking appearances as a guide, 'a man' in (7) appears to occupy direct object position exactly on a par with 'Jones' in 'Smith met Jones.' And one might be tempted to say that just as 'Jones' refers to the object asserted by 'Smith met Jones' to have been met by Smith, so 'a man' refers to the object asserted by 'Smith met a man' to have been met by Smith.* **But what object is that?**

*Russell himself makes an odd claim, though just in passing, in this connection.* **He says that 'a man' in 'Smith met a man' denotes an arbitrary man.** *That, I think, is an unfortunate way of phrasing a correct and essential point - unfortunate because this way of phrasing matters can make it sound as if there is an arbitrary man that Smith met. Of course, one cannot meet an arbitrary man. If one meets a man, one meets a particular man. One meets Jones or Black or Brown or someone (or more) of the men that there are in the universe. Notice though, and this I think is what Russell was really driving at, that there is no particular man such that 'Smith met a man' is true only if Smith met that very man. 'Smith met a man' thus stands in sharp contrast with 'Smith met Jones'.* **If Smith met Brown but not Jones then 'Smith met Jones' is false but 'Smith met a man' is true. And the same goes for any particular man you care to name. So there is no particular man x such that 'Smith met a man' entails that 'Smith met x'.**

*A Meinongian might be tempted to conclude that therefore 'Smith met a man' expresses a relation between Smith and a new kind of object,* **an arbitrary object***, and that 'a man' refers to such an object.* **The way to extinguish that temptation, Russell claims, is to see that the appearance that the 'a man' occupies object position and that 'Smith met a man' expresses a relation between Smith and some object is illusory.** *The Russellian logical form of the Sentence 'Smith met a man' is closer to that of an existentially quantified Sentence of the form:*

*(8) (There exists: x)(man x and Smith met x).*





> *Notice that no single constituent of (8) directly corresponds to 'a man' in 'Smith met a man.' Where we have 'a man' in (7) we have in (8) the existential quantifier, the predicate 'man' and two occurrences of the variable - none of which, taken either individually or in various combinations, is a constituent of (8) which directly paraphrases or translates 'a man' as it occurs in (7). That is why Russell concludes that 'a man' in (7) functions as an 'incomplete symbol' which, in effect, disappears under analysis."*

Our objections to the above ideas can be resumed in the following general points, preceding particular objections from the perspective of Arabic:

1- Enforcing the *Referential Doctrine* on Natural Language made grammatical constituents like *Subject* and *Object*, in Russell's eyes, *necessarily representing objects*, in direct contradiction to a basic rule of Natural Language, namely: *Grammar roles are independent of the meaning of expressions occupied by them.*

   In a Sentence like: *'Her courage is enormous.'* the *Subject: 'her courage'* cannot refer to an object, even from a logical point of view. There isn't any way around the *trueness*, that English Grammar allows us to directly attribute a property to another property. And even agreeing to what Logicians require, i.e., accepting that the surface structure of this Sentence is not revealing its true meaning (so that we need to convert it to an existential formula of second order) doesn't help understand how English Language works: A *'hidden meaning'* neither concerns English Grammar nor questions definition, purpose or organization of its constituents. To the contrary: *It underlines the fact, that roles of Sentence constituents, as determined by that Grammar, are independent of whatever Logicians think those Sentences or parts of them are really expressing.*

2- Realizing that Definite Descriptions like: *'The present king of France'* don't refer to anything in the current state of the world, Russell jumped to the conclusion that Definite- as well as Indefinite Descriptions are *incomplete Symbols*, which need to be eliminated, because they:

   > "…. *are not proper grammatical constituents of the Sentences in which they occur"*

   Here again one must ask: Which Grammar is he referring to? *Evidently: English Grammar shall always admit non-referring Descriptions as proper constituents, whether this is acceptable to Logicians or not.*

   Moreover: An important logical objection to the idea of *incomplete Symbols* is the following: If in the current status of the world there is no referent of the expression: *'The present king of France'*, what happens, when/if this status changes and France becomes a monarchy again? Is the status of an *incomplete symbol* a permanent one or are we allowing some *incomplete Symbols* to become sometimes *'complete'*, i.e., full constituents of Sentences? Russell





accepted the fact, that *trueness* is relative to a set of axioms: Was he willing also to accept the same for *reference*? It doesn't seem to have been the case. Note that when we use possible worlds Semantics to reason about knowledge, for example, we can always find a possible world, in which France is a monarchy. *In fact: Possible worlds Semantics allows us to find references in some obscure world to any Definite Description, however absurd such a Description may seem, making the incomplete-Symbols-invention pointless.*

3- Eliminating Grammar constituents, which are deemed *incomplete*, changes the meaning. No surprise then that:

> *'… no single constituent of (8) directly corresponds to 'a man' in 'Smith met a man.'*

Both Sentences don't express the same, actually: *'There exists a man and Smith met this man'* **is not** saying: *'Smith met a man'.* We are not talking in the latter about the existence of an unspecified man, but about an action performed by a known one.

In ordinary Language, the Sentence *'Smith met a man.'* has one of the following meanings:

a- *Smith met a man*, whose name is known to the speaker, but not relevant for the context of the conversation, because it is about Smith's action.

b- *Smith met a man*, whose name the speaker does not know.

c- *Smith met an ordinary man*, who possesses all properties of the predicate *'man'* as understood by all speakers), including the property of having a name (what Russell probably meant, when he called him *'arbitrary'*).

> Obviously: All three interpretations falsify the claim that:

> *"… there is no particular man x such that 'Smith met a man' entails that 'Smith met x' …"*

Someone knowing names of all men encountered by Smith (let us say he met three: 'Henry', 'Ali' and 'Mustafa') can easily find such a particular man (by mere enumeration). *For him: 'Smith met a man' necessarily entails exactly one of the Sentences: 'Smith met Henry' or 'Smith met Ali' or 'Smith met Mustafa'.*

The same can actually be shown using proper methods of Logics: Applying Skolemization[28] to (8) gives us the equisatisfiable Sentence Set:

$$S = \{'man\ c', 'Smith\ met\ c'\}$$

for a suitable constant c, where the following holds: *[(8) is true **iff** S is true].*

---

[28] As per [28]: *"Skolemization is a way of removing existential quantifiers from a formula. Variables bound by existential quantifiers which are not inside the scope of universal quantifiers can simply be replaced by constants."*





This clarifies an important point: Even adapting Russell's existential translation, the meaning of Indefinite Description: *'a man'* does *not* contradict finding a particular man fulfilling the Sentence, contrary to what Russell claimed.

4- In his attempted solutions of *No-Object*- and *Negative Existential* Puzzles, Russell seems to apply the following argument, in which the cascading effect of his ideas about Natural Language can be observed:

From:

  a. Definite Descriptions necessarily stand for objects (*Referential Doctrine* and *Dogma2*),

  b. Definite Description: *'The present king of France'* does not stand for any object in the current state of the world,

  c. In this Sentence, the Grammar constituent: *Subject* is: *'The present king of France'*,

  d. The Sentence: *'The present king of France is bald.'* is not meaningless,

Deduce:

  e. There are *incomplete Symbols*, i.e., ones, which have meanings only in a context, not in isolation,

  f. *All* Definite- and Indefinite Descriptions in Natural Language are *incomplete Symbols*,

  g. *'The present king of France'* is a Definite Description and cannot really be a term and so cannot really serve as the *'logical subject'*,

  h. It cannot be a *grammatical Subject* as well, because *incomplete Symbols* cannot form Sentence constituents,

  i. Which means that we are misled by the surface grammatical form of this Sentence, *incomplete Symbols* must be eliminated to reveal the *'hidden'* logical meaning.

Taking *b.*, *c.* and *d.* alone, one could have either doubted in the correctness of *a.* or admitted that *b.* is a relative assertion, not enabling jumping to any of the conclusions (e-i). Moreover: Since conclusions f. and h. are linguistic generalizations, one would have expected them to be supported by linguistic evidence as well.

In Reality: Jumping to unsupported linguistic conclusions seemed to be a routine exercise, practiced by Russell, Frege and their followers, in a time, where Natural Language was thought to be *'defect'* and in need of *'repair'*.

All the above has already been said, in a way or another, and Russell's ideas about Natural Language, thoroughly criticized by many prominent western Linguists, foremost Chomsky.





What is the position of Arabic?

Remembering that *AntiDogmas* presented in this work are principles of Arabic, uncontested among Linguists throughout the time, we realize the following:

1- The *Referential Doctrine* is clearly contradicted by *AntiDogma1*. As seen above in the Description of Russell's cascading argument: This fact is sufficient alone to falsify his theories about Definite- and Indefinite Descriptions. Moreover: Our findings in the last sections showed, that this doctrine has a serious side-effect: Anomalies, which cannot be removed unless *the Priority of meanings* principle is respected (*Actor Anomaly*), which is not possible in any view of Natural Language Semantics, seeing reference to objects as the sole purpose of Descriptions.

2- Russell's *incomplete Symbols* idea directly contradicts *AntiDogma3*, which states that: *'Verbs and Nouns have intrinsic meanings, independent of any context'.* This includes, of course, Definite- and Indefinite Descriptions.

One can argue, that Russell's meaning-notion is referential in a very *material sense* and therefore different, in principle, from the meaning-notion presented in *AntiDogma3*, which prioritizes *the imaginary*. But even admitting this, we still have the following serious issues:

   i-    The fact itself, that there exists such a different meaning-notion, supported by overwhelming linguistic evidence, cannot count in favor of Russell's claim, that Descriptions have *'no meaning in isolation'.* To the contrary: It is another indication that his meaning theory is missing essential parts (we remember that Frege introduced *'senses'* to account for different meaning notions, which Russell refused to do).

   ii-   The notion supported by *AntiDogma3* places meaning at Symbol-, not at word- or expression levels. According to this: *Incomplete Symbols*, allegedly present in Arabic also, must exhibit some meaning, even if they have no reference and even when they are dissolved in existential expressions. How is this meaning accounted for?

         Take for example: *'Smith met a frequent liar'.* The *incomplete symbol*: *'a frequent liar'*, which represents the grammatical *Object* in this Sentence turns, according to Russell, into a single predicate when we translate the Sentence to:

              *(There exists: x)(frequent liar x and Smith met x).*

         But this is not the only thing we are saying in the original Sentence. We say also that: *'Smith met a liar'.* In plain English: *'a frequent liar'* is *'a liar'*, of course, but in





Russell's Logics they are two different predicates, semantically unrelated to each other. For expressing our intention, something like this Sentence is needed:

*(There exists: x)(frequent liar x and liar x and Smith met x).*

Which raises the question: How many different predicates must be mentioned explicitly to capture our intention correctly, when we translate Sentence constituents into Logics in this way? And more importantly: Where do we get this vital Semantic information from? Where, in Russell's System, are linguistic Rules of the form:

*(For all: x) (frequent liar x implies liar x)*?

As per Russell: The Logics counterpart of the Arabic Sentence: *'Smith iLtᵃQa KᵃZZᵃBᵃⁿ'* would be:

*(There exists: x)(KᵃZZᵃB x and Smith iLtᵃQa x).*

As already mentioned in section (i): The meaning nuance *'the one who repeatedly performs the verb'* is stored in the morpheme: *{@ a $$ a %}*, intrinsically linked to: *{@ a $ i %}* via a meta-symbolic rule of the form:

> *For all x and all root-Symbols '@', '$', '%': [{@ a $$ a %}(x) implies {@ a $ i %}(x)]*

Since Russell's System does not foresee any meta-symbolic analysis (like the one needed to form and use this rule): The *incomplete symbol*: *'KᵃZZᵃB'* retains a meaning which can never be truly reflected.

3- In section (i) (*Principle of constructing Noun Sentences in Arabic*), we have seen that, in Arabic, Noun Sentences forming assertions about Indefinite Descriptions are only complete, when they start with existential expressions. This means that, contrary to what Russell claims: *Arabic Grammar is reflecting existence issues, similar to those, which are thought to be hidden behind surface structures in English.* More on this point in the next section.

4- We remember that, as per *AntiDogma2*, Indefinite Descriptions denote in Arabic an unspecified imaginary entity, representing the mental concept on hand (here: *'KᵃZⁱB'* = *'lying'*). This entity is also the grammatical *Object* of the Sentence and, as already seen, we don't care, whether it exists or doesn't exist, because it is not an *'object'* in the Logicians sense. To contemplate the importance of separating *the conceptual-* from *the referential* meaning of Indefinite Descriptions, consider the following semantically correct Sentence:

*S = 'I demanded that you read a book, but not a particular one.'*

Suppose we have only three books: *b1, b2, b3*. The second part of *S* is logically equivalent to:





*((I didn't demand that you read b1) And*

*(I didn't demand that you read b2) And*

*(I didn't demand that you read b3))*

According to Russell, the first part of S should translate into:

*'There exists book x: I demanded that you read x'*,

which means that *S* contains a contradiction, in contrast to what any normal Language speaker would be willing to admit. We shall, henceforth, call this anomaly: *'Inadequacy of existential translations'*.

### vii. The relation between Grammar and Meaning in Arabic Sentences

Is Grammar independent of meaning? In [29] Chomsky states:

> *"Grammar is best formulated as a self-contained study independent of Semantics. In particular, the notion of grammaticalness cannot be identified with meaningfulness"*

What is stated here is the well-known *Principle of the Autonomy of Syntax*, which Chomsky defends by criticizing the equivalence of meaningfulness and grammaticality in both directions.

For Chomsky: Neither the Sentence: '*Colorless green ideas sleep furiously*' has a meaning, nor seems there to be any Semantic reason why, for example, the interrogation in: '*Have you a book on music*?' forms a valid English Sentence, while '*Read you a book on music*?' doesn't or *'The book seems interesting'* is acceptable and *'The child seems sleeping'* is not[29].

According to [30]: The main objective of Chomsky's Principle is to grant a definition of Syntax, which is non-dependent on Semantics. His intent is to subtract Syntax and Grammar from the *'intuitive nature'* of meaning. Intuition must have for him the same space and role in Linguistics as it has in other sciences: i.e., guiding the first insights, to be then replaced by purely formal analyses.

Chomsky attacks the traditional idea, that notions like grammatical *Subject* or grammatical *Object* have to be defined, respectively, in terms of the Semantic notions of agent of an action and patient of an action. To this purpose, he brings counterexamples, such as Sentences like *'John received a letter'*, or '*The fighting stopped*', where the grammatical *Subjects* do not satisfy such Semantic requirements. Chomsky's alternative proposal is to define *Subject* and *Object* in purely syntactic and formal terms, i.e., *by means of particular configurations of syntactic Descriptions.* Later on, notions like *Agent* or *Patient* also entered the theory as *thematic roles* and played a crucial part in the *Government & Binding* version of Chomskian framework.

---

[29]When *'seem'* is used to describe a perception or an appearance, it is followed by an adjective or an adjective phrase, not a verb.





The *autonomy of syntax* is used to sort out the formal and Semantic aspects within grammatical components. The idea is to set apart syntactic and Semantic notions, so that the issue of their interaction is posed at the empirical level. Chomsky's Principle had (and still has) the crucial role of allowing the relationship between formal syntactic computational devices and the formation of Semantic unitary expressions to be set as an empirical question, rather than as a methodological assumption (not to mention: *A Dogma imported from Logics*). Chomsky's position with this regard is:

> *"It seems clear, then, that undeniable, though only imperfect correspondences hold between formal and Semantic features in Language. The fact that the correspondences are so inexact suggests **that meaning will be relatively useless as a basis for grammatical Description.** [...] To put it differently, given the instrument Language and its formal devices, we can and should investigate their Semantic function [...]; **but we cannot, apparently, find Semantic absolutes, known in advance of Grammar, that can be used to determine the objects of Grammar in any way."** [29]*

Can we say the same about Arabic?

Although answering this question in a thorough manner is beyond the scope of this paper, we can, nevertheless, make the following observations about Syntax-Semantics relationship in Arabic:

1- *Grammaticalness is a necessary condition for meaningfulness*: If an Arabic Sentence is semantically correct, then it must also be grammatically correct, but not vice versa (*Role of Grammar,* Section (ii))[30].

2- Marking Sentences with a correct set of *ḥarakāt* is a sufficient condition for grammaticalness (*Necessity of ḥarakāt***,** Section (ii)). Since *ḥarakāt* are *meaning-particles*, this is strong evidence of the involvement of meaning in Grammar.

3- More evidence: *Harakāt* used in marking grammatical roles of Descriptions are a measure of their Definiteness: The more indefinite the Descriptions are, the more they are eligible to the acceptance of *ḥarakāt*. Definite Descriptions can accept either rigid, kind of built-in *ḥarakāt* (Arabic: *'AL$^a$M$^a$t$^u$    alB$^i$N$^{a u}$*) or semi-rigid ones, not distinguishing all grammatical cases (Words accepting those latter *ḥarakāt* are called: *'m$^a$MNuAt min al-S$^a$RF'*). The idea is to mark a word according to the degree of its ability to express common

---

[30] Many Arabic Grammar Rules, especially those related to *ḥarakāt*, are caused by phonetics. For example: The two Sentences: S1 = *'SH$^a$R$^i$B$^u$  al-KH$^u$MR$^i$  fî  KH$^a$T$^u$R$^{i n}$'* and S2 = *'SH$^a$R$^i$B$^{u n}$  al-KH$^u$MR$^i$  fî  KH$^a$T$^u$R$^{i n}$'* both say the same thing: *'The alcohol consumer is in danger'*, but S1 is grammatically correct and S2 is not. Fact is: S2 cannot be pronounced properly, because of the morpheme: *'un'*, which should have been replaced, as in S1, by; *'u'.* Is S2 a semantically correct Sentence, although it cannot be correctly pronounced? Arabic Linguistics refuse to accept meaningfulness of such Sentences on the grounds, that they cannot be uttered correctly by any native Language speaker, i.e.: *There is no empirical evidence of their existence in the first place.* Note that including phonetic- and other meaning-unrelated reasons in Semantics is in contrast to the strict demarcation, proposed in Chomsky's principle.





behavior within the abstract type, class or concept it represents. The mark indicating the highest degree of common behavior (Arabic: *'t^aM^aKK^oN'*) being: *Tanween*. Such a degree is fore mostly expressed by *Common Nouns*, which are always marked with *Tanween*, whenever they occur in a Sentence. For example, in:

*'R^aAYt^u R^aG^uL^{an}' = 'I saw a man'* and *'R^aAYt^u al-R^aG^uL^a' = 'I saw the man'*

the marking *'an'* (*tanween*) indicates the indefiniteness of *'R^aG^uL'*, while *'al-R^aG^uL'*, performing the same grammatical role, is marked only with *'a'*.

4- Yet more evidence: In the last section, we have seen, how Arabic Grammar Rules are sensitive towards Definiteness as well. Indefinite Descriptions receive special treatment, when they are used as *Mubtada* in Noun Phrases: *They must be preceded by existential assertions, otherwise Sentences are deemed ungrammatical.*

5- The correspondence between Sentence constituents *Subject* and *Object* and notions of *Agent* and *Patient* is way more abstract, than the one thought of, assuming a relation between two objects. Agent-Patient Models are main-stream Semantic Models, attempting to explain Grammar Rules of Arabic [15]. However: As per *AntiDogma1*, Sentence Constituents don't denote objects, but *meanings,* which are: *Patterns recognized by the mind either by definition, perception or abstraction.*

Going back to Chomsky's examples: When we say: *'John t^aL^aQQA R^iS^aL^at^{an}' = 'John received a letter'*, the Arabic verb: *'t^aL^aQQA'*, corresponding to the English verb: *'receive'*, is based upon morpheme: *{t^a @ a $$ a % a}*, which expresses *'compliance'*. This helps clarifying the Semantic picture: We are not describing someone giving John a letter, in which case John would be the *Patient*, but the mental act itself of getting the letter, an act of compliance, in which John is the complying party, i.e., the *Agent*. The same analysis is valid for *'t^aW^aQQ^aF^a al-Q^iT^aL^{u'}* = *'The fighting stopped'*.

This section ends with an analysis, from the perspective of Arabic, of examples used usually to show the inadequacy of classical *Subject-Predicate* Models for expressing logical contents of English Sentences (see: [31] for example). The analysis here shows how easily Arabic Grammar handles logical concerns posed in all those cases:

a- *'John hit Smith' = 'John D^aR^aB^a Smith'* and *'Smith was hit by John' = 'D^oR^iB^a Smith min John'*:

Both Sentences express the same thing, but in the first English one, *'John'* is the *Subject*, while in the second English one, *'Smith'* is the *Subject*. Critics complain, that, if Grammar is supposed to reflect logical content, then the only *Subject* of the verb *'hit'* in any of its passive or active forms should be *'John'*, i.e.: There shouldn't be two different *Subjects*. In the second Arabic

Sentence: *'Smith'* is not a *Subject*, but a Grammar category between *Subject* and *Object* called: *'Deputy Subject'* (Arabic: *'NaIB FaIL'*), which bears the Grammar markings of a *Subject*, but is logically an *Object*. This category is used only within the scope of passive verbs (morphemes like: *{@ᵒ Šⁱ %ᵃ}*). Both Arabic Sentences are thus expressing the Logics intended here correctly: *There is only one Subject and the passive form of the Sentence doesn't change this fact.*

b- *'There is a problem'* = *'Honaka MᵒSHKⁱLᵘⁿ'* and *'It rained'* = *'aMTᵘRᵃt al-SᵃMᵃAᵘ'.*

In the first English Sentences: *'There'* acts as filler material, occupying the grammatical *Subject* position of an existential construction. In the second: *'It'* is similarly used to stand in the grammatical *Subject* position of the meteorological verb *'rain'*. Both cases are different in Arabic: The first is a Noun Sentence, in which: *'MᵒSHKⁱLᵘⁿ'* = *'a problem'* is *Mubtada* (corresponding to the *Subject*) and *'Honaka'* is the *Khabar* (corresponding to the *Predicate*), asserting existence. *Khabar* comes before *Mubtada* in this case, because the latter is an Indefinite Description. The second Arabic Sentence is a Verb Sentence, where the *Subject*: *'al-SᵃMᵃA'* = *'the heaven'* plainly follows the verb *'aMTᵘRᵃt'*, in which the last character *'t'* indicates the feminine form of the *Subject* and no pronouns are used. In both cases: *Subject positions are filled with concrete Descriptions, not mere 'fillers'.*

c- *'Not everybody is coming'* = *'laysᵃ al-GᵃMⁱOᵘ AaTiYᵃⁿ'.*

In one possible grammatical analysis of this English Sentence the *Subject* is *'Not everybody'*, in which case it looks as if the entity: *'Not everybody'* possesses the property of: *'coming'*, an absurd Semantic interpretation. In the Arabic equivalent Sentence: *'laysᵃ'* = *'Not'* is a syntactic operator, applied to the positive Sentence, placing on: *'al-GᵃMⁱO' (Mubtada)* and *'al-GᵃMⁱOᵘ AaTiY' (Khabar)* different markings: *'u'* and *'an'*, respectively, to underline this function. The Arabic constituent corresponding to the *Subject* of this Sentence is: *'al-GᵃMⁱO'*, making the meaning-revealing scope look like: *'laysᵃ (al-GᵃMⁱOᵘ AaTiYᵃⁿ)'* = *'Not (everybody is coming).'*

All the above and much more, omitted here to avoid unnecessary length, shows the existence of a vivid relationship between Grammar and meaning in Arabic, surpassing similar phenomena observed in Indo-European Languages. We note that:

*1-* Although deciding grammaticalness of a Sentence is largely simplified through the use of *ḥarakāt*, this is not sufficient to determine the exact meaning: There might exist more than one set of *ḥarakāt*, causing ambiguity. *Meaning* remains as unattainable through formal methods as it is in other Languages, including those of Logics. In the next part we shall see, however, that *it is fortunately not meaningfulness, but grammaticalness, which sits at the heart of computation.*

2- There is an overwhelming abundance of *meaning-particles* in the surface structure of an Arabic Sentence, covering almost all aspects relevant to Logics: *Constituency, Naming, Reference, Definiteness, Quantification and Predication.* How can there be, then, any need for a *'deep logical structure'*?

### viii. Quantification anomalies and other Paradoxes which are not

As per Chomsky's quote in the introduction, the current state of the art in Linguistics, regarding *deep structure* and its relation to Logics is as follows:

> *'[…] relating deep structure to logical form was given up in the 1960s, with the discovery of surface structure effects on meaning. Decades ago deep structure was given up altogether as superfluous'*

As expected, this position, although concurring with our findings in Arabic, leaves many loose ends from the Logicians point of view.

Fact is, that *'meaning'* in Computational Linguistics is still built upon Tarskian notions of Model-based Semantics, adopting almost all ideas of Frege and Russell and manifested clearly in Montagues work.

To understand why this is the case, one must go back in time to witness the emergence of what was called then: The *'Generative Semantics'* movement (quotes from [30]):

> *"Stemming directly from within the generative enterprise, the Generative Semantics movement brought a strong attack to the thesis of the Autonomy of Syntax, by defining 'deep syntax' as actually a logico-Semantic level. Not very differently, Montague presented a new Model of logical Grammar according to which - thanks to a proper pairing of Semantic and syntactic operations – Sentences are analyzed in such a way to exhibit the logical form directly on their syntactic sleeves […]*
>
> *Crucially, starting from the assumption that the linguists' concerns and the logician's are consistent with each other, Generative Semantics was able to raise a whole wealth of issues concerning the relation between grammatical structure and logical form, thus showing the inadequacies of the first transformational Models to tackle these aspects of the theory of Language. **Many problems that attracted the attention of generative Semanticists, like quantification, bound anaphora, etc., passed then to occupy a key position in the later developments of the Chomskian framework**, thus leading towards more elaborated hypotheses on the syntax-Semantic interface […]*

*The crucial element of novelty in this line of analyses is that **deep structure is now regarded as an abstract level with the same format as first-order logic representations**. In other terms, first order logic structures are syntactically 'wired' in deep syntactic Descriptions. In fact, the latter contain not only lexical items, but also abstract elements, such as variables bound by quantifiers or logical operators, like negations, modals, etc. These abstract constructs are then turned into surface phrase structures by the application of various types of transformations - such as for instance Quantifier Lowering - which replace and inserts lexical items or delete abstract elements.*

*In summary, bringing to its extreme limits the assumption that deep structure has to provide all the compositionally relevant information, **generativist Semantics came to abandon the idea itself that deep structure is syntax at all**, thus performing a radical departure from the Principle of the Autonomy and Chomsky's Standard Theory architecture of Grammar. **Instead of postulating syntactic representations that feed an interpretive Semantic component, Semantics was intended as a generative device that produces the deep layer directly encoding the logical form of Sentences-, which is then turned by transformation into surface structures.** Therefore, generative Semantics pursues the view that in Grammar "there is no dividing line between syntax and Semantics", exactly because logico-Semantics aspects - ranging from quantifier scope, to presuppositions, implicatures and speech acts are directly built in deep structures."*

From the quotes above it seems clear, why the idea of the existence of a *'deep structure'* became, with time, superfluous: On the one hand, Linguists, guided by Grammar aspects, found an abundance of surface structure effects on meaning. On the other: Logicians, whose main concern was the logical Model behind the Sentence, found no added value in assuming the existence of a Natural Language layer, which turns out to be very similar to common logic representations.

Where does Arabic stand?

Let us first note that Arabic is concerned with a truly existing two-way relationship between Syntax and Semantics, which is not similar to anything known from Logics. As we saw in the last sections: There is an abundance of *meaning-particles* in Arabic surface structures, but those particles are *not* coinciding neither in nature nor in manifestation with Logicians views.

This fact provides a motivation to re-analyze problematic Sentences, claimed by Logicians to be Language-anomalies or paradoxes, and investigate: First, whether the same problems exist in Arabic contexts and second, if they do exist, whether there are grammatical means to overcome them, making the idea of a separate logical layer unnecessary.

As a thorough study of the two questions bypasses the scope of this work also, we have chosen to analyze some representative examples of alleged Language anomalies occurring in quantification contexts, before making our concluding remarks for this part.

We include in our analysis some known anomalies related to Identity statements as well, since they were raised in the context of choosing modal

Logics with possible world Semantics as a *'deep layer'* in the place of ordinary first order Logics:

### 1- Partee's paradox:

A notorious example of what is called *Language Anomalies* is *Partee's Paradox*, for which it is said (see: [32], for example), that from the three following Sentences:

1. The temperature is ninety.

2. The temperature is rising.

3. Therefore, ninety is rising.

The obvious invalidity of 3. can only be recognized (without abandoning Leibniz's Law of substitution), if the underlying formulization captures the fact that the first premise makes a claim about the temperature *at a particular point in time*, while the second makes an assertion about *how it changes over time*. One way of doing that, proposed by Montague, is to adopt Intentional Logics for Natural Language, thus allowing *'the temperature'* to denote its *extension* in the first premise and its *intension* in the second. Montague took this example as evidence that a Nominal denotes an individual concept, defined as functions from a world-time pair to an individual. Later analyses built upon this general idea, but differed in the specifics of the formalization.

In first order Logics, the above problem Description may be expressed as follows:

1. Temperature = 90
2. Rising(Temperature)
3. Rising(90)

Where 3. Is a valid conclusion.

This analysis, which is based upon the *Referential Doctrine*, assumes, that English Sentences using Copula are identity statements, of course.

We saw in previous sections, that this assumption is incorrect. Moreover: Such an example doesn't apply for Semitic Languages, which don't use Copula, notably: Arabic. Moreover: *Anti-Dogma1* contradicts the Referential Doctrine as we already established. *All this makes the whole problem formulation completely nonsensical to Arabic.*

### 2- Operator scope ambiguities:

A lot has been written about *Scope Ambiguities* in the state-of-the-art literature of Linguistics and Logics. The common understanding is that Semantic Operators are responsible for such ambiguities and that they are best understood by contemplating logical, not syntactic phenomena.

We show here that some selected, important types of scope ambiguities either *do not exist in Arabic at all*, or can be easily explained using Grammar Rules and adequate *Syntactic* Operators.

The following analysis of our selected examples shows that, because of the adoption of Russell's idea of substituting Definite- and Indefinite Descriptions by existential quantifiers, artificial ambiguities induced in this way fail to reflect correctly the intended meanings of Arabic Sentences.

**a- Quantification scopes:**

Are *All-Quantifiers* used in Logics correctly modelling their counterparts in Natural Language?

Take for example the Quantifier: *'Everyone'*.

As per [33]: The word *'Everyone'* is an indefinite pronoun. That is to say, it is a pronoun that refers to an indefinite group of people. *'Everyone'* (one word) is a synonym for *'Everybody'* (although *'Everybody'* is slightly less formal), and it means all the people, every person. *'Everyone'* always refers to humans, or to humanity in general.

On the other hand: The phrase *'Every one'* (which combines a Modifier and a Noun) is more explicit, referring to each individual or thing in a particular group. *'Every one'* is usually followed by the preposition *'of'*. In practice, *'Every one'* is a near synonym of *'Each one of a set'*, so it does not necessarily refer to people at all.

There is, hence, a difference between the *collective*- and the *distributive* meaning of this quantifier: While *'Everyone'* refers to the collection as a whole (called henceforth: *'the linguistic collective'*), *'Every one'* indicates each and every member of this collection.

Obviously: The distributive meaning is the one modelled by *All-Quantifiers* in Logics, since this is enforced through directly binding variables referring to objects to quantifiers, and we must assume in all the following examples, that the *linguistic collective* will be irrevocably lost:

i-   *'Everyone loves someone'* = *'Al-G$^a$M$^i$O$^u$ y$^u$H$^e$BB$^u$ SH$^a$KHS$^{an}$ ma'*

To be able to analyze this Sentence using logical quantifiers, it must be re-written to become:

*'Every one loves someone'* = *'K$^o$ll$^u$ W$^a$H$^i$D$^{in}$ y$^u$H$^e$BB$^u$ SH$^a$KHS$^{an}$ ma'*

It is claimed, that this Sentence has the following two possible meanings, of which the second is a collective one:

     1-   *For Every x: Some y exists such that: love(x y)*

     2-   *Some y exists such that: For Every x: love(x y)*

If the *linguistic collective* is lost, as we just saw: Where does the collective meaning in Sentence 2- come from?

There are two aspects of the collective meaning in 2- (which we call henceforth: *'the logical collective'*), both of them caused by the mere usage of quantifier-bound variables and their respective places relative to each other, not by any other Semantic consideration related to the verb itself:

  a-  *That all elements of the collection are Agents of the verb*
  b-  *That there is one and only one Patient*

Looking at the two different Quantifiers in the Arabic Sentences: *'Al-G$^a$M$^i$O'* and *'K$^o$ll$^u$ W$^a$H$^i$D'*, we find that, as per Arabic Semantic Rules: The first one stands for the majority of elements of a collection, while the second relates to each and every element of the same collection. *'Al-G$^a$M$^i$O'* is, therefore, out of question as a means to reflect aspect *a-* of the *logical collective*.

While *'K$^o$ll$^u$ W$^a$H$^i$D'* fulfills *a-*, we must ask: Is it possible to use this Quantifier to express aspect *b-* as well, i.e.: Making the entity referred to by the Indefinite Description *'person' Object* of the verb *'to love'*, while in the same time expressing, that every single member of the collection *'loves'* this one and only one entity?

Returning to the Sentence, we realize that the Indefinite Description *'person'* used as the *Object*: *'SH$^a$KHS$^{un}$'* is a *Common Noun*. From Section v (point 2 of the features of Indefinite Descriptions in Arabic) we remember that: *Common Nouns are words denoting types of things. None of those things are intended to be singled out, with respect to the properties included in the given mental concept.*

This clearly contradicts our intention to single out an entity, whose love is shared by everyone. It also explains, why modifying the Sentence to express *sameness* of the Object of love can only be done in Arabic using Definite Descriptions:

  o  *'K$^o$ll$^u$ W$^a$H$^i$D$^{in}$ y$^u$H$^e$BB$^u$ N$^a$FS$^a$ **Al-**SH$^a$KHS$^{i}$'* = *'Every one loves **the** same person'*
  o  *'Al-K$^o$ll$^u$ y$^u$H$^e$BB$^u$ **Al-**SH$^a$KHS$^a$ Z$^a$T$^u$**h$^u$**'* = *'Every one loves **the** same person'*
  o  *'H$^o$n$^a$k$^a$ SH$^a$KH$^{un}$ y$^u$H$^e$BB$^u$**h$^u$** Al-K$^o$ll$^u$'* = *'There is a person, **who** is loved by every one'* [31]
  o  *etc…*

---

 The pronoun *'hu'* is the *Object* in the verb: *'y$^u$H$^e$BB$^u$**h$^u$**'* and like all pronouns attached to a Verb: *Definite*.

Resuming our analysis of this example: *The logical collective, manifested in 2- cannot be expressed using Indefinite Descriptions in Arabic. Hence: Only meaning 1- can correspond to the re-written Sentence, the original Sentence being completely out of the scope of usual logical analysis, since the linguistic collective is lost.*

ii-     *'A man climbed every tree' = 'R$^a$G$^u$L$^{un}$ t$^a$S$^a$LL$^a$Q$^a$ k$^o$ll$^a$ SH$^a$G$^a$Ra$^{tin}$'*

This Sentence has, similar to the previous one, the following two interpretations, as per the Logicians analysis:

a-     *Some x exists: For every y: Man(x) and Tree(y) and climbed(x,y)*

b-     *For every y: Some x exists: Man(x) and Tree(y) and climbed(x,y)*

We note, when converting this Sentence to Arabic, that the literal translation: *'R$^a$G$^u$L$^{un}$ t$^a$S$^a$LL$^a$Q$^a$ k$^o$ll$^a$ SH$^a$G$^a$Ra$^{tin}$'* doesn't correspond to a complete Sentence.

As mentioned before (*Principle of constructing Noun Sentences in Arabic*, section i): The Indefinite Description *'R$^a$G$^u$L$^{un}$'* cannot assume the role of *Mubtada*, unless it is preceded by an existential assertion, like in: *'H$^o$n$^a$k$^a$ R$^a$G$^u$L$^{un}$ t$^a$S$^a$LL$^a$Q$^a$ k$^o$ll$^a$ SH$^a$G$^a$Ra'* = *'There is a man, who climbed every tree'*, which is clearly of the form a-, not b-.

We could also formulate a Noun Sentence, in which the *All-Quantifier* is *Mubtada*, creating a passive construction, like in: *'k$^o$ll$^u$ SH$^a$G$^a$R$^a$t$^{in}$ t$^a$S$^a$LL$^a$Q$^a$h$^a$ R$^a$G$^u$L$^{un}$'* = *'Every tree was climbed by a man'*, indicating the meaning: *b-, not a-*.

Another alternative of completion is to convert the Sentence to a Verb-Sentence, in which the Indefinite Description must come after the Verb, like in: *'t$^a$S$^a$LL$^a$Q$^a$ R$^a$G$^u$L$^{un}$ k$^o$ll$^a$ SH$^a$G$^a$Rat$^{in}$'* = *'climbed a man every tree'*, for which the literal translation to English doesn't provide a correct Sentence. Here, the Indefinite Description assumes the *Subject* position and is outside the scope of the All-Quantifier, leaving no doubt, that one man only is meant (interpretation: a-).

Can we place the *Subject* in this Sentence after the *Object*, like in: *'t$^a$S$^a$LL$^a$Q$^a$ k$^o$ll$^a$ SH$^a$G$^a$Rat$^{in}$ R$^a$G$^u$L$^{un}$'* = *'climbed every tree a man'*, so that we keep the Indefinite Description within the scope of the *All-Quantifier*? As per Arabic Grammar Rules: No, we can't, unless some special conditions related to the *Object* are fulfilled, which are out of question in this case.

Resuming our findings: *The literal translation of the English Sentence doesn't form a correct Arabic one. Completing the Arabic Sentence, by moving the Indefinite Description, either provides unambiguous meanings, or prevents ambiguities to occur, if Grammar Rules are strictly followed. In all completion cases: The meaning is unambiguously understood, without having to convert the Indefinite Description to an existential Quantifier.*

**b- Negation- within Quantification scopes:**

iii-        *'Everybody didn't come' = ' Al-GᵃMⁱOᵘ Lam yatⁱ '*

This Sentence is usually investigated to account both for the ambiguity of *all…not* constructions and the lack of such an ambiguity (for the most part) of *some…not* constructions [34]:

> *"An explanation for these facts is offered by Horn (1989: chapter 7.3.3). According to him, the reason why the NEG-Q reading is not available for (13)a' is that this meaning can unambiguously be expressed by (13)b' [..]. Horn claims that* **inherently negative quantifiers are less marked than negated quantifiers**.
>
> *(13)*
>    a.  *Everybody didn't come. a'. Somebody didn't come.*
>    b.  *Not everybody came. b'. Nobody came.* **NEG-Q**
>    c.  *Nobody came. c'. Not everybody came.* **NEG-V**
>
> *But the situation is different for (13)a with a universal quantifier. There is no lexicalized quantifier to express this NEG-Q reading [..]. While the availability of the fully lexicalized, and thus unmarked, 'nobody' restricts the interpretation of (13)a' to NEG-V, the morphologically and syntactically more marked 'not everybody' has a "relatively weak restrictive effect on the use of (13)a to convey its potential NEG-Q meaning" (Horn 1989: 499). This explanation predicts that "NEG-Q readings will be available for those predicate denials which do not have a lexicalized paraphrase" (Horn 1989: 499) – and this prediction, according to Horn, turns out to be accurate, if one examines actual Language use"*

Do we see in Arabic the same phenomena?

We notice first, that, as in the example Sentence *(i)* in this section: *The linguistic collective* is lost and we need to re-write Sentences: *a* and *a'* as follows:

> *a''- 'Every one didn't come' = 'Kᵘllᵘ WᵃHⁱDⁱⁿ Lam yatⁱ'*
>
> *a'''- 'Someone didn't come' = 'SHᵃKHSᵒⁿ ma Lam yatⁱ'*

In Arabic there is a clear distinction between Negation Operators used with Verbs and those used with Nouns. *'Lam'* is an operator used with Verbs, so that both Sentences must be NEG-V. No adequate translation of the Sentences may choose Negation Operators used for Nouns in the place of *'Lam'*.

While *a''-* means: *'Nobody came'*, because the Quantifier is *Mubtada* and the Verb phrase is its *Khabar*, i.e., we are asserting something about every single one of the members of the collection, namely: That he didn't come, *a'''-* is not a correct Arabic Sentence in the first place, because it starts with an Indefinite Description (unless it is understood as an answer to a question in a previous context).

As mentioned several times for such cases: Moving Indefinite Descriptions behind the Verb (in a Verb Sentence) or behind the *Khabar* (in a Noun Sentence) corrects the situation and we get in this case:

$$a'''\text{- 'Didn't come someone'} = \text{'Lam yat}^i \text{ SH}^a\text{KHS}^{on} \text{ ma'}$$

which isn't a correct English Sentence, but means in Arabic: *'Some particular person didn't come'*, a meaning, different from any of the meanings *b.* and *c.*

What, if we move the Quantifier in *a''*- behind the Verb to get:

$$a''\text{- 'Didn't come every one'} = \text{'Lam yat}^i \text{ K}^u\text{ll}^u \text{ W}^a\text{H}^i\text{D}^{in}\text{'?}$$

While also not being correct in English: This Sentence means in Arabic: *'Not every one came'*, indicating the *NEG-Q* nuance, because the negated Verb phrase is meant now to include the Quantifier. None of all this has to do with lexicalized paraphrases of quantifier expressions.

In summary: *The grammatical distinction between Negation Operators working on Verbs and others working on Nouns reduces the ambiguity to a minimum. This latter may result from moving syntactical constituents of the Sentences around, not from any other Semantic considerations beyond the order of Syntax.*

iv- *'I demanded that you read not a single book'* = *'T*$^u$*L*$^a$*Bt*$^u$* minka an la t*$^a$*QRA*$^a$* K*$^i$*T*$^u$*B*$^{an}$* W*$^a$*H*$^i$*D*$^{an}$*'*

As proposed by Logicians [35]: Scopes of the Indefinite Description and the Negation-Operator create two different meanings, the second, especially when *'single'* is understood to mean *'particular'*:

1-    *I demanded that: Not (Some x exists: Book(x) and you read x)*
*(I demanded that there is no single book that you read) = (T*$^u$*L*$^a$*Bt*$^u$* minka an la [yUG*$^a$*D*$^a$* K*$^i$*T*$^u$*B*$^{un}$* W*$^a$*H*$^i$*D*$^{un}$* anta t*$^a$*QR*$^a$*A*$^u$*h*$^u$*] )*

2-    *Not (Some x exists: (Book(x) and I demanded that you read x))*
*(It is not the case that there is a particular book, that I demanded you read) =*
*(la [yUG*$^a$*D*$^u$* K*$^i$*T*$^u$*B*$^{un}$* W*$^a$*H*$^i$*D*$^{un}$* T*$^u$*L*$^a$*Bt*$^u$* minka an t*$^a$*QR*$^a$*A*$^u$*h*$^u$*'])*

We note first that 2- is logically equivalent to:

2'- *For all x: (Book(x) implies Not (I demanded that you read x))*
*(I didn't demand that you read any book)*

Which is different in meaning from: *'I didn't demand that you read any **particular** book'*, the explicit intention behind 2.

As already seen in section v (point 4, *Inadequacy of existential translations*): Correctly modeling the meaning of the adjective *'particular'* attributed to an Indefinite Description requires referring to mental entities, not existing objects of the world.

Eventually it is not surprising, that we don't have the same problem in Arabic: The adjective *'single'='W*$^a$*H*$^i$*D*$^{an}$*'*, which is attached to the Indefinite Description, is understood to refer to a count, contradicting *'many'* and making the overall meaning equivalent to:

*'I demanded that you read more than one book' = 'T$^u$L$^a$Bt$^u$ minka an t$^a$QRA$^a$ aKTH$^a$R$^a$ min K$^i$T$^u$B$^{in}$ W$^a$H$^i$D$^{in}$'.*

However: In case the adjective *'particular'='m$^u$AYY$^a$N'* is used, like in: *'T$^u$L$^a$Bt$^u$ minka an la t$^a$QRA$^a$ K$^i$T$^u$B$^{an}$ m$^u$AYY$^a$N$^{an}$'*, ambiguity arises between the following two meanings:

    i-      *I demanded that you read more than one book*
    ii-     *I demanded that you don't read some specific book*

Because of two possible nuances of *'particular'='m$^u$AYY$^a$N'*: One indicating the count and the other indicating definiteness. *Both meanings are not equivalent to 1- or 2-, which is a clear indication, that scopes of Semantic Quantifiers are not the source of ambiguity here.*

### ix.  Summary of findings of part A

Starting from findings of the last Section, we summarize what we have shown in part A:

1-    *Scope ambiguities* are drastically reduced, when, instead of substituting an existential quantifier for each Definite/Indefinite Description (as prescribed by Russell's *Dogma2*), the same Descriptions are understood to stand for imaginary entities, representing mental concepts, not objects (*AntiDogma2*):

    a. Interfering scopes of *syntactical* operators, caused by constituent's movements, are, in addition to ambiguous lexical meanings, the true sources of ambiguity.

    b. The fact, that Arabic Grammar treats Indefinite Descriptions in a special way, by requiring existential assertions and Verbs to precede them, reduces freedom of movement of constituents and thus also related ambiguities.

    c. Because the *logical collective* cannot be modeled using Indefinite Descriptions, some studied forms of scope ambiguities do not occur in Arabic in the first place.

    d. Fregean Logics does not permit to model the *linguistic collective*, so that not all possible meanings of a Sentence in Arabic or English can be modelled or assumed to reside in any *'deeper, logical form'*.

2-      Alleged Language paradoxes, like *Partee's paradox*, caused by applying *the Referential Doctrine* (*Dogma1*), while interpreting *ToBe*-constructions as identity statements, are artificial, erroneous constructions, not related to how Natural Language, notably Arabic, truly works.

3-      Semantic Roles of Arabic Sentence Constituents, like: *Subjects*, *Objects*, etc. do not contradict assigned logical roles (*Agent* and *Patient*), as long as an adequate level of abstraction is chosen.

4-      The relation between Grammar and meaning in Arabic is bi-directional: Grammaticalness is a necessary condition for meaningfulness and there is an abundance of *meaning-particles* in Grammar.

5-      Applying the *Referential Doctrine* causes Language anomalies to occur. We have identified two in particular: The *Actor Anomaly* and the *Inadequacy of existential translations Anomaly*. Both disappear, when *AntiDogma1* and *AntiDogma2* are applied instead.

6-      The fact, that, as per *AnitDogma2*, an Arabic Noun Sentence, asserting some property about an Indefinite Description, can only be grammatically correct, if the Indefinite Description is preceded by an existential assertion related to it, defeats the purpose or Russell's idea of substituting existential quantifiers for Indefinite Descriptions (*Dogma2*).

7-      Russell's meaning-notion (*Dogma2*) is referential in a strictly materialistic sense and therefore different, in principle, from the meaning-notion presented in *AntiDogma3*, according to which the overall meaning of all Descriptions, whether Definite or Indefinite is composed from *Root*- and *Template-morphemes*, which are meta-symbolic structures, not taken into account at all in Fregean Logics.

8-      Russell's notion of *incomplete Symbols* is fundamentally flawed: In Arabic, Nouns and Verbs will always hold meaning nuances, independent of any context of use.

9-      Frege- and Russell's incorrect interpretation of *ToBe*-constructions lead to imagining Identity statements, where there are none. This can be demonstrated, not only in Arabic, but also in English. *Identity*- and *Substitutivity Puzzles* are, therefore, not genuinely reflecting how Natural Language works.

10-     Semantic Models of Logics cause formal Systems adopting them to *over- and under-accept* Natural Language Sentences. On the one side they relativize otherwise fixed Symbol denotations and on the other: Reject *no-Object*-denotations. *Logical validity of a Sentence is unrelated, in principle, to its Semantic validity in Natural Language.*

11-    Adhering to Arabic Language *Anti-Dogmas* enables not only a solid theoretical foundation for linguistic Models, but also overcoming processing shortcomings, through extensive use of *meaning-particles*, which help in disambiguating denotations of words, phrases and Sentences.

12-    *Anti-Dogmas* reflect the existence of *descriptive necessities* in Arabic, which provide, philosophically, a middle-way between: Accepting logical necessities as a basis for assigning denotations to Descriptions while *Naming* and leaving those denotations scientifically unaccounted for.

13-    To classify an Arabic Sentence, without using *ḥarakāt*, an important type of *meaning-particles*, is exponentially more expensive in most cases, than when *ḥarakāt* guide derivations.

All the above points may be regarded as good reasons for rejecting Logicians *Dogmas* as adequate, formal Semantic assumptions to be used for modeling Arabic.

We end this part with a quote from [36], where Chomsky answers an author's question with regard to Logicians *Dogmas*, notably: *The Referential Doctrine*:

> "I think you're right in focusing on the 'logicians view of Language and their insistence to enforce objects of the world on relations linking phrases to their meanings in natural Language.' I've put the matter a bit differently in my own work for many years: Natural Language does not have the relation of reference/denotation in the logicians' sense – the 'Word-Object' sense.

> This undercuts much contemporary work but I don't think it would have much disturbed Frege or Russell, or their prominent successors like Tarski and Carnap, all of who regarded human Language as 'defective' and sought to construct ideal Languages in which the posited relations hold – like meta-mathematics, in which the numeral 3 denotes the number 3. Even Quine, who comes closest to concern for natural Language, was really interested in what he calls 'regimented Language'.

> I think you're right about the pernicious effect of applying the referential doctrines of the logicians to natural Language, where they don't work."

## VII-    Part B: From Language Sentences to Satisfiability Problems

### i.   Why is Language recognition NP-complete?

In the first part of this work, we have seen that, other than adequately formalizing reference, which is the main pillar of *'meaning'* for Logicians, Arabic requires equally important Semantic aspects to be formally dealt with.

An adequate formal treatment of Arabic has also to include: Modelling mental entities, existing per se, without being manifested as objects, regarding Indefinite Descriptions to be referring to such entities, recognizing *meaning-particles* in Syntax and using them to classify types and roles of Sentence constituents, utilizing meta-Symbols to form Nouns and Verbs, letting the latter possess basic meaning nuances, independent of any interpretation or context, all that within a flexible Grammar framework, correctly incorporating Semantic features into Syntax Rules.

Unfortunately, when aspects of meaning are either not understood or not accounted for, *arbitrariness* is the only remaining alternative.

This can also be observed in artificial Languages, like first order Logics: As mentioned in the introduction, Satisfaction procedures, seeking to verify validity of logical formulas, implement Assignment Functions in a way completely independent of both Semantics and Syntax, and: *There is neither a logical nor a structural reason for the choice of a particular variable to refer to a particular object in a domain D or for selecting a particular order in which variable assignments are made.*

This must be done by simple trial and error, until a correct solution is found. We called this property in the introduction of this work: *Selection Arbitrariness*. It is a direct consequence of the strict requirement, that Symbols bear no meanings for themselves (*Dogma3*).

Of course, there is nothing simple in trial and error: As formal Logics became the foundation of modern computation, trial- and error-based procedures soon found their place at the heart of complexity in the form of *Brute Force Procedures*, applied whenever there are no apparent rational criteria to choose between some given alternatives. Seeing them today from the perspective of Energy consumption: *Brute Force Procedures are some of the worst inventions human beings brought upon planet earth.*

In NLP it became quickly clear, while studying properties of so-called *Lexical Functional Grammars* (*LFGs*), for example, that even finding the right type of a word in an English Sentence (i.e., whether it is Noun or Verb) can become, in the worst case, an expensive endeavor, requiring trial and error. It turned out, that this fact was enough to *classify LFG-word problems as NP-complete problems*.

*LFG* views Language as being made up by multiple dimensions of structure [37]. Each of these dimensions is represented as a distinct structure with its own Rules, concepts, and form. The primary structures that have figured in *LFG* research are:

Representation of grammatical functions (*f-structure*)
Structure of syntactic constituents (*c-structure*)

This dissociation of syntactic structure from predicate argument structures (essentially a rejection of *Chomsky*'s *Projection Principle*[32]) is crucial to the *LFG* framework. While *c*-structure varies somewhat across Languages, the *f*-structure representation, which contains all necessary information for the Semantic interpretation of an utterance, is claimed to be universal.

The lexical entry (or Semantic form) includes information about the meaning of the lexical item, its arguments, and the grammatical functions (e.g., *Subject*, *Object*, etc.) that are associated with those arguments. Grammatical functions play an essential role in *LFG*, however, they have no intrinsic significance and are located at the interface between the lexicon and the Syntax. *LFG* imposes the restriction of *Direct Syntactic Encoding*, which prevents any syntactic process from altering the initial assignment of a grammatical function.

A thorough look into the *3SAT*-reduction used in the proof of NP-completeness of LFG word problems given in [38] reveals the deep reason for the presumable intractability:

> "*One and the same terminal item can have two distinct lexical entries, corresponding to distinct lexical categorizations; for example, baby can be both a noun and a verb. If we had picked baby to be a verb, and hence had adopted whatever features are associated with the verb entry for baby to be propagated up the tree, then the string that was previously well-formed, "the baby is kissing John", would now be considered deviant. If a string is ill-formed under all possible derivation trees and assignments of features from possible lexical categorizations, then that string is not in the Language generated by the LFG. **The ability to have multiple derivation trees and lexical categorizations for one and the same terminal item plays a crucial role in the reduction proof**: it is intended to capture the satisfiability problem of deciding whether to give an atom $X_i$ a value of T or F.*"[33]

Accordingly, the essential problem lies in the fact, that one element of Syntax may possess two different Semantic roles and there is no syntactical way to distinguish between them. No surprise there: *SAT* is the problem of efficiently linking Syntax to Semantics of simple Propositional Logics formulas, upon which *Dogmas* are applied, which prohibit Symbols from exhibiting intrinsic meanings.

Can we have Terminal Items with multiple lexical categorizations in Arabic also?

Yes, of course[34,] but we have also: *Noun distinction criteria* (part A, section (ii)), which are *meaning-particles*, helping to distinguish between those categorizations in a syntax-based way. However: Those criteria relate only to Arabic Grammar and cannot be applied else where.

If we want to get a deeper and simpler understanding of the efficiency puzzle, we need a more direct way to look at *SAT*, which belongs to the artificial Language of Propositional Calculus.

---

[32]   Under the Projection Principle, the properties of lexical items must be preserved while generating the phrase structure of a Sentence.

[33]   [38] p. 103

[34]   Like the word 'ahmad', which might, according to the context, be a Verb or a Noun.

Take the *2CNF* formula: S={{$x_0,x_4$} {$x_1,x_2$} {$x_2,x_3$}}, for example[35].

Propositional variables/literals $x0$, $x1$, $x2$, $x3$, $x4$ are all called by Frege: *Unsaturated*, having no meaning for themselves (similar to Russell's *incomplete Symbols*), because elements of the Domain: {*True*, *False*} need to be substituted for them to give $S$ any one of the two meanings. $S$ is also called *unsaturated*, until it becomes variable-free. To reach this stage, we might need, in the worst case, to try both possibilities of substitution for all literals. We do this either via a Truth-Table or using equivalent structures like *Binary Decision Diagrams* (*BDDs*) seen in Figures (1-a) and (1-b).

*Selection Arbitrariness,* the property of any conventional *Satisfaction procedure,* implies that there are no rational criteria to help us choose a suitable order of substitution of literals and verify validity of $S$ in the shortest number of steps possible. Such a choice is crucial for the size of resulting *BDDs* as can be seen in the Figures (slightly expanded for better illustration). They show unique non-terminal node counts of 5 and 10 respectively.

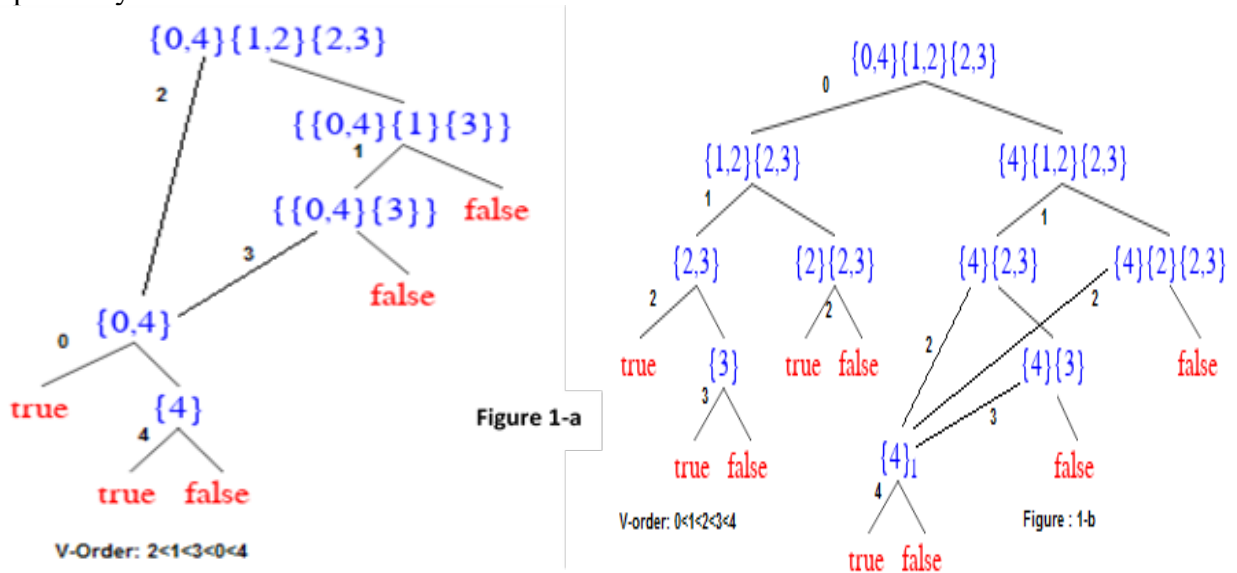

Figure 1-a

Suppose, we choose the ascending order to the right (we call it: *canonical*), i.e., there is a program $P$, which, given a *CNF* formula, applies the canonical order of substitution to produce *BDDs*. For the above $S$, we get 10 nodes as can be seen.

Renaming the variables of $S$ in the following way: [$x2 > x0$, $x1 > x1$, $x3 > x2$, $x0 > x4$, $x4 > x5$] produces a new clause set: $S' = \{\{x_4, x_5\} \{x_0, x_1\} \{x_0, x_2\}\}$, which is logically equivalent to $S$. How many unique nodes shall $P(S')$ produce? The reader is invited to check that the resulting *BDD* will look, if we ignore the markings on the edges, exactly like the one in Figure (1-a).

*By Adequate Renaming* of variables/literals, we have thus reduced the number of processing steps to half of the original amount. Are there sufficient criteria for a renaming of *S* to be *adequate*, i.e., to be delivering small sizes of resulting *BDDs* in all cases (*Small*: Polynomial in the length of *S*)?

---

[35] Taking *2CNF* formulas as examples does not cause any loss of generality, since they exhibit the same phenomena of relevance for this work as their *3CNF* counterparts.

One might argue, that not names of variables/literals, but *structural features* of the set *S* were responsible for the reduction: Literal $x2$ is present in two clauses and selecting it reduces significantly the number of nodes in the left part of the *BDD*. Similarly: Choosing unit clauses before others helped cut the resulting trees early on, etc. *Adequate Renaming* seems, therefore, only to be structural analysis in disguise.

Indeed: Current state-of-the-art *SAT* solvers, because of all reasons discussed in [39], for example, have many built-in structural pre-processing options, which reduce search spaces enormously. However: Their success is restricted only to special types of clause sets, ruling out the possibility that structural analysis alone could lead to *Adequate Renaming* in the general case.

Can we build upon this fact and try to find out, whether there are any sufficient *Semantic criteria* hiding behind *Adequate Renaming*?

If we try to do so using formal Systems, based upon *Dogmas*, then we face a number of obstacles.

First: *Dogma3,* according to which, Symbols don't have intrinsic meanings, forces us to give up the idea, that $x2$ or $x4$, for example, stands for any particular, fixed Semantic nuance. As per this Dogma and *Dogma1*: Literals are just empty vessels, which refer to objects: *True* or *False*. One consequence is our belief, that interchanging their names in *S* cannot have any effect on the evaluation of the formula.

Fact is, however, that while renaming has no effect on the final result of our evaluation (i.e., whether *S* is satisfiable or not), as just seen: *It has enormous effect on the way we attain this result.*

Second obstacle: Tarski's idea, that *'taking variables as names of objects is a Semantic notion'*, not amenable to formalization (see: *Footnote 1*), undercuts any efforts to find a reason, why the name *'x2'*, for example, could serve a different Semantic angle, than the name *'x4'*. Moreover: Even if such angles exist in form of criteria, adopting Tarski's view leaves us without any clue on how to express those criteria in syntactic form.

As usual, *arbitrariness* is the only remaining alternative: *Sticking to Logicians Dogmas forces us to try all possible naming conventions to find the best one.*

Can Arabic help out here?

What happens, if we drop Logicians *Dogmas* and apply *AntiDogmas* of Arabic, this time within the realm of Logics itself?

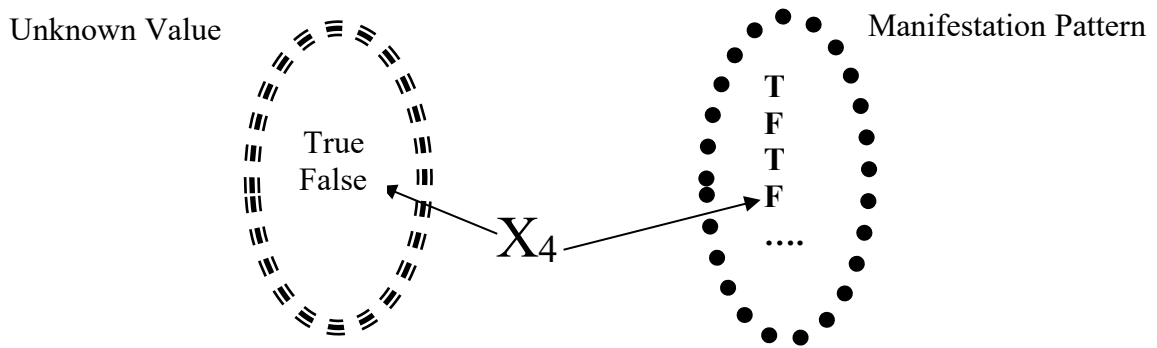

**Figure 1-c:** Semantic decomposition of a
Literal Name as per [40]

It turns out, that we gain significantly doing so:

1- As per *AntiDogma1* and *AntiDogma3*: The name of a Literal constitutes, similar to Noun- and Verb Symbols in Arabic, of two distinct *meta-Symbols* standing for two Semantic/cognitive entities *(i.e., two meaning-particles)*:

    a. *Meta-Symbol 'x'*, referring to *'the Unknown Value'* of a propositional Sentence*, which is not an object, but a cognitive pattern (singleton): *'True'* or *'False'*.

    b. The index *'i'* (in Figure (1-c) it is *'4'*), which refers to a sequence of such singletons, describing the way this particular unknown value manifests itself in the Truth-table (see *Observation-2* below).

2- Literal names, i.e.: 'x2' or 'x4' etc., remain referring to the above two constituent Semantic patterns, independent of the clause sets they are used in. They are hence: *Descriptive necessities of Propositional Logics.*

3- Practically: Both natures of a Literal *L* (in [11] called: The *'container'*- and the *'pattern'* nature), add generic information to what *Sat-Solvers* know, while instantiating *L*, namely: The relative order of the variable represented by *L* in the *Canonical Truth Table*, which is, as shall be seen below, vital from the point of view efficiency.

In the next sections, we aim to convince the reader that this simple Semantic Model, describing the relation between Syntax and Semantics of a propositional formula and inspired by Arabic, is not only practically feasible as a basis for *SAT*-solving methods, *but also successful.*

## ii.  <u>Using Binary Decision Diagrams to solve *kSAT* problems</u>

A *Binary Decision Diagram* (*BDD*) implementing a Boolean function f is a directed acyclic graph. The graph has two leaf nodes 0 and 1. Each non-leaf node is labeled with a Boolean variable v and has two out-going edges labeled 0 (the left edge) and 1 (the right edge). Every path in a *BDD* is unique, i.e., no two paths contain nodes with the same variables. If we arbitrarily trace out a path from the root node to the leaf node 1, then we have automatically found a value assignment to function variables for which the function will be 1, regardless of the values of the other variables. If any variable appears in a branch only once, the *BDD* is called *Free Binary Decision Diagram (FBDD)* [41].

The following generic procedure produces an *FBDD* for a set of *CNF* formulas representing a *kSAT* problem

*PR*:
Inputs: Arbitrary Clause Set S, where clauses contain at most k literals
Output: FBDD
Data Structure: Store of resolved Sets and their FBDDs (ST)

Steps:
Until all Literals are selected:
1- Select any Literal $x$, which was not chosen before, from a Clause C$\in$ S.
2- Put $x$ = TRUE in S forming S'
3- If (S' evaluates to TRUE)
      leftResult = TRUE-Node
      Else
      if (any C'$\in$ S' Evaluates to FALSE)
      leftResult = FALSE-Node
4- Put $x$=FALSE in S forming S''
5- If (S'' evaluates to TRUE)
      rightResult = TRUE-Node
      Else
      if (any C''$\in$ S'' Evaluates to FALSE)
      rightResult = FALSE-Node
6- Search for S' in ST if not S' TRUE/FALSE
      If found
            Put leftResult = FBDD of S' (create Common Node)
            Else
                - Put leftResult = PR(S')
                - Store S' as well as leftResult in ST
7- Search for S'' in ST if not S'' TRUE/FALSE
      If found
            Put rightResult = FBDD of S'' (create Common Node)
            Else
                - rightResult = PR(S'')
                - Store S'' as well as rightResult in ST
8- Create node Result such that: S is Clause Set of Result and:
a- leftNode(Result) = leftResult
b- rightNode(Result) = rightResult
9- Store S as well as Result in ST
10- Return Result

### iii.   Literal Ordering is NP-complete

Let us call the content of a stack which registers the Literal choices made by *PR* in step 1, while solving problem *p*: A *Variable Ordering*. As already mentioned: Figure (1-a) shows Ordering $\prod_p = [2<1<3<0<4]$ which makes the number of nodes generated in the final *FBDD* half the number needed, if we chose $\prod'_p = [0<1<2<3<4]$ of Figure (1-b). We call $\prod'_p$ *Canonical Ordering*, because it represents the order in which variables are listed from left to right in a *Canonical Truth Table*[36]:

| $x_0$ | $x_1$ | $x_2$ | $x_3$ | $x_4$ |
|---|---|---|---|---|
| 0 | 0 | 0 | 0 | 0 |
| 0 | 0 | 0 | 0 | 1 |
| 0 | 0 | 0 | 1 | 0 |
| 0 | 0 | 0 | 1 | 1 |
| 0 | 0 | 1 | 0 | 0 |
| 0 | 0 | 1 | 0 | 1 |
| 0 | 0 | 1 | 1 | 0 |
| ….. | | | | |

Canonical Truth Table – T2

A distinct feature of an *FBDD* is that Orderings chosen may be different for different branches. If only one Ordering is used for the whole graph, we call the resulting graph an *Ordered Binary Decision Diagram* (*OBDD*).

Since the number of possible Orderings may be very large even for a reasonable number of variables: Finding for a problem *p* an optimal Ordering $\prod_p$, i.e., one which enables the construction of minimal *BDDs*, is in general NP-complete [42].

### iv.   The link between CNF formulas, Truth Tables and BDDs

Because of their syntactical character: Truth tables and *BDDs* can obviously be regarded as extensional definitions of *'meaning'* of a *CNF*. They both explicitly contain all possible variable/value pairs against which a given *CNF* may be evaluated.

As per [12] two trivial, but contradicting observations can be made:

**Observation-1:** *It is possible to change any Ordering* $\prod_p$ *in a BDD to* a *canonical one* $\prod_p{}^c$ *by renaming variables in the canonical Truth Table.*

In the above example: Renaming $[x_2>x_0,\ x_3>x_2,\ x_0>x_3]$ makes the smaller *FBDD* achievable via a Canonical Ordering for $S^{Renamed}=\{\{x_0,x_1\}\{x_0,x_2\}\{x_3,x_4\}\}$, which is logically equivalent to *S* (as per *Consequence3* of *Dogma3*).

It follows from *Observation-1*, that we may focus our attention on the study of conditions, under which a Canonical Ordering produces *FBDDs* with small node counts, instead of searching in all ordering possibilities for suitable choices.

---

[36] We mean by a *'Canonical Truth Table'* one in which variable/value combinations are listed in an ordered way as depicted in T2.

A second intuitive observation is:

**Observation-2:** *Any Literal $x_i$ refers in the Canonical Truth Table to a variable, whose repetitive pattern of 0s and 1s has length $2^{N-i}$. This pattern is given by formula: $[2^{N-i-1}(0)2^{N-i-1}(1)]$, where N is the total number of variables. Negative literals refer to the same pattern-formula as positive ones, but with 1s preceding 0s.*

To fully appreciate this observation: A graph may be drawn, in which the x-axis represents rows of a *Canonical Truth Table* and the y-axis Boolean values given for a particular 2CNF formula *f*. This graph is called in [12]: *Pattern-Domain of f (PD$_f$)*.

Figure (1-d) shows for *Canonical Truth Table* T2: $PD_{\{x0,x4\}}$, $PD_{\{x2,x3\}}$, $PD_{\{x2\}}$, respectively. A Pattern Length Repetition of a variable *v* (*PLR$_v$*) is the number of times a Truth pattern of *v* is repeated within the $2^N$ rows of the Truth table. The Pattern Length Repetition of the Literal with the least index in a clause C/Clause Set S: *Pattern Length Repetition of C/S* (*PLR$_C$/PLR$_S$*).

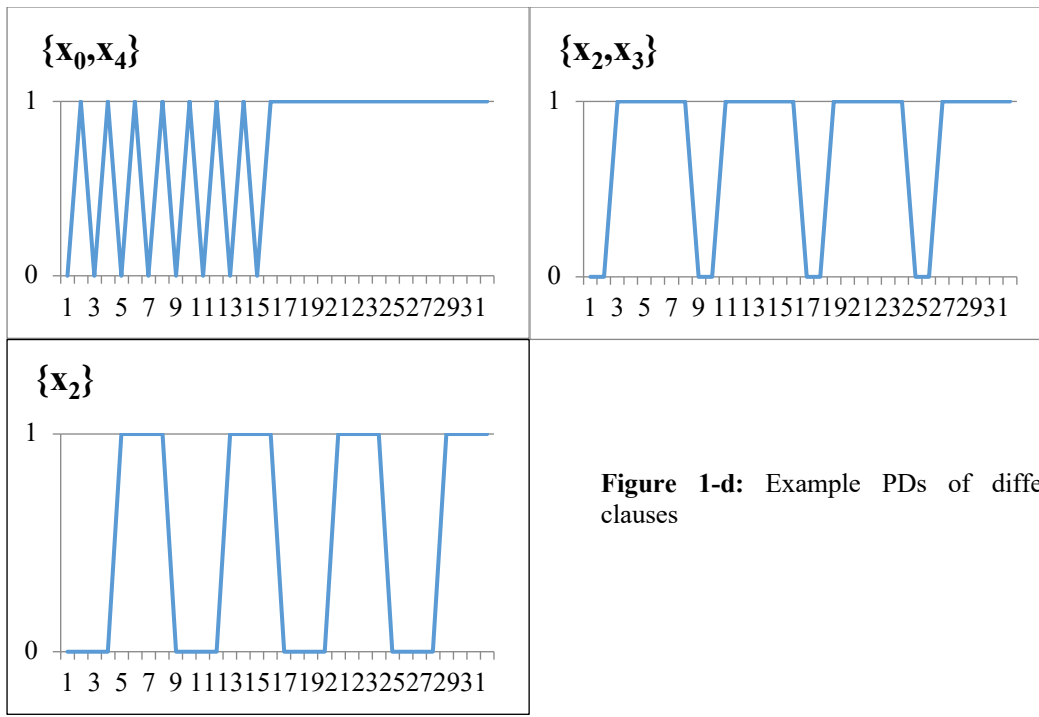

**Figure 1-d:** Example PDs of different clauses

Getting information from the index of a Literal goes against the *no-meaning postulate* of *Dogma3* and hence against *Observation-1* as well. It also contradicts *Dogma1*, according to which Literal Symbols refer to objects (*True* or *False*), not to patterns.

While *Dogmas* lead to such conceptual inconsistencies, *AntiDogmas* enable a natural and non-contradicting explanation of both observations: *Although any two indices: i, j in $X_i$ and $X_j$, $i \neq j$, refer to different patterns (Observation-2), all literals refer also to 'The Unknown Value', which is only one entity, representing a distinguished, clear concept. This is why substituting them for each other doesn't affect Truth (Observation-1).*

Before going into technical details of the algorithmic implementation of our ideas, we need to understand first what contribution *Observation-2* may have to the efficiency of Satisfaction procedures.

**v.   The Puzzle: Is there an efficient way to find an optimal Ordering of Literals?**

The reader must have noticed that we are dealing in this paper with at least two different types of NP-complete problems: On the one hand: *3SAT (in general: kSAT)*, the decision problem which Language recognition reduces to, and on the other: *the construction of minimal BDDs,* an optimization problem.

NP-completeness theory tells us, however, that they are, with respect to computability, two faces of the same coin, i.e.: The question of efficiently finding an optimal Ordering (leading to a minimal *BDD*) can be solved by finding sufficient and general conditions, under which *PR* generates small *FBDDs*, while solving *3SAT* (*kSAT*) problems. If we include *Observation-1*, we know also that such suitable conditions may, without loss of generality, relate to the use of Canonical Orderings only[37].

The *NP-Puzzle* can be reformulated thus in the following way: *What are suitable, syntactic conditions, which, when imposed on any CNF formula representing a 3SAT (kSAT) problem, enable PR, using a Canonical Ordering, to produce small FBDDs?*

Before moving further, we have to be clear about what we mean by *'small FBDDs produced by PR'*.

In the state-of-the-art literature, sufficient conditions for exponentially sized *BDDs* are given. In [43], for example, a Lower Bound technique, which is influenced by the algorithmic point of view following [44], is used to explain the methodology behind the majority of Lower Bound results known for some important functions like multiplication. As per [45], it turns out that variants of the following observation were constantly used:

*"**Lemma:** Let f be a Boolean function of n variables. Assume that m is an integer, $1 < m < n$, if for m any m-element subset Y of the variables N(f, Y) = $2^m$ holds[38], then the size of any read-once branching program (FBDD) computing f is at least $2^{m-1}$."*

In [45] a proof is given, in which it is shown that the sufficient condition of this **Lemma** leads to a complete binary tree in the first *(m-1)* levels of any *FBDD* computing f, so that we can formulate the following:

***Fact 1:** Let f be a Boolean function of n variables. If for m any m-element subset Y of the variables N(f, Y) = $2^m$ holds, $1 < m < n$, then: The first (m-1) levels of any FBDD computing f must be a complete binary tree.*

---

[37] In our first publication on the subject [10] we have shown that the Algorithm used for creating a compact *FBDD* for *3CNF* formulas can also be seen as an efficient 2-approximation Algorithm for *MinFBDD*. To solve *MinFBDD* in an exact way with the same Algorithm we need, however, to add to it the reduction between *MinFBDD* and *3SAT,* of course.

[38]  N(*f*,Y) denoting the number of different sub-functions obtained under all possible assignments to Y.

A small *FBDD* produced by *PR* is therefore: One which does not contain a complete binary tree for any non-constant number of variables:

**Fact 2:** *An FBDD produced by PR for the kCNF-representation of a Boolean Function f of N variables is exponential in size, relative to N iff there exist m variables, where m is not constant: $1 < m < N$, N = factor \* m, factor >= 1, such that: The FBDD contains a complete binary tree for those m variables.*

**Proof:** If an *FBDD* contains a complete binary tree for $m$, a non-constant number of variables $<= N$, then the total number of nodes in the *FBDD* would be $>= 2^{(N/factor)} - 1$. Other direction: If the *FBDD* is exponential in size, relative to $N$, but no non-constant number of variables form any complete binary tree within it, then the only way to add more nodes to the *FBDD* would be by adding more depth. This is because: Suppose some portion $p$: $p = m/c$, $c$ constant, out of the $m$ variables comes before the rest in the instantiation order. Then: Since $p$ don't form initially a complete binary tree, as per assumption, some of their value-combinations will remain missing, when the rest $m-p$ variables are instantiated in their turn. To cover this gap, variables in $p$ must be re-evaluated against each other again. This can only be achieved through selecting variables more than once, which contradicts the definitions of both *PR* and an *FBDD*. (QED)

## VIII-   Part C: Solving the *Puzzle* using AntiDogmas

### i.   Where the exponential behavior comes from

This part has two main objectives:

a-   Understand and formalize the conditions under which usage of Canonical Orderings produces small *FBDDs*, showing a practical Algorithm doing just that.

b-   Prove efficiency properties of the given Algorithm.

To get an intuitive understanding of what the proposed conditions may be, we focus our attention on constructing *FBDDs* for $S$ in the above example, only using Canonical Orderings.

More particularly: We would like to investigate node counts whenever one single clause is resolved against an *FBDD* constructed for the beginning of a Clause Set.

(Figure 1-e) shows two starting alternatives for $S$: $S'' = \{\{x_1, x_2\} \{x_2, x_3\}\}$ and $S''' = \{\{x_0, x_4\} \{x_1, x_2\}\}$. Node counts are clearly different. Remembering that Figure (1-b) depicted the *FBDD* for the whole $S$, we have two possibilities of node-count-growth from M=2 to M=3, where M is the number of clauses in $S$: From 4 ($S''$) to 10 or from 6 ($S'''$) to 10. In both cases we notice a blow-up of the number of nodes resulting from *'copying'* almost all of previously constructed nodes.

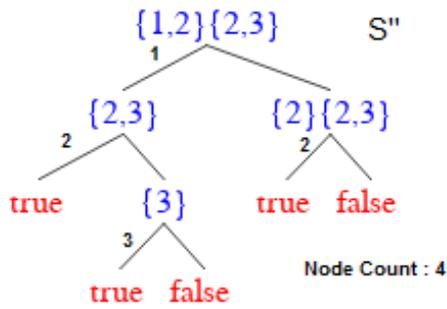

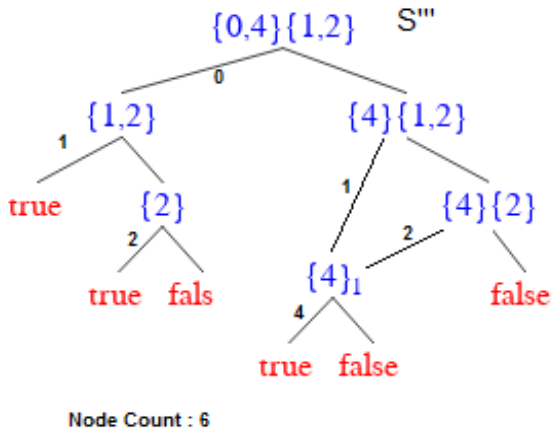

**Figure 1-e:** Starting alternatives

What about $S^{Renamed}$? Figure (1-f) shows a node-count-growth from 3 to only 5 in the *FBDDs* constructed for $S^{iv}=\{\{x_0,x_1\}\{x_0,x_2\}\}$ and $S^{Renamed}$, respectively.

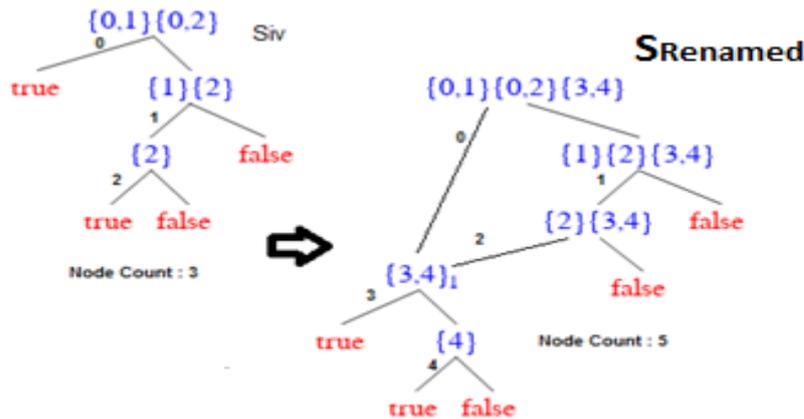

**Figure 1-f:** Smaller growth rate

Obviously, the nature of growth in the case of $S^{Renamed}$, is different: The full *FBDD* is constructed from the previous one by just adding two additional nodes to the lowest *FBDD*-level.

How can we explain this?

(Figure 1-g) shows $PD_{S''}$ and $PD_{\{x0,x4\}}$. The first one is for starting position $S''=\{\{x_1,x_2\}\{x_2,x_3\}\}$ Figure (1-e, top), the second for the clause which completes the tree to give the *FBDD* for $S=\{\{x_1,x_2\}\{x_2,x_3\}\{x_0,x_4\}\}$ (Figure 1-b).

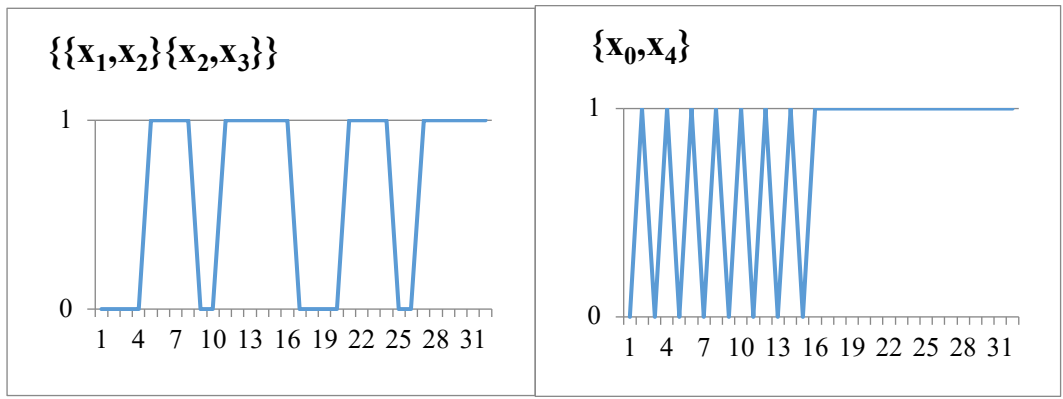

**Figure 1-g:** PD of an already processed Clause Set S'' compared to the PD of a new clause

As seen: $PD_{S''}$ consists of one self- repeating pattern $P_1$="0000111100111111", where $PLR_{S''}$ = 2 (i.e., $PD_{S''}$=2 $x$ $P_1$), $P_1$ representing the concatenation between sub-patterns for Clause Sets: {2}{2,3}="00001111"& {2,3}="00111111" seen in (Figure 1-e, top) .

When we want to resolve $P_1$ with $PD_{\{x0, x4\}}$ = $P_2$ & $P_3$, which has $PLR_{\{x0,x4\}}$=1, where $P_2$="0101010101010101", $P_3$="1111111111111111" as seen in (Figure 1-g, right), it is clear that we need $P_1$ to be bit-ANDed against each one of $P_2$ and $P_3$.

This explains why all nodes of the *FBDD* for *S''* had to be copied once as can be seen in (Figure 1-b). Clause {4} is appended there to all copies of nodes representing the result of bit-AND operation between $P_1$ and $P_2$. Obviously: Because $PLR_{\{x0,x4\}}$ < $PLR$ $_{S''}$ this Copy-Operation (called in [#2SAT is in P]: *Split*) was necessary.

What about *PDs* of the following (Figure 1-h) which relate to $S^{Renamed}$?

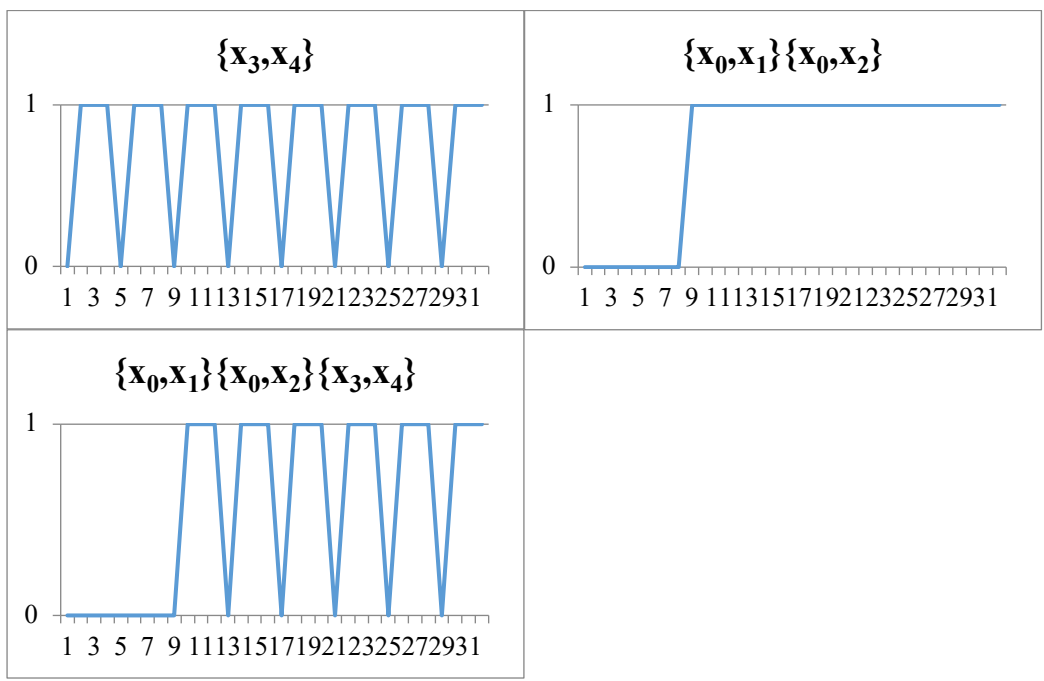

**Figure 1-h:** PD of an already processed Clause Set {{$x_0,x_1$}{$x_0,x_2$}} is bit-ANDed with PD of {$x_3,x_4$} to form PD of {{$x_0,x_1$}{$x_0,x_2$}{$x_3,x_4$}}

Here the new, to-be-resolved clause $C=\{x_3,x_4\}$ has $PD_C=8x"0111"$, $PLR_C=4$, while $PD_{\{0,1\}\{0,2\}}=(8x"0")$ & $(24x"1")$ is a pattern which repeats itself only once, i.e., $PLR_{\{0,1\}\{0,2\}}=1$. This gives us the opportunity to resolve the new incoming pattern of C with sub-patterns of $PD_{\{0,1\}\{0,2\}}$ only once and then refer to the result of this resolution whenever needed. This is reflected in the *FBDD* by including node {3,4} (Figure 1-f, bottom) as a *Common Node* between two constructed branches, thus reducing drastically the total amount of unique nodes.

Resuming this explanation: *It turns out that resolving a clause C with a Clause Set S, where $PLR_C < PLR_S$ necessitates Split-Operations. Such Operations are important causes of FBDD blow-ups. On the other hand: In the case of $S^{Renamed}$ resolving S with a clause C does not induce Splits, when $PLR_C > PLR_S$.*

In [12] we called this and three other conditions: *Linear Order (l.o.)*. The core of the work here is showing that Algorithms observing a slightly strengthened version of the l.o. condition *cannot* produce exponentially sized *FBDDs*.

## ii.  A Satisfaction Procedure

### 1.  The s.l.o. Condition

A *kCNF* formula set $S$ is called strongly linearly ordered (s.l.o.)[39] **if** the following conditions hold for Clauses $\mathbf{C}_i$, $\mathbf{C}_j$ and literals $\boldsymbol{a_x}$, $\boldsymbol{b_y}$, where $i, j$ are indices of clauses in S, x, y indices of literals:

- a-  $\forall a_x, b_y \in \mathbf{C}_i$: *if x<y then $a_x$ comes before $b_y$,* i.e., Literals are sorted in ascending order of indices within clauses.
- b-  $\forall a_x \in$**Literals of S**, $\forall \mathbf{C}_i \in$**S**: **if $a_x \notin$ LEFT($a_x$, $\mathbf{C}_i$) then $\forall b_y \in$ LEFT($a_x$, $\mathbf{C}_i$): $x > y$**, where LEFT($a_x$, $\mathbf{C}_i$) is the set of all literals appearing in S to the left of $a_x$ in clause $\mathbf{C}_i$ (all new indices of literals occurring for the first time in a clause of S are strictly greater than all the literal indices occurring before them in S).
- c-  $\forall \mathbf{C}_i$, $\mathbf{C}_j \in$**S**: **if size($\mathbf{C}_i$) < size($\mathbf{C}_j$) then $i < j$** (Shorter clauses come first).
- d-  $\forall \mathbf{C}_i$, $\mathbf{C}_j \in$**S**, $x$, $y$ **indices** *of* **head literals of $\mathbf{C}_i$, $\mathbf{C}_j$, respectively: if $x < y$ then $i < j$** (clauses are sorted in ascending order of head literals).
- e-  **Condition c- has priority over Condition d-.**

---

[39] In [12] we included prioritizing negative literals on positive ones in the l.o. condition, which is omitted here, and did not include sizes of clauses in the sorting process (as in condition c-).

## 2.  **Renaming Algorithm**

The Clauses Renaming Algorithm (CRA) as described in [12] is a procedure which takes an arbitrary Clause Set S as input[40], renames its literals yielding a new, logically equivalent S' as output. This procedure consists of the following steps:

**CRA:**
**Inputs**: Arbitrary kCNF Clause Set **S** of size M
**Output**: Clause Set S'
**Steps**: -
1. Enumerate clauses in S (starting with 0) in ascending order.
2. For each clause $C_i$:
   a) Arrange literals in ascending order[41] within $C_i$
   b) Create a matrix whose rows represent variable/Literal names/indices while columns represent clauses. This matrix is called: *Connection Matrix*.
3. For all clauses $C_i$ and all literals in $C_i$:
   - Create a new row and write column values TRUE or FALSE according to whether the Literal appears in the corresponding clause or not.
4. Rename all variables in the Connection Matrix in ascending order.
5. Reconstruct the clauses again using the new variable names. This reconstruction may be done by simply substituting each Literal in the original Clause Set with its new Literal name/index.

Example: If $S = \{\{0,5\} \{0,2\} \{1,3\} \{1,4\} \{2,3\}\}$, then the Connection Matrix of S is:

|   | $C_0$ | $C_1$ | $C_2$ | $C_3$ | $C_4$ |
|---|-------|-------|-------|-------|-------|
| 0 | **True** | **True** | False | False | False |
| 5 | **True** | False | False | False | False |
| 2 | False | **True** | False | False | **True** |
| 1 | False | False | **True** | **True** | False |
| 3 | False | False | **True** | False | **True** |
| 4 | False | False | False | **True** | False |

Transformed (via step 4 of CRA) to:

|   | $C_0$ | $C_1$ | $C_2$ | $C_3$ | $C_4$ |
|---|-------|-------|-------|-------|-------|
| **0** | **True** | **True** | False | False | False |
| **1** | **True** | False | False | False | False |
| **2** | False | **True** | False | False | **True** |
| **3** | False | False | **True** | **True** | False |
| **4** | False | False | **True** | False | **True** |
| **5** | False | False | False | **True** | False |

The new clause set for the above reads $S' = \{\{0,1\} \{0,2\} \{3,4\} \{3,5\} \{2,4\}\}$. Note that $S'$ is not s.l.o. (*condition d- is breached*), since one iteration is usually not sufficient. This can

---

[40] In [12] it was applied to *2SAT* problems, while here we use it for *kSAT*.
[41] In [12] the following extra phrase was added: *"such that literals which were not renamed before and appear more often in other clauses become head literals before those which appear less often or which only appear in $C_i$."*

also be seen using another example: $S = \{\{0,5\}\ \{0,2\}\ \{3\}\ \{1,4\}\ \{2,3\}\}$, where the above Connection Matrices look like this:

|   | $C_0$ | $C_1$ | $C_2$ | $C_3$ | $C_4$ |
|---|---|---|---|---|---|
| 0 | **True** | **True** | False | False | False |
| 5 | **True** | False | False | False | False |
| 2 | False | **True** | False | False | **True** |
| 3 | False | False | **True** | False | **True** |
| 1 | False | False | False | **True** | False |
| 4 | False | False | False | **True** | False |

|   | $C_0$ | $C_1$ | $C_2$ | $C_3$ | $C_4$ |
|---|---|---|---|---|---|
| **0** | **True** | **True** | False | False | False |
| **1** | **True** | False | False | False | False |
| **2** | False | **True** | False | False | **True** |
| **3** | False | False | **True** | False | **True** |
| **4** | False | False | False | **True** | False |
| **5** | False | False | False | **True** | False |

Yielding: $S' = \{\{0,1\}\ \{0,2\}\ \{3\}\ \{4,5\}\ \{2,3\}\}$, which is not s.l.o., because of condition c-.

### 3.  Renaming and Ordering Algorithm

As per [12] the Clauses Renaming & Ordering Algorithm (CRA⁺) is a procedure which takes an arbitrary Clause Set $S$ and applies CRA repetitively. After each step the intermediate Clause Set is sorted as required by the l.o. condition, before iterating back. This is done until renaming indices in two consecutive steps yields the same, i.e., the output Clause Set S' becomes l.o. The following recursive pseudo-formal Description of this procedure is an adoption for s.l.o.:

**CRA⁺:**
**Inputs**: An arbitrary kCNF Clause Set **S**
**Output**: s.l.o Clause Set S'

**Steps**:
  i- set currentSet = S,
  ii- while (currentSet is not s.l.o.)
        i.   currentSet = CRA(currentSet)
        ii.  sort currentSet as instructed in the s.l.o. condition
  iii- S'=currentSet
  iv- return S'.

Example: Continuing the procedure of last section for $S =\{\{0,5\}\{0,2\}\{3\}\{1,4\}\{2,3\}\}$, which resulted after one iteration of *CRA* in *currentSet* = $\{\{0,1\}\{0,2\}\{3\}\{4,5\}\{2,3\}\}$ lets us, as per step (ii), pull the unit clause to the beginning of the formula and move $\{2,3\}$ before $\{4,5\}$ to get: *currentSet* = $\{\{3\}\{0,1\}\{0,2\}\{2,3\}\{4,5\}\}$, for which we need another renaming iteration and the following matrices:

|   | $C_0$ | $C_1$ | $C_2$ | $C_3$ | $C_4$ |
|---|-------|-------|-------|-------|-------|
| 3 | **True** | False | False | **True** | False |
| 0 | False | **True** | **True** | False | False |
| 1 | False | **True** | False | False | False |
| 2 | False | False | **True** | **True** | False |
| 4 | False | False | False | False | **True** |
| 5 | False | False | False | False | **True** |

|   | $C_0$ | $C_1$ | $C_2$ | $C_3$ | $C_4$ |
|---|-------|-------|-------|-------|-------|
| 0 | **True** | False | False | **True** | False |
| 1 | False | **True** | **True** | False | False |
| 2 | False | **True** | False | False | False |
| 3 | False | False | **True** | **True** | False |
| 4 | False | False | False | False | **True** |
| 5 | False | False | False | False | **True** |

Which reads after renaming: *currentSet* = $\{\{0\}\{1,2\}\{1,3\}\{0,3\}\{4,5\}\}$ and is still not s.l.o., since we need to move $\{0,3\}$ to the second position: $\{\{0\}\{0,3\}\{1,2\}\{1,3\}\{4,5\}\}$, then iterate again:

|   | $C_0$ | $C_1$ | $C_2$ | $C_3$ | $C_4$ |
|---|-------|-------|-------|-------|-------|
| 0 | **True** | **True** | False | False | False |
| 3 | False | **True** | False | **True** | False |
| 1 | False | False | **True** | **True** | False |
| 2 | False | False | **True** | False | False |
| 4 | False | False | False | False | **True** |
| 5 | False | False | False | False | **True** |

|   | $C_0$ | $C_1$ | $C_2$ | $C_3$ | $C_4$ |
|---|-------|-------|-------|-------|-------|
| 0 | **True** | **True** | False | False | False |
| 1 | False | **True** | False | **True** | False |
| 2 | False | False | **True** | **True** | False |
| 3 | False | False | **True** | False | False |
| 4 | False | False | False | False | **True** |
| 5 | False | False | False | False | **True** |

Giving:     {{0}{0,1}{2,3}{1,2}{4,5}},     which     gets     sorted     to     become: {{0}{0,1}{1,2}{2,3}{4,5}}, an s.l.o. clause set[42].

### 4.  *PR'*

Putting all the above together in a procedure gives us the following alternative for *PR* (modifications are **bold** and <u>underlined</u>):

*PR'*:
Inputs: Arbitrary Clause Set S, where clauses contain at most k literals
Output: FBDD
Data Structure: Store of resolved Sets and their FBDDs (ST)

Steps:
**0-  <u>S = CRA[+](S), i.e., convert S to s.l.o.</u>**
**1-  <u>Use the Canonical Ordering</u>:**

    a. **<u>Select Literal x0</u>**
    b. Put $x0$ = TRUE in S forming S'
    c. If (S' evaluates to TRUE)
        leftResult = TRUE-Node
        Else
        if (any C'$\in$ S' Evaluates to FALSE)
            leftResult = FALSE-Node
    d. Put $x0$ = FALSE in S forming S''
    e. If (S'' evaluates to TRUE)
        rightResult = TRUE-Node
        Else

---

[42] We omitted the last iteration to avoid unnecessary length.

if (any C''∈ S'' Evaluates to FALSE)
    rightResult = FALSE-Node

f.  Search for S' in ST if not S' TRUE/FALSE
    If found
      Put leftResult = FBDD of S' (create Common Node)
    Else
      **Put leftResult = *PR′*(S')**
      Store S' as well as leftResult in ST

g.  Search for S'' in ST if not S'' TRUE/FALSE
    If found
      Put rightResult = FBDD of S'' (create Common Node)
    Else
      **rightResult = *PR′*(S'')**
      Store S'' as well as rightResult in ST

h.  Create node Result such that: S is Clause Set of Result and:
i.  leftNode(Result) = leftResult
j.  rightNode(Result) = rightResult
k.  Store S as well as Result in ST

10- Return Result

Figure 2 shows for the *3CNF* formula $S = \{\{0,1,2!\}\{1,3,4\}\{1!,5\}\{2,3\}\}$ an *FBDD* produced by *PR* (2-a) and another one produced by *PR′* (2-b).

The drastic improvement is apparent in the respective node counts.

In the next section we show that this improvement is not a coincidence and hides deeper properties of the new Satisfaction procedure.

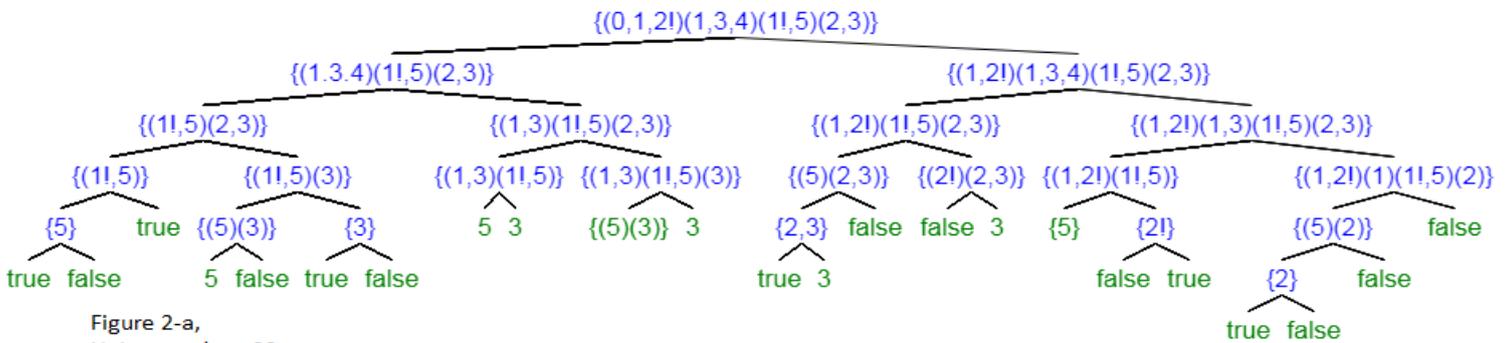

Figure 2-a,
Unique nodes : 22

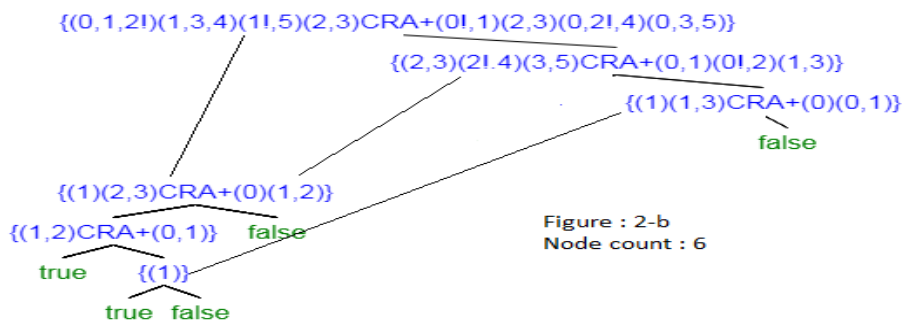

Figure : 2-b
Node count : 6

# I.  The non-exponential behavior

The reader may have noticed that s.l.o. (*condition b-*), i.e., the fact that new indices used by $CRA^+$ in literal names are always greater than all indices appearing before them, implements the inequality $PLR_C > PLR_S$ of section (i) of this part, according to which a clause $C$, occurring at the rear of a clause set $S$, should have a strictly larger $PLR$-value than the rest of $S$, in order for the *FBDD* to remain free of *Splits*.

We shall see in what follows that this condition, when coupled with sorting $S$ in ascending order of clause-sizes, s.l.o. *(condition c-)*, and sorting indices of head literals *(condition d-)*, while making sure that Literal indices appear within a clause sorted as well *(condition a-)*, suffices to guarantee non-exponential node counts.

We start with the *2SAT* case for better illustration of the ideas, then show the same behavior for *3SAT*. Note first the following conventions and facts:

i- Since $CRA^+$ is used in every recursive step, index $a$ may be renamed to become $a',a'',a''',...$ in one or more nodes. For simplification, we refer to it in all nodes as: $a^i$, similarly for: $b^j$ and $c^k$.

ii- $PR'$ uses Canonical Ordering to instantiate clauses. For indices $a$, $b$, $c$ of literals in any clause set, where $a<b<c$: $c$ is selected after both $a$ and $b$. Even when $CRA^+$ renames literals containing those indices in different nodes: Unless its relative position is changed due to sorting, $c^k$ remains fulfilling the inequalities: $a^i < c^k$ and $b^j < c^k$ for all i, j, k in all nodes, so that it comes *always* after the first two.

iii- Clauses containing $X_a$ as a head literal are called an: $X_a$-*block* (for example: $\{\{X_c,..\}\{X_c!, ..\}...\}$ is $X_c$-*block* in the below $S'$).

**Theorem 1:** Let $S$ be a clause set in *CNF* form, expressing a *2SAT* problem, then: The *FBDD* generated by $PR'$ for $S$ *cannot* contain any sub-graph in the form of a complete binary tree of depth $m >= 3$ for any $m$ variables.

**Proof:** Suppose this was not the case. Then, we would be able to find three variables, say $X_a$, $X_b$ and $X_c$, whose literals appear as head literals in $S$, such that $PR'$ forms, while instantiating their clauses, a complete binary tree within the resulting *FBDD*[43].

Let $S' = CRA^+(S)$ be the s.l.o. set resulting after step 0- of $PR'$ and:

$S' = \{\{X_a , ..\}\{X_a! , ..\}.. \{X_b , ..\}\{X_b , ..\} ..\{X_c , ..\}\{X_c! , ..\} ...\}$ for example.

$X_c$ or its negation cannot appear neither as a head- nor as tail literal in any clauses in a later $X_d^j$ –block, where $c<d$, because of s.l.o. conditions a- and d-.

We can thus distinguish only two cases:

**_Case1_:** $X_c$ appears for the first time in $S'$ as a head literal of clause $\{X_c , ..\}$.

---

[43] Obviously: Variables appearing only as tail literals do not contribute to the formation of complete binary trees because those literals form, after instantiation of their corresponding head literals, unit clauses.

Any indices coming before $c^k$ in any node must therefore be $< c^k$, i.e., $c^k$ cannot change its relative position and come before such indices. As per (ii) above: $X_c^k$ –block is instantiated with values of $X_c$ always *after* $X_a^i$ –and $X_b^j$ –blocks. Because values of $X_a^i$ and $X_b^j$ do not affect any clauses in the $X_c^k$-block, which can only contain tail literals with indices $> c^k$ as per s.l.o. (condition a-): $X_c^k$ *-block appears in its entirety in all nodes produced in the FBDD prior to instantiating literal $X_c^k$.*

It forms a *Common Node*, acting as a sink within the *FBDD* (as seen in Figure-3), blocking any effort to extend the sub-graph resulting from the prior instantiation of $X_a^i$- and $X_b^j$– blocks to a complete binary tree for all three variables.

***Case2*:** $X_c$ appears for the first time in *S′* as a tail literal before the $X_c^k$-block as in:

$S′ = \{\{X_a, ..\} \{X_a! , \boldsymbol{X_c}\}..\{X_b, ..\} \{X_b, ..\}..\{\boldsymbol{X_c , ..}\}\{\boldsymbol{X_c! , ..}\}…\}$.

Or

$S′ = \{\{X_a, ..\} \{X_a! , .. \}..\{X_b, \boldsymbol{X_c}\} \{X_b, ..\}..\{\boldsymbol{X_c ,..}\}\{\boldsymbol{X_c! , ..}\}…\}$.

Here again it is clear (illustration in Figure-4) that we cannot extend any sub-graph already created for the first two variables to a complete binary tree for all three variables, since in any tree generated for $X_c^k$ –block: *Unit clauses resulting from previous steps (like in: {..{$X_c$}..{$X_c$ , ..}{$X_c!$ , ..}…}) are prioritized by PR′, as per s.l.o. (condition c-), thus creating a graph in which either the right- or the left-side is missing (i.e., having the value 'false').*

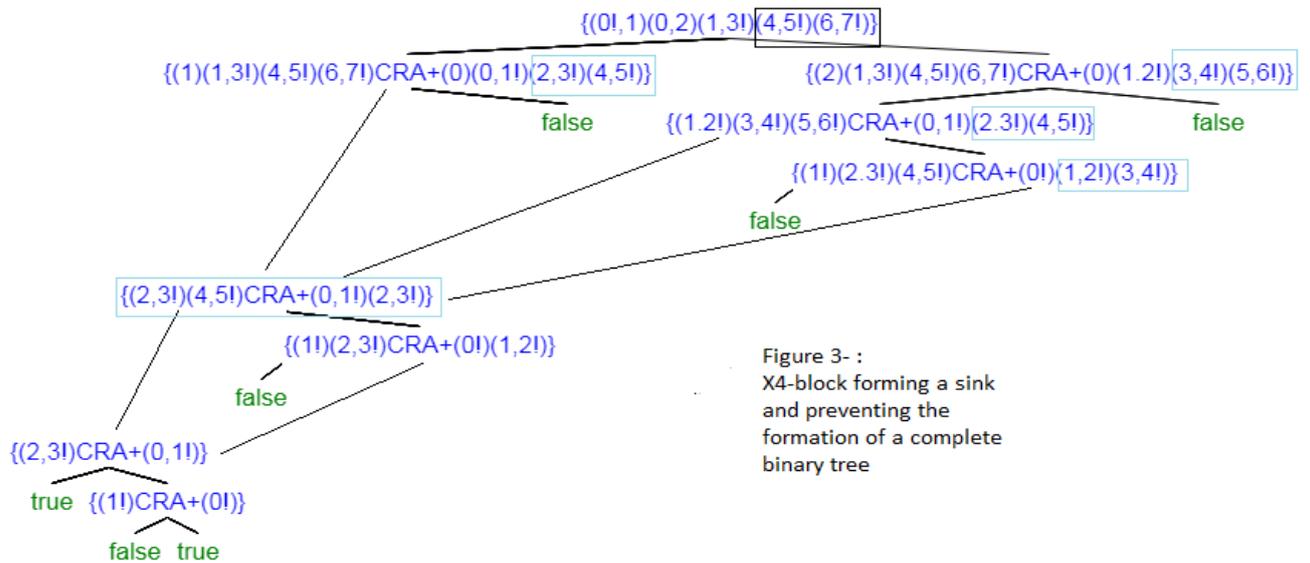

Figure 3- :
X4-block forming a sink and preventing the formation of a complete binary tree

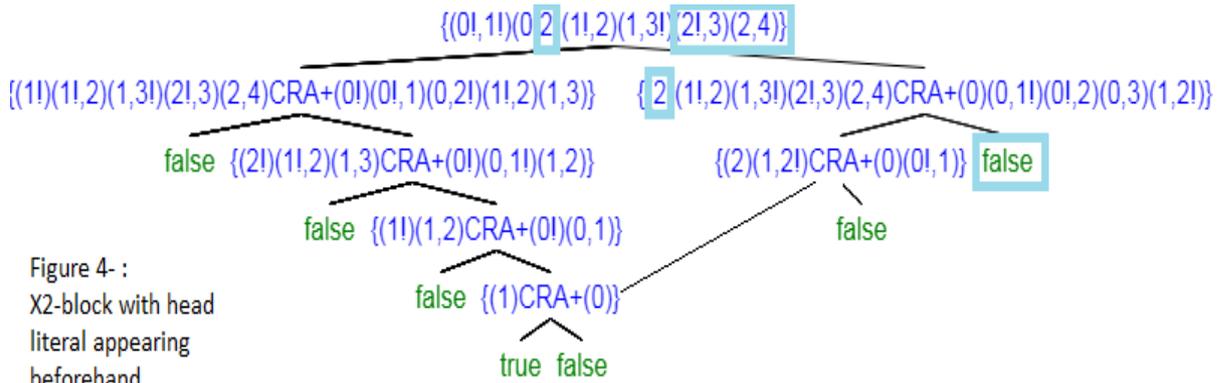

Figure 4- :
X2-block with head
literal appearing
beforehand

(QED)

**Theorem 2:** Let $S$ be a clause set in *3CNF* form, expressing a *3SAT* problem, then: The *FBDD* generated by *PR′* for $S$ *cannot* possess any sub-graph in the form of a complete binary tree of depth $m >= 4$ for any $m$ variables.

**Proof:** The same arguments used in Theorem 1 hold for the *3SAT* case, if for any three chosen variables for the construction of a complete binary tree: $X_a$, $X_b$ and $X_c$, literals of $X_c$ either appear for the first time as head literals of $X_c^k$ –blocks or as tail literals before that.

The only new case we need to investigate is the one, where $X_c^k$ appears as a *middle-literal* of a clause *before $X_c^k$*–block as in:

$$S' = \{\{X_a,...\}\{X_a!, X_c, X_i\}..\{X_b,...\}\{X_b...\}..\{X_c,...\}..\},$$

for example (see Figure-5)[44].

In that case: We notice that the instantiation of $X_c^k$–blocks, in which $X_c^k$ appears as head literal of two-sized clauses, like in: $\{... \{X_c^k, X_i^m\} ..\{X_c^k,..\}..\{X_c^k!, ..\}...\}$ is the result of putting $[X_a = \text{true}]$ in $\{X_a!, X_c, X_i\}$ and has to lead in a further step to one branch, in which the unit clause $\{X_i^m\}$ comes first. Putting $[X_a = \text{false}]$ lets the whole clause disappear.

If, for any reason, *PR′* chooses a Literal other than $X_i^m$ to be instantiated after $X_c^k$, then those literals must be forming unit clauses as well, because of s.l.o. (condition c-). *PR′* moves unit clauses always to the beginning, so that any instantiation must result in a tree evaluated to *'false'* in one of its branches (see *'false'*-box in Figure-5).

This means that even if a complete binary tree is generated for all three variables after the instantiation of $X_c^k$–blocks: *PR′ cannot extend the depth of this complete tree into the next level, because at least one of its branches will be missing.*

(QED)

---

[44] Here also, $X_c^k$ or its negation cannot appear anywhere in clauses in a later $X_d^l$–block, where $c<d$, because of s.l.o. (conditions a-, d-).

Figure 5 : An example
of an X3-block whose
head literal '3' appears
as a middle-literal as
well

## II.  <u>What about Lower Bounds?</u>

Showing that *PR′* produces small *FBDDs* provides more evidence, supporting the results in [10] and [12], according to which similar pattern-oriented procedures have a worst-case number of nodes given by $O(M^4)$ *only*, produced in a polynomial number of steps in *M*, where *M* is the number of clauses in a *3CNF/2CNF* clause set, which makes *P*=*NP*[45].

As per [12], this result has been shown in three different ways:

i.   *FBDDs* of polynomial sizes for arbitrary *2CNF* formulas enable the definition of efficient Model counting solutions resulting in solving *#2SAT*, the *#P-complete* problem in a polynomial number of steps (Theorem 1- b).

ii.  Uniformly linking efficient *1CNF*- and *2CNF*-versions of pattern-algorithms, while proving small, upper bounds on unique node counts for those two base cases, enables formulating the strongest possible induction hypothesis, namely: That there is a polynomial time Algorithm producing polynomial number of unique nodes in an *FBDD* (which means: *P*=*NP*). This in its turn facilitates using the described *kSAT-FGPRA* to solve *(k+1)CNF* formulas via equisatisfiable translations in the induction step, completing thus a second way of showing that *P*=*NP* (Theorem1-a).

iii. A third way was shown in [10] by directly solving *3CNF* using procedures, which are guaranteed to produce also *2-approximations* of yet another NP-complete problem, namely: *MinFBDD*.

In all three publications, the existence of known Lower Bounds was taken up as a challenge, against which our results were compared. This was successfully done in [12] and [10] and is repeated here from a different angle:

There is a common Lower Bounds technique used in Communication Complexity, which makes use of the concept of *fooling sets*. With the help of fooling sets, a Lemma similar to the one of section (v) (part B) helps obtaining

---

[45] Until the time of completion of this paper, this result hasn't been recognized, despite the fact, that all three previous publications [10], [11], and [12] were not only peer-reviewed by mathematicians and availed online in popular public archives since 2016, but also investigated by a German scientific committee, formed by German administration under German Secrecy Act §93 StGB, which failed to report a single mistake in reasoning or proofs. This shouldn't come as a surprise, though, knowing that trivial flaws of Logicians, like the ones shown here, and their contrast to known facts from Arabic, needed more than one hundred years to come to light in this work as well.

Lower Bounds. Instead of arguing over all possible variable orderings and/or sub-functions, one can argue over balanced partitions.

**Definition** (see [46]): A Fooling Set for a Boolean function $f$ and a balanced partition (L,R) is a set A(L,R) which contains pairs $(l,r)$ of assignments. For two different pairs $(l,r)$ and $(l',r')$, it has to hold $f(l \cdot r) \neq f(l' \cdot r)$, where $(l \cdot r)$ denotes the complete assignments resulting from assignments l and r.

**Lemma (Fooling Set):** If there is a fooling set with size $c^n$ and$>1$ for every balanced partition (L,R) for a function $f$, then every BDD representing the function has a size of at least $O(c^n)$ .

While the full proof can be read in [47], an intuition on why this Lemma is correct is given in [48]:

*"Given an arbitrary variable ordering, there is the corresponding balanced partition (L, R). Directly below the last variable from L, **the width of the BDD has to be at least as wide as the size of the fooling set** [...] Otherwise, there would be l and l' from the fooling set leading to the same node at this level of the BDD, which would result in a violation of the fooling set. It is due to the fact that no r will be able to produce a different result for these l and l'. Consequently, the complete BDD has to be bigger than the fooling set."*

As per [48], the following Fact, very similar to **Fact 1** of section (v) (part B), is obvious then:

**Fact 3:** *If there is a fooling set of size $2^m$, for every balanced partition (L, R) of a Boolean Function f of N variables, $1<m<N$, then any BDD calculating f contains a complete binary tree of depth m+1 in its first levels.*

Recall: In **Fact 1** we saw that, if for an integer $m$ and any $m$-element subset $Y$ of the variables of a Boolean Function $f$ of $n$ variables, where $1 < m < N$, the number of different sub-functions obtained under all possible assignments to $Y$ is $2^m$, then the first $(m-1)$ levels of any *FBDD* computing $f$ must be a complete binary tree.

Lower Bound techniques seem thus to build upon the idea, that producing *BDDs* for particular Boolean Functions, possessing some non-constant subsets of variables, whose full combinatory variable/value-extensions need to be included in the search for validating assignments (as prescribed by the Semantics of those functions), *necessitates* forming complete binary trees in the first levels of *BDDs*.

We have just seen in *Theorem 2* of last section that any *FBDD* generated by *PR'* for a *3CNF* clause set *S*, *cannot possess such complete subtrees for any non-constant depth.*

How is this compatible with Lower Bound results known for many practical Boolean Functions, like multiplication for example?

Note that *Theorem 2* would only be in *direct* contrast with Lower Bounds on sizes of *FBDDs*, obtained for *3CNF*-representations of Boolean Functions resolved via Satisfaction procedures.

Moreover: When converting a *kCNF*-representation of a function to the *3CNF*-presentation, reductions usually use *equisatisfiable* transformations, i.e., the given Boolean function *f* is translated into another function *f'*, which is not equivalent, but always satisfied, when f is satisfied and vice versa [49].

A thorough study of literature reveals that no Lower Bound results were reported, neither for *3CNF*-representations nor for equisatisfiable transformations of Boolean Functions.

No surprise there: Combinatory insights about how functions behave can only come from studies of the behavior of the functions themselves, not equisatisfiable versions of them, *which are not expected to behave the same way in the first place.*

Can reported Lower Bounds indirectly contradict our findings?

This can happen, if either the Lower Bound proof assumptions are also applicable to our methods or the investigated types of *FBDDs* in the Lower Bound results are shown to be equivalent to the ones we produce:

Let *f* be a Boolean Function for which a Lower Bound *LB* on the size of the *FBDD* is known, *f'* an equisatisfiable *3CNF* formulation of *f*.

The reasons why *LB* isn't applicable to *f'* has been informally summarized in [12], we mention here the main points again:

1- The *k*CNF Description of *f* is sometimes the *only way* to guarantee that, for any *m*-element subset *Y* of the input variables of *f*, different sub-functions obtained under all possible assignments to *Y* are truly distinct.

For example, in the projective planes case we quote the following part of the Lower Bound proof [45], page 15:

"*Proof of the theorem. We show that for every q-element subset A of the variables, $N(f_{\Pi}, A) = 2^q$ holds, i.e., each Truth assignment to the variables in A yields a different sub-function on the remaining variables.* **Since each line defines** a **clause of the function** $f_{\Pi}$, *it follows from the Fact[46] that for an arbitrary q-element subset A of the variables there exist q clauses such that each variable from A appears in exactly one of them, and each variable appears in* a *different clause.*"

Obviously: When *f* is formalized in *3CNF*, a line for projective planes with *q>3* cannot be represented by only one clause making the above argument inapplicable.

2- From the logical point of view, *f* and *f'* are *not* equivalent. This means that *FBDDs* constructed for them are not expected to be equivalent as well. There may be Models for *f* which are not Models for $f$ and vice versa. As $f$ and $f'$ are equisatisfiable, they may disagree for a particular choice of variables.

---

[46] *Lemma* of section (v) (part B) could only be applied to the blocking Sets problem, because of the following combinatory property shown to hold for projective planes [45]: *"Fact: Let J={p1,...,pt} be a set of t<=m distinct points of the projective plane P, then there exist distinct lines {l1,...lt} such that for i>=1, j <=t we have pi ∈lj iff i=j."*

A typical equisatisfiable translation from *kCNF* to *3CNF* usually looks like:

$$C' = (A \lor B \lor x_1) \land (\neg x_1 \lor C \lor x_2) \land (\neg x_2 \lor D \lor E)$$

for a $k$=5 clause:

$$C = (A \lor B \lor C \lor D \lor E),$$

for example.

Note that while C has a Model in which B=TRUE, $x_2$=TRUE and all other variables including $x_1$ are FALSE, this is not a Model for the translated *3CNF* formula C'.

In such constellations: The number of variables in clauses of $f'$ are strictly larger than the number of variables in clauses of $f$ and consequently: Sub-function properties, necessary for the application of the combinatory *Lemma* of section (v) (part B) are disturbed by the introduction of new variables[47], which have no place in the definition of $f$ and must be treated as *Don't Cares*, i.e., variables whose Truth values don't matter for the overall Truth-value of the formula. Treating variables as *Don't Cares* makes the *FBDD* Non-Deterministic, causing all Lower Bounds for Deterministic *FBDDs* to be inapplicable per definition[48].

3- If *LB* is a Lower Bound on the size of any Non-Deterministic FBDD[49] constructed for $f$, as the one given in [50] for example, not necessarily using *Lemma* of section (v) (part B), call a Non-Deterministic *FBDD*: *NFBDD* and a Deterministic *FBDD*: *DFBDD*, then: For *LB* to be applicable to $f'$, the following must be true:

   "*An NFBDD satisfying f can be transformed into a DFBDD satisfying f'*"[50].

But this is not the case:

One might think that erasing variable names from nodes in a *DFBDD* for $f'$ is enough to transform it into an *NFBDD* for $f$, or that putting a proper marking on nodes in a *NFBDD* of *f* can give us a correct *DFBDD* for *f'*.

---

[47] Similar arguments hold for the fooling set Lemma as well.

[48] It must be mentioned here that introducing new variables is known, since the 90s, to disturb exponential Lower Bounds obtained for multiplication-*BDDs*, for example. In [51] a method for using *BDDs* to verify multipliers while avoiding exponential complexity is shown. Normally the outputs of an $n$ by $n$ bit multiplier circuit are represented by *BDDs* with $2n$ variables, since the circuit has $2n$ inputs. In the method described there, the outputs of the circuit are represented by a *BDD* with $2n^2$ variables, instead. The size of this *BDD* is cubic in $n$.

[49] Recall: A Deterministic *FBDD* is an *FBDD* in which every node is marked with a variable name, while a Non-Deterministic *FBDD* has some unmarked nodes [52].

[50] Note that if $f$ and $f'$ are equivalent, agreeing on all used variables, this is trivially true.

Both ways are not possible as can be seen in the following simple counter-example using *f* = C, *f'* = C' of point 2. (Figure 6-a, -b):

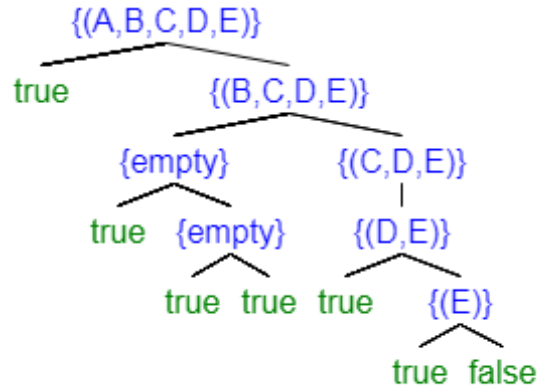

Figure 6a-: NFBDD for {A,B,C,D,E}

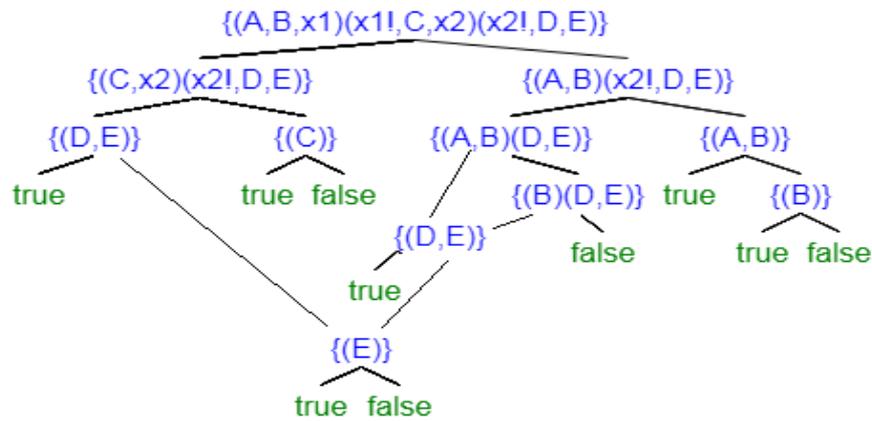

Figure 6b- : DFBDD for : $(A \lor B \lor x_1) \land (\neg x_1 \lor C \lor x_2) \land (\neg x_2 \lor D \lor E)$

If we mark *'empty'*-labeled nodes in Figure 6a- with $x_1$ or $x_2$, in any order, and try to use the resulting *DFBDD* for *f'*, it tells us (because all leaf nodes are TRUE starting from the point of insertion of *'empty'*-labeled nodes), that it is sufficient to set: B=TRUE, $x_2$=TRUE to get *f'* satisfied, irrespective of values put for the rest of the variables, which is wrong, since *f'* can become FALSE when all other variables are set to FALSE as we have seen above.

Failing from the other side is also easy to see: Erasing markings from the first two levels of the *DFBDD* in Figure 6b- (which represent variables $x_1$ and $x_2$ in this order), so that it can be used as an *NFBDD* for *f*, will give us a branch where setting C=FALSE renders *f* FALSE, irrespective of values of other variables, which is wrong as well. Even if we try writing on the erased markings the names of any variables, other than C, like A or B for example, then the *DFBDD* will *not* reflect correctly what happens when A or B are set to TRUE in *f* (which must lead to TRUE nodes immediately as in Figure 6a-).

Resuming our arguments showing that Lower Bound results known in the literature *cannot* contradict our findings here:

1- Only Lower Bounds obtained for *3CNF* representations of Boolean Functions and processed using Satisfaction-procedures can directly contradict our findings. We are not aware of such results[51].

2- Lower Bounds obtained for *kCNF*-representations of Boolean Functions are non-contradictory as well, because:

   a. Combinatory Lemmas used may not work for equisatisfiable translations, like the ones used in *3CNF* representations (the case of projective planes).

   b. Even assuming such Lemmas do work for some cases: Since functions and their equisatisfiable counterparts are *not equivalent*, *BDDs* (Deterministic or Non-Deterministic) constructed for them are not equivalent as well and hence: *Not comparable*.

## III.  Experimental Evidence: Multiplication Circuits

Since the publication of [10], in which we presented for the first time the family of pattern-oriented SAT-Solving Algorithms, continuous development- and reproducible testing efforts of our methodologies[52] have succeeded in providing preliminary validation results of the new theory [53]. We call those validation efforts: *'preliminary'*, because they involved testing our algorithms only on instances of small, hard problems[53], known to pose difficulties for most Sat-Solvers, especially those, which produce *BDDs*.

In spite of this restriction: *We were able to obtain, among other forms of evidence, that, as predicted in the last section, multiplication Lower Bounds don't apply to our algorithms.*

We selected for our tests *3CNF*-Clause Sets as provided by the well-known Indiana universities *CNF* generator [54], which is a tool producing X-bit-length integer factorization and multiplication circuits.

To ensure that no trivial factors are found for prime numbers, Indiana circuits require, for *n-bit* inputs, that the first factor has no more than *(n-1) bits* and that the second factor has no more than *n/2* (rounded up) *bits*.

---

[51] All Lower Bounds known to exist for multiplication, for example, were *not* obtained for Satisfaction procedures using *CNF*-representations of this function, but for direct bit-wise multiplication.

[52] This information is published with the kind permission of *GridSAT Stiftung*, a German non-profit foundation. All results are reproducible by obtaining an open-source license from *GridSAT* and downloading open-source programs implementing our Algorithms, and then applying them to the indicated *CNF* clause sets. Clause Sets can be obtained via email-request to the author or through, e.g., Indiana-university online *CNF* Generator.

[53] *Small*: Number of Variables and/or Clauses in *Thousands*.

This particularity and the fact, that any generated *3CNF*-Clause Set represents one single circuit, makes multiplication using this method (below: *ibit-multiplication*) dependent on the interval chosen: Numbers are multiplied only using the smallest circuits, in which they can be represented. For example: 10-bit numbers are multiplied using a 10-*ibit* multiplier, not a 22-*ibit* multiplier.

One nice property of Indiana's factorization *CNF* generator is the possibility to generate *CNFs* for multiplication from factorization instances. This is simply done by discarding unit-clauses generated at the end of any predicate. If the original *CNF*-formula represented factorization of *X-ibit* integers, the one without unit-clauses represents multiplication of two integers of different lengths, both smaller than *X-ibit*.

The following node count values were recorded for the multiplication of integer numbers from: 4- to 23-*ibits* of length (14 readings, Table: 3): By increasing the number of *ibits* from 4- to 23, using N-Bit, Carry-Save multiplication (variables from 12 to 658), the resulting sizes of *FBDDs* increased from 50- to 53104412 unique nodes.

The validation rationale we used was simple: *As per known Lower Bound results, an exponential blow-up in the sizes of FBDDs must be observable, even when the distance between consecutive ibit-lengths is not more than hundreds of variables only. When those distances grow, corresponding growth-factors of sizes of FBDDs must also grow.*

Let the size of an *FBDD* for an *X-ibit* circuit, using $n_1$ variables in its *3CNF*, be $size_1$, then its exponential growth-factor, which is the ratio between the size of the *FBDD* of the next *(X+1)-ibits* circuit, using $n_2$ variables (called: $size_2$), and $size_1$ must be given by:

- *$Size_2 / size_1 = base^{n_2 - n_1}$, for an estimated base $> 1$*

The reader can clearly see in Table T4 (also illustrated in Figure 7) that: *The estimated exponential base becomes smaller and smaller with larger and larger variable distances. It may be expected to practically diminish at a relatively low number of ibits.* This evidence contradicts the expectation in the validation rationale and is, hence, sufficient to show the inexistence of exponential Lower Bounds, when applying our methods to Indiana-type multiplication circuits.

| Ibits | N | <u>Unique nodes</u> |
|---|---|---|
| 4 | 12 | 50 |
| 5 | 22 | 139 |
| 7 | 52 | 1303 |
| 8 | 68 | 2389 |
| 10 | 116 | 17806 |
| 11 | 138 | 26475 |
| 13 | 204 | 166407 |
| 14 | 232 | 219374 |
| 16 | 316 | 1236655 |
| 17 | 350 | 1519226 |
| 19 | 452 | 7946916 |
| 20 | 492 | 9345347 |
| 22 | 612 | 46489244 |
| 23 | 658 | 53104412 |

Sizes of FBDDs for *ibit* Multiplication – T3

| ibits | <u>distance in variables</u> | <u>estimates for exponential base</u> | <u>growth factor of bdd</u> |
|---|---|---|---|
| 5 | 10 | 1.1076 | 2.78 |
| 7 | 30 | 1.077 | 9.374100719 |
| 8 | 16 | 1.039 | 1.833461243 |
| 10 | 48 | 1.043 | 7.453327752 |
| 11 | 22 | 1.018 | 1.486858362 |
| 13 | 66 | 1.028 | 6.285439093 |
| 14 | 28 | 1.01 | 1.318297908 |
| 16 | 84 | 1.02 | 5.637199486 |
| 17 | 34 | 1.006 | 1.228496226 |
| 19 | 102 | 1.016 | 5.230897839 |
| 20 | 40 | 1.004 | 1.175971534 |
| 22 | 120 | 1.013 | 4.974587246 |
| 23 | 46 | 1.003 | 1.142294592 |

Estimation of Exponential bases for *ibit* Multiplication – T4

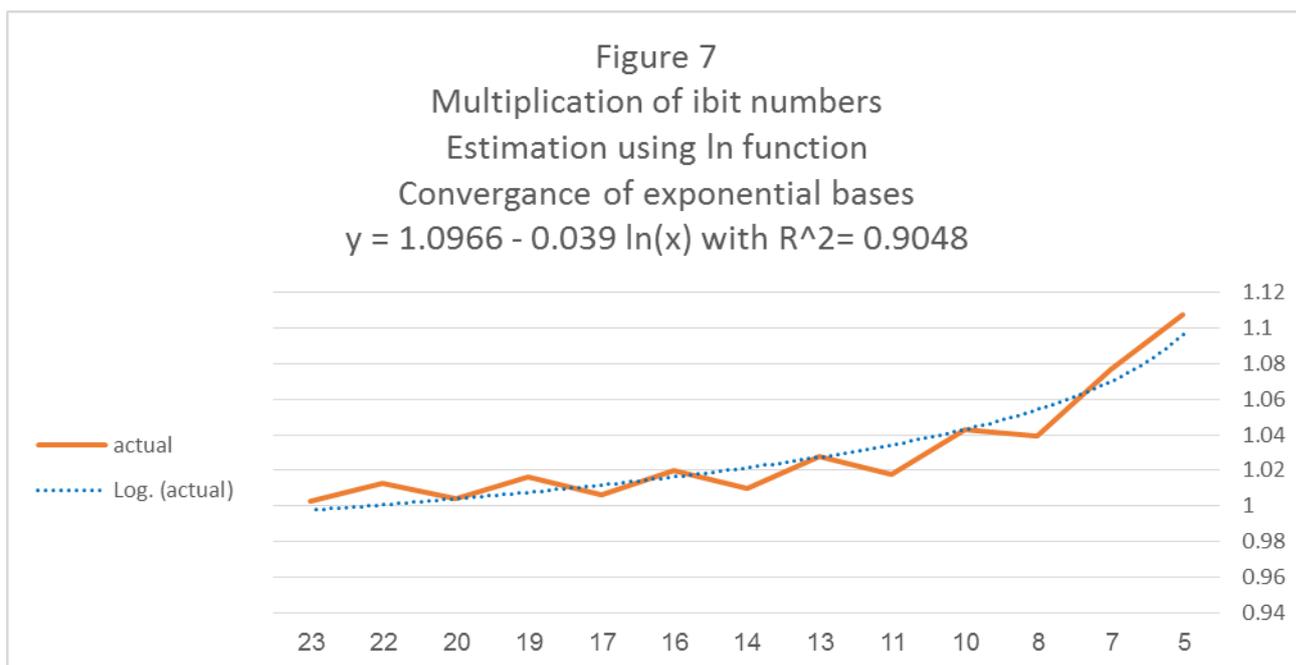

Figure 7
Multiplication of ibit numbers
Estimation using ln function
Convergance of exponential bases
y = 1.0966 - 0.039 ln(x) with R^2= 0.9048

## V-     Discussion of results and future work

The relation between Syntax and Semantics in formal and Natural Languages is one of the most debated topics in modern Logics and Linguistics.

Since its beginning, however, this debate was *indoctrinated*: Western Logicians and Philosophers insisted on enforcing their logical views on Natural Language, not only ignoring counter-evidence from the latter, but also falsely re-interpreting Language mechanisms to reflect their logical postulates, claiming, for example, that surface structures were *'deceiving'* and hide behind them an alleged, deeper logical layer, which, after decades of research, turned out to be superfluous as we know now.

Result of this indoctrination was the false, but unfortunately commonly accepted pretense, that Natural Language is imprecise and/or inadequate to be seen as a formal System, a view shared by almost all founding fathers of mathematical Logics: Frege, Russell, Tarski, Quine, Wittgenstein and many more.

One could have expected a revision of such Doctrines, when, by the middle of last century, the generative Grammar movement emerged and Chomsky's work falsified the claim, that Language is not amenable to formalization. But this was too much to ask. *Doctrines were already deeply embedded in working Models of Linguistics, Logics and Computation, so that they became: Dogmas.*

One could have even expected Dogmas, if claimed to be applicable to all Natural Language contexts, to compare to founding principles of at least one Natural Language, which played a crucial role in the philosophical, scientific and technological enlightenment of western civilization, namely: Arabic. This was also too much to ask, knowing what deliberate contrast to science and rationality Arabic and Islam were falsely accused of.

There was and still is in the state-of-the-art of Computational Linguistics an unforgivable absence of formal investigations of the Semantics of Arabic, according to Arabic's own principles, not to any imported ones.

Victim of such subjective attitudes and such shameful absence is, as always: *The scientific, objective Truth:*

The three Dogmas of relevance in this work consist of assuming that Symbols refer solely to objects of the world (*Dogma1*) and this reference is uniquely determined by the context of a Sentence, not by any intrinsic features of the used Symbols (*Dogma3*). Descriptions are allegedly incomplete Symbols, which need to be substituted by existential assertions to reflect logical meaning (*Dogma2*).

Traditional Philosophy of Classical Arabic contradicts all three Dogmas: Symbols refer to meanings not to things (*AntiDogma1*), *'meanings'* being cognitive patterns, not necessarily manifested in objects. They are embedded in permutations of characters used in Symbols. A Symbol contains its own, interpretation-independent Semantic nuance, which is only complemented by the context of usage, but never erased or nullified (*AntiDogma3*). Descriptions, whether Definite or Indefinite refer to an imaginary entity, representing a mental concept, existing per se and not necessary holding any logical attributes (*AntiDogma2*).

Calling undisputed principles of Arabic: *Anti-Dogmas* serves the sole purpose of comparing and contrasting them to Logicians ideas.

Fact is, however, that there is a substantial epistemological difference between both types of postulations: While Arabic *Anti-Dogmas* represent overwhelming linguistic evidence, known from Arabic for millennia now, *Dogmas* of Logicians contradict all such evidence:

1- Alleged *Identity*- and *Substitutivity Puzzles* (as well as: *Partee's paradox*) are fake constructions, resulting from adopting *Dogma1* (*Referential Doctrine*) and misunderstanding *ToBe*-expressions as identity statements, a misconception, leading to serious meaning anomalies (like: *The Actor Anomaly*). Such constructions are only possible in Languages, where Copula are used to express the relation between *Subjects* and *Predicates*. Arabic not being one of them, defeats the idea, that Copula-related issues can have any universal Semantic bearing at all.

2- Russell's view of the denotation of Indefinite Descriptions (*Dogma2*), backed up by the excessively materialistic notion of reference, prescribed by *Dogma1,* leads, in turn, to another meaning anomaly: *Inadequacy of existential translations*, which shows the necessity of admitting mental entities as referents in ordinary Natural Language use.

3- The fact, that, as per *AnitDogma2*, an Arabic Noun Sentence, asserting some property about an Indefinite Description, can only be grammatically correct, if the Indefinite Description is preceded by an existential assertion related to it, questions the purpose of Russell's idea (*Dogma2*) of substituting existential quantifiers for Indefinite Descriptions.

4- Russell's notion of *incomplete Symbols* is fundamentally flawed: In Arabic, Nouns and Verbs will always hold meaning nuances, independent of any context of use, even when Descriptions using them *'dissolve'* in existential statements as Russell suggests.

5- In Arabic also: Since Descriptions are *not* replaced by existential quantifiers, scope ambiguities, usually attributed by Logicians to *Semantic operators*, are caused, when they occur, either by lexical ambiguities or ambiguities due to meaning, type and location of used *syntactical operators*. Because *the logical collective* cannot be modeled via Indefinite Descriptions, some studied forms of scope ambiguities do not occur in Arabic in the first place.

6- Fregean Logics does not permit, to model *the linguistic collective*, so that not all possible meanings of a Sentence containing quantifiers in Arabic or English is captured, *in principle*.

7- This Logics also *relativizes* meanings of Symbols, in principle, so that acceptance of semantically invalid Natural Language Sentences becomes possible. This feature distorts, as per Skolem, also some important set-theoretic notions, rendering them unattainable.

All this and much more, omitted here to avoid unnecessary length, amounts to saying that *Dogmas* of Logicians cause formal Systems adopting them to *over- and under-accept* Natural Language Sentences, making them false imitations of how Language, especially Arabic, works.

What about Computation?

If we mean to ask this question from the perspective of Computational Linguistics, we have seen, that any computational System implementing *AntiDogmas* of Arabic in form of a *Diacritic-Grammar,* supported by adequate *I'rab-procedures* (Section (ii), part A), fulfills the following properties:

    i. It enables overcoming processing complexity through extensive use of *meaning-particles*, which help in disambiguating denotations/roles of words, phrases and Sentences, avoiding thus the use of brute-force recognition procedures.

    ii. Meta-Symbolic Layers of Arabic, embedded in such System, express Language- as well as Meta-Language Rules in a way, facilitating efficient higher order deductions, going beyond anything computationally possible using first order Logics.

    iii. Classifying an Arabic Sentence without using *ḥarakāt* is exponentially more expensive, in most cases, than when *ḥarakāt* guide derivations in an *I'rab-procedure*, which is another incentive of the said System.

However: We didn't look at Computation in this work to facilitate NLP.

Putting features of Arabic into action in the context of *SAT* problems was our main aim in parts *B* and *C*.

The obvious reason: *SAT-Solution methodologies also depend on the way we understand the relation between Syntax (a CNF formula) and Semantics (its Canonical Truth Table).*

Our investigations revealed interpretation-independent Semantic patterns (*meaning particles*) hidden behind names of literals used in *CNF* formulas.

In Literal *'X_i'*: While Symbol *'X'* denotes always *'The Unknown Value'*, index *'i'* reveals a fixed pattern present in the *Canonical Truth Table*. When a *Satisfaction-procedure* sees indices in Literal names as reflecting lengths of such patterns, construction of small *FBDDs* is possible through inducing a specific linear order between those indices by means of simple renaming and sorting operations.

This paper strengthens, hence, the belief, that the *NP-problem* cannot be solved using means of Logics alone. It shows in fact, that those means of Logics (engraved in the above *Dogmas*) were the reason, why the *NP-problem* existed in the first place.

It challenges the false idea, that Logics cannot profit from Natural Language: *Inspired by the way Arabic uses Meta-Symbols to determine meanings of words, phrases and Sentences, similar Meta-Symbols were found in CNF formulas, waiting there to be discovered for more than half a century now.*

Our work has also a practical side, which challenges the relevance of known Lower Bounds on *FBDD* sizes. We not only demonstrate their theoretical inapplicability to the here described methods, but show practical, reproducible evidence, that sizes of *FBDDs*, created for some known type of multiplication circuits, *don't increase exponentially* with the number of used variables in the corresponding *3CNF* formulas.

Findings presented here should not be perceived as a defensive refutation of Logicians views towards Natural Language only, but foremost as a counter-attack, led by Arabic and aimed at the heart of current computation-paradigms in general and those used in NLP in particular: One way to reconcile Natural Language with Logics again is to go a step back to Natural Deduction- (or Aristotelian) Systems and try to integrate Arabic meta-symbolic Layers into them. Another way is to formalize Arabic Grammar machinery directly using *AntiDogmas*, utilizing the abundance of *meaning-particles* to reduce Semantic issues to syntactic ones, whenever possible. This direction builds also upon Chomsky's ideas. Both research directions are currently being investigated by the author and may result in future publications.

We genuinely believe, that our work will remain unrecognized for many more years to come, but published it, nevertheless, as a reminder of the fact, that miracles of Nature, like the ones existing in Natural Languages, notably Arabic, remain strongly present, applicable, in imitation, to all areas of modern technology and life, even when people deliberately deny their existence.

ACKNOWLEDGEMENT

I would like first to thank all my family, foremost father and mother, for the patience and understanding they have shown me during the busy years of my concentration on the study of Arabic and computation.

Many thanks go also to *GridSAT Stiftung*, a German non-profit foundation which fosters non-compromising Open- and Free Everything, and all its supporting members for their assistance throughout this time.

To the esteemed Noam Chomsky go special thanks and acknowledgments for finding time and interest to read briefings of my work and answering my sometimes naive questions.

My thanks go foremost to Allah (*SWT*), Who guided my ideas throughout time to what they have become today, showing me the significance of that conscious-level, obscured between certainty and dogmatism, which motivates humans to search for Truth all their lives, while knowing that their means will always be limited. He calls this free, courageous spirit in His wise Quran: *'Iman',* a word which has no adequate translation in any other Language.